\pdfoutput=1

\documentclass[11pt]{article}

\usepackage[]{ACL2023}

\usepackage{times}
\usepackage{latexsym}
\usepackage{graphicx} 
\usepackage{amsmath}  
\usepackage{amssymb}
\usepackage{booktabs, tabularx}
\usepackage{caption}
\usepackage{subcaption} 
\usepackage{multirow}
\usepackage{amsfonts}
\usepackage{bm}
\usepackage{tikz}
\usepackage{listings}
\usepackage{linguex}

\definecolor{Blue}{RGB}{115,191,249}

\newcommand{\bluecirc}[1]{%
  \tikz[baseline=(char.base)]{%
    \node[
      draw=Blue,   
      fill=Blue,   
      thick,
      circle,
      minimum size=3mm,
      inner sep=0pt,
      text=black,
      font=\small\ttfamily\bfseries
    ] (char) {#1};%
  }%
}

\newcommand{\bluecircc}[1]{%
  \tikz[baseline=(char.base)]{%
    \node[
      draw=Blue,   
      fill=Blue,   
      thick,
      circle,
      minimum size=3mm,
      inner sep=0pt,
      text=black,
      font=\tiny\ttfamily\bfseries
    ] (char) {#1};%
  }%
}

\usepackage[T1]{fontenc}

\usepackage[utf8]{inputenc}

\usepackage{microtype}

\usepackage{inconsolata}

\usepackage{titlesec}

\title{Tug-of-war between idioms' figurative and literal interpretations in LLMs}

\author{
Soyoung Oh\textsuperscript{1}, Xinting Huang\textsuperscript{1}, Mathis Pink\textsuperscript{2}, and Michael Hahn\textsuperscript{1, $\dagger$}, Vera Demberg\textsuperscript{1,3, $\dagger$} \\
Saarland University\textsuperscript{1}, Max Planck Institute for Software Systems\textsuperscript{2}, \\ Max Planck Institute for Informatics\textsuperscript{3} \\
\texttt{\{soyoung, xhuang, mhahn, vera\}@lst.uni-saarland.de}, \\ \texttt{mpink@mpi-sws.org}
}

\begin{document}
\maketitle

\begingroup
  \renewcommand\thefootnote{$\dagger$}
  \footnotetext{Co‐corresponding authors.}
\endgroup

\begin{abstract}
Idioms present a unique challenge for language models due to their non-compositional figurative interpretations, which often strongly diverge from the idiom's literal interpretation. In this paper, we employ causal tracing to systematically analyze how pretrained causal transformers deal with this ambiguity. We localize three mechanisms: (i) Early sublayers and specific attention heads retrieve an idiom's figurative interpretation, while suppressing its literal interpretation. (ii) When disambiguating context precedes the idiom, the model leverages it from the earliest layer and later layers refine the interpretation if the context conflicts with the retrieved interpretation. (iii) Then, selective, competing pathways carry both interpretations: an intermediate pathway prioritizes the figurative interpretation and a parallel direct route favors the literal interpretation, ensuring that both readings remain available. Our findings provide mechanistic evidence for idiom comprehension in autoregressive transformers\footnote{Code and dataset are available at \url{https://github.com/sori424/idiom_processing}}.

\end{abstract}

\section{Introduction}
\begin{figure}[!h] 
\centering
\includegraphics[width=0.5\textwidth]{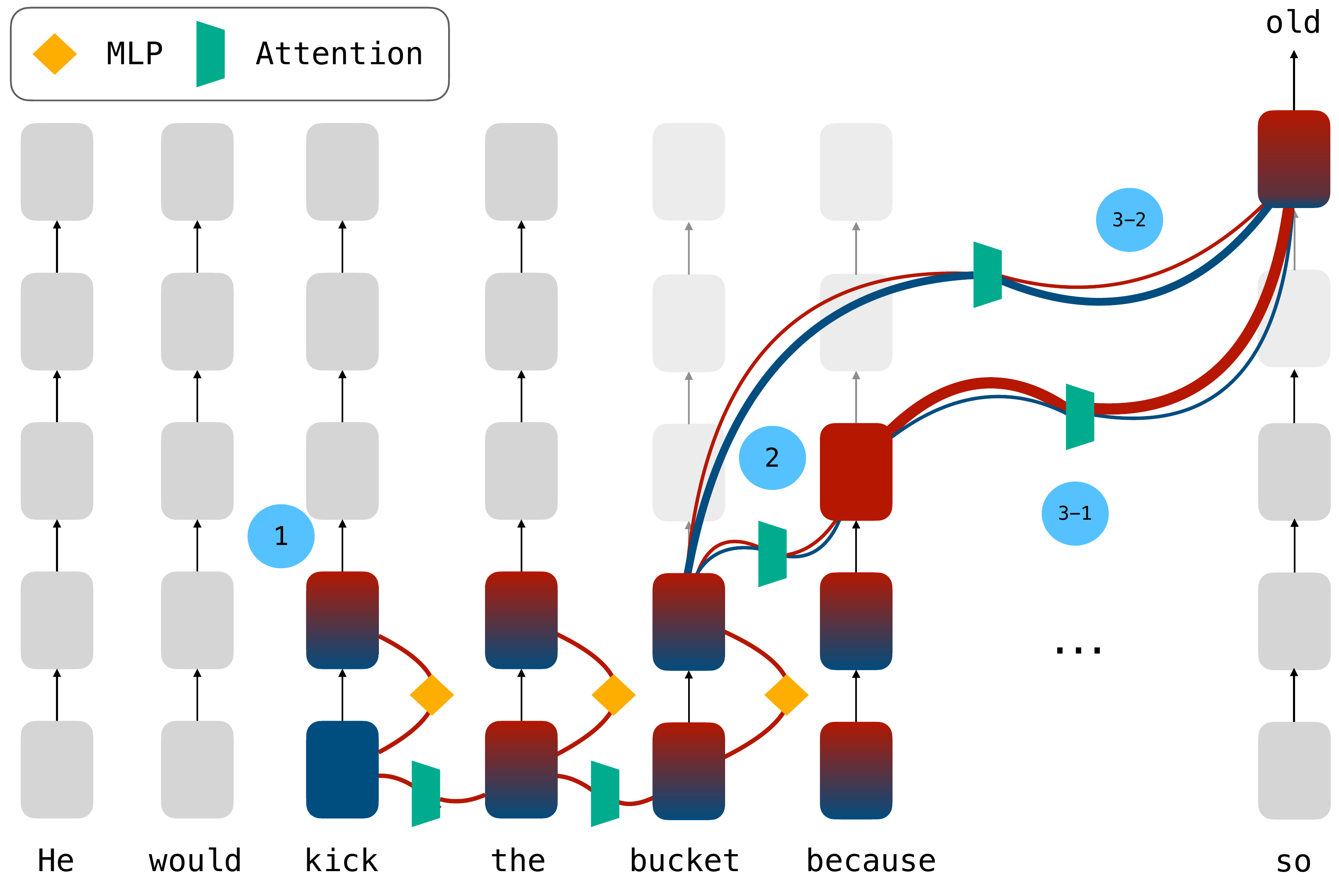}
  \caption{The \textcolor[RGB]{166,44,22}{\textbf{figurative}} and \textcolor[RGB]{30,76,124}{\textbf{literal}} interpretations are highlighted in the blocks and paths. We find three main steps for idiom processing: \bluecirc{1} \textbf{Idiom retrieval step:} Early layers (i.e., layers 0-3) attention and MLP are actively retrieving the idiom's figurative interpretation while storing both figurative and literal interpretations in the residual stream. 
  \bluecirc{2} \textbf{Selective interpretation step:} At the token immediately following the idiom span, the model begins to encode a representation that favors the figurative interpretation over the literal one, starting from the middle layers.
  \bluecirc{3} \textbf{Interpretation routing:} For final prediction, the model passes literal interpretation via both \textbf{a direct compositional semantic path} (\bluecircc{3-2}), as well as the intermediate pathway that prioritizes the figurative interpretation (\bluecircc{3-1} \textbf{figurative path}).}
  
  \label{fig:summary}
\end{figure}

Idioms pose a challenge to standard semantic composition because, as multi-word expressions, their figurative meanings don't follow from the literal meanings~\cite{beck2020context}. For instance, ``kick the bucket''  conveys the figurative interpretation ``to die'', yet its literal sense ``physically kicking a container'' remains semantically available. 
Psycholinguistics has investigated how humans understand idioms, whether they access the figurative interpretation directly or first have to access the literal interpretation of the idiom and only in a later stage suppress that interpretation and access the figurative interpretation. These findings have led to various models describing human idiom processing~\cite{gibbs1980spilling, CACCIARI1991217, bobrow1973catching, cacciari1988comprehension}.

Yet for LLMs, it's unclear how they process idioms, potentially limiting further improvements in their idiom comprehension~\cite{kabra2023multi, liu2023multilingual, knietaite2024less}. One explanation may lie in how the model transforms raw input token embeddings into richer representations~\cite{tian2023idioms, haviv2022understanding, dankers2022can}. Within the autoregressive transformer, token representations are incrementally refined through a series of residual layers, which consist of multi-head self-attention (MHSA) and MLP sublayers, each adding their outputs to update the representation~\cite{vig2020investigating}. This multi-stage transformation suggests that idiom processing proceeds via a sequential shift from compositional to non-compositional representations~\cite{haviv2022understanding, dankers2022can}. 


In this paper, we first aim to identify how the idiom is interpreted by a transformer Llama3.2-1B (we also run experiments on Llama3.1-8B, Qwen2.5-0.5B, Qwen2.5-7B), where its figurative interpretation is located, and how it is retrieved by the model compared to its literal counterpart. To this end, we pinpoint specific components of the model that have specialized in processing idioms by boosting their figurative interpretation, while suppressing the compositional semantic literal interpretation. Having identified these, we proceed to investigating the information flow in the model, with the goal of identifying specific pathways through which the figurative interpretation is passed forward to the final prediction, and whether this differs from the pathways of passing forward information about an idiom's literal interpretation. Additionally, given that the context plays a key role in disambiguating the interpretation of an idiom~\cite{mi2024rolling, fazly2009unsupervised, dankers2022can, holsinger2013representing}, we investigate how context interacts with lexical information of idioms.


We address these questions through knockout analyses \cite{nanda2023progress, wang2022interpretability, geva2023dissecting}, where we separately ablate activations of each component (layer-wise MHSA and MLP, individual attention heads) in the model to observe their importance for retrieving idiom's figurative interpretation, and contextual disambiguation. Furthermore, to trace the flow of these interpretations, we employ activation patching experiments~\cite{wang2022interpretability, meng2022locating, hanna2023does, conmy2023towards, stolfo2023mechanistic}, in which we ablate one or both interpretations from specific pathways within the model. Figure \ref{fig:summary} summarizes our main findings. Our contributions are:

\begin{itemize}
    \setlength{\itemsep}{0pt}
    \item We find that early layers and idiom specific heads causally retrieve figurative interpretations, while suppressing the corresponding literal counterparts (Section~\ref{sec:disambiguation}).
    \item We identify a selection mechanism at the token immediately following the idiom, where an intermediate path carries a strong flow toward the figurative interpretation. In parallel, a bypass pathway preferentially carries literal information directly to the final prediction, circumventing the intermediate pathway (Section~\ref{sec:information_flow}).
    \item We localize where context resolves idiomatic ambiguity, showing that early MLP/MHSA layers suppress figurative retrieval under literal context, while later MHSA integrates surrounding tokens to refine the context consistent interpretation (Section~\ref{sec:context}).
\end{itemize}
\section{Idiom interpretation task}
\label{sec:data_gen}


We conduct our experiments in two setups. First, we investigate how language models process idiomatic expressions, which permit both figurative and literal interpretations. In the second setup, we introduce disambiguating context by adding preceding sentences.

We formalize a next-token prediction task and construct a dataset. Idioms are embedded into a sentence template designed to reveal their interpretation through continuation \texttt{`X (would) IDIOM because X was so/too/a/the'}\footnote{Here, \texttt{X} is instantiated with pronouns and morphologically adapted; X is instantiated with various pronouns (he/she/it/they); the word ``would'' is inserted only in some of the idioms to make the sentence sound more fluent; idioms are morphologically fit into the sentence; depending on sg/pl form of X, the second part of the template contained was or were.}. This template preserves idiom ambiguity while eliciting causal completions (e.g., for `kick the bucket', figurative completions include `old/sick', while literal completions include `angry/mad'.) Since there is no single ground truth continuation, we automatically construct token sets consistent with either the figurative or literal interpretation. The tokens in each set are mutually exclusive and chosen for their high likelihood of association with the respective interpretation. Below, we outline the dataset construction process, and see Figure~\ref{fig:datagen}.


\noindent\textbf{1.~Idiom Extraction:} We select 245 idioms from a psycholinguistic paper~\cite{cronk1993idioms}, each of which received high literality (humans rated the literal meaning as plausible). This ensures  ambiguity, which is necessary to evaluate the model's change between literal and figurative interpretations.

\noindent\textbf{2.~Sentence Generation:} We generate
a literal paraphrase and a figurative paraphrase for each idiom using the Llama3.3-70B-instruct model~\cite{grattafiori2024llama}. The sentences below are three examples of the generated \textit{ambiguous sentence} ($s_a$), \textit{figurative paraphrase} ($s_f$), and \textit{literal paraphrase} ($s_l$): 

\ex.[$s_a$] \small{He would \textit{kick the bucket} because he was so}
    
\ex.[$s_f$] \small{He would \textit{die} because he was so}

\ex.[$s_l$] \small{He would \textit{kick the container} because he was so}    


\noindent\textbf{3.~Token Set Generation:} To generate 
sets of next-word completions that are indicative of the figurative vs.~the literal interpretation, we prompt the Llama3.3-70B-instruct model with paraphrases ($s_f$, $s_l$) and obtain logit distributions. Let $z^{(f)}$ and $z^{(l)}$ denote the logits corresponding to $s_f$ and $s_l$ as input, then we compute the element-wise difference between the two sets $\Delta z_v = z_v^{(f)} - z_v^{(l)}$, where $v \in V$ is the vocabulary size of the model. Since tokens that score high in one paraphrase tend to score low in the other, $\Delta z_v$ captures each token’s interpretation-specific relevance. We take the 20 tokens with the largest $\Delta z_k$ as the figurative candidate set $C_f$, and the 20 with the smallest $\Delta z_k$ as the literal candidate set $C_l$.


\noindent\textbf{4.~Context Generation:} Using each paraphrased sentence, we generate a preceding context sentence that disambiguates the interpretation of the subsequent idiom. The examples below illustrate the generated figurative context ($FC$) and literal context ($LC$).

\ex.[$FC$] \small{His breath rattled after every step. He would die because he was so}
    
\ex.[$LC$] \small{The vending machine ate his coins. He would kick the container because he was so}




\section{Methodology}
\label{sec:method}
\subsection{Information flow in transformers}

Given an input token sequence $\mathbf{t}=[t_1,\dots,t_N]$ over a vocabulary $V$, each token is embedded as $\mathbf{x}_i^{0}=\mathrm{PE}(\mathrm{emb}(t_i), i)\in\mathbb{R}^d$, where $\mathrm{PE}$ denotes a positional encoding function that injects position $i$ into the token embedding~\cite{vaswani2017attention}, and $d$ is the model's hidden dimension. From layer 0, these vectors $\mathrm{x}_i^{0}$ are carried forward and accumulated over the subsequent $L$ layers. At each layer $\ell$, the hidden state for token $i$ is updated from its previous value $\mathrm{x}_i^{\ell-1}$ by adding the multi-head self-attention (MHSA) output $a_i^\ell$ and MLP output $m_i^\ell$: 
\[
\mathrm{x}_i^{\ell} = \mathrm{x}_i^{\ell-1} + a_i^{\ell} + m_i^{\ell}.
\]

\noindent\textbf{MLP Sublayers}\indent Every MLP sublayer computes a local update for each representation:
\[
m_i^\ell = W_F^\ell \, \sigma\Bigl(W_I^{\ell}(a_i^\ell + \mathrm{x}_i^{\ell-1})\Bigr),
\]
where $W_I^\ell \in \mathbb{R}^{d_{ff} \times d}$ and $W_{F}^\ell \in \mathbb{R}^{d \times d_{ff}}$ are parameter matrices with inner-dimension $d_{ff}$ and $\sigma$ is a nonlinear activation function.



\noindent\textbf{MHSA Sublayers}\indent Each attention layer's output $a_i^{\ell}$ aggregates information of multiple parallel attention heads $h$. The parameter matrices of each MHSA sublayer comprise three projection matrices $W_Q^{\ell},\,W_K^{\ell},\,W_V^{\ell}\in\mathbb{R}^{d\times d}$, and $W_O^{\ell} \in\mathbb{R}^{d\times d}$. The columns of each projection matrix and rows of the outputs matrix can be split into $H$ equal parts, corresponding to the number of attention heads $W_Q^{\ell,j},\,W_K^{\ell,j},\,W_V^{\ell,j}\in\mathbb{R}^{d\times \tfrac{d}{H}}$, and $W_O^{\ell,j}\in\mathbb{R}^{\tfrac{d}{H}\times d}, \, j=[1, H]$. 
Let $X^{\ell-1}\in\mathbb{R}^{N\times d}$ be the matrix of all token representations at layer $\ell-1$, $A^{\ell,j} \in \mathbb{R}^{N\times N}$ encodes the weights computed by the $j$-th attention head at layer $\ell$, where it's masked to a lower triangular matrix. The MHSA output is the sum over individual attention heads matrices:


\[
a_i^{\ell} = \sum_{j=1}^H \big( A_{i}^{\ell,j} X^{\ell-1} W_V^{\ell,j} \big) W_O^{\ell,j}.
\]


After $L$ layers, the final hidden state $\mathrm{x}_T^L$ is mapped to vocabulary logits via an unembedding matrix $W_U \in \mathbb{R}^{|V|\times d}$, so that $y = W_U\,\mathrm{x}_T^L \in \mathbb{R}^{|V|}$, and $\mathrm{softmax}(y)$ defines the next‐token distribution.  Thus we can view a transformer as a computational graph $G\colon\mathcal X\to\mathcal Y$, where token embeddings, MHSA sublayers, MLP sublayers, and the unembedding all communicate via the residual stream.  We refer to~\citet{vaswani2017attention, geva2023dissecting} for further details.

\subsection{Knockout}

If we think of a model as a computational graph $G$, then certain behaviors of the model can be attributed to specific subgraphs $G_{sub} \subseteq G$~\cite{rai2024practical}. Identifying such subgraphs provides a valuable mechanism for interpretability, enabling the isolation of individual components and their contributions to the model’s behavior. A knockout removes specific components to quantify their contribution to particular behaviors~\cite{nanda2023progress, wang2022interpretability, geva2023dissecting}. Every activation in $G_{\text{sub}}$ can be replaced by either zero or reference mean:
\[
\mathrm{x}_i^\ell \;\longmapsto\; \tilde{\mathrm{x}}_i^\ell =
\begin{cases}
0 \quad \text{or} \quad \mu_i^\ell & \text{if } (\ell, i) \in G_{\text{sub}}, \\
\mathrm{x}_i^\ell & \text{otherwise}.
\end{cases},
\]
then $G$ with those components ``knocked out'' is
\[
G_{\text{knock}} = G\left( \dots, \tilde{\mathrm{x}}_i^\ell, \dots \right).
\]

\noindent By replacing the activations of these components, we eliminate the contributions of $G_{sub}$ while keeping other computations in $G$ fixed. 





\subsection{Activation patching}
Activation patching is a technique to identify which activations in a model contribute to a particular output~\cite{wang2022interpretability, meng2022locating, hanna2023does, conmy2023towards, stolfo2023mechanistic}. This involves running the model on input A, then selectively patching certain activations with those obtained from a paired input B:
\[\resizebox{0.5\textwidth}{!}{$
G_{\text{patch}}(A \hookleftarrow B) = G \bigl( \mathrm{x}^1_i(A), \dots, \mathrm{x}^{\ell}_i(B), \mathrm{x}^{\ell + 1}_i(A), \dots \bigr).$}
\]

By comparing the original output for input A with the altered output after patching, we can precisely measure how much the introduced activations from input B shift the model’s output. 

\subsection{Metric for measuring interpretation shift}

To measure whether the model assumes the figurative or literal interpretation when given each prompt \(s \in \{s_a, s_f, s_l\}\), we define two scores based on cumulative probabilities:
\[
F(s) \;=\; \sum_{c_f \in C_f} p(c_f \mid s); \;
L(s) \;=\; \sum_{c_l \in C_l} p(c_l \mid s).
\]
After applying an intervention, we quantify its causal effect:
\[
\Delta I(s) \;=\; I_{\mathrm{interv}}(s) \;-\; I_{\mathrm{origin}}(s), \, I \in \{F, L\},
\]
where \(\Delta I <0\) means the intervention disrupts the corresponding interpretation, implying the component is necessary; \(\Delta I = 0\) shows no effect; \(\Delta I  > 0\) implies the component is not required (or may even inhibit) that particular reading.  
\section{Localizing idiom's figurative interpretation retrieval}
\label{sec:disambiguation}
We deliberately preserve idiomatic ambiguity by analyzing idioms without the context, so we can localize where, and through which components, the model shifts between figurative and literal interpretations.

\subsection{Probing sublayers through knockout}
To locate the mechanisms for retrieving figurative interpretations, we knock out the activations at idiom span tokens by replacing them with their mean values computed over all $s_a$. By knocking out the activations across each layer, we assess how the probabilities of predicting specific candidate tokens ($\Delta F(s_a), \, \Delta L(s_a)$) change. To narrow down the points to which we can ascribe a functional role for a specific computation component, we limit the intervention to a MLP and MHSA in each block. 


\begin{figure}[h]
  \centering  
  \begin{subfigure}[b]{0.23\textwidth}
    \centering
    \includegraphics[width=\textwidth]{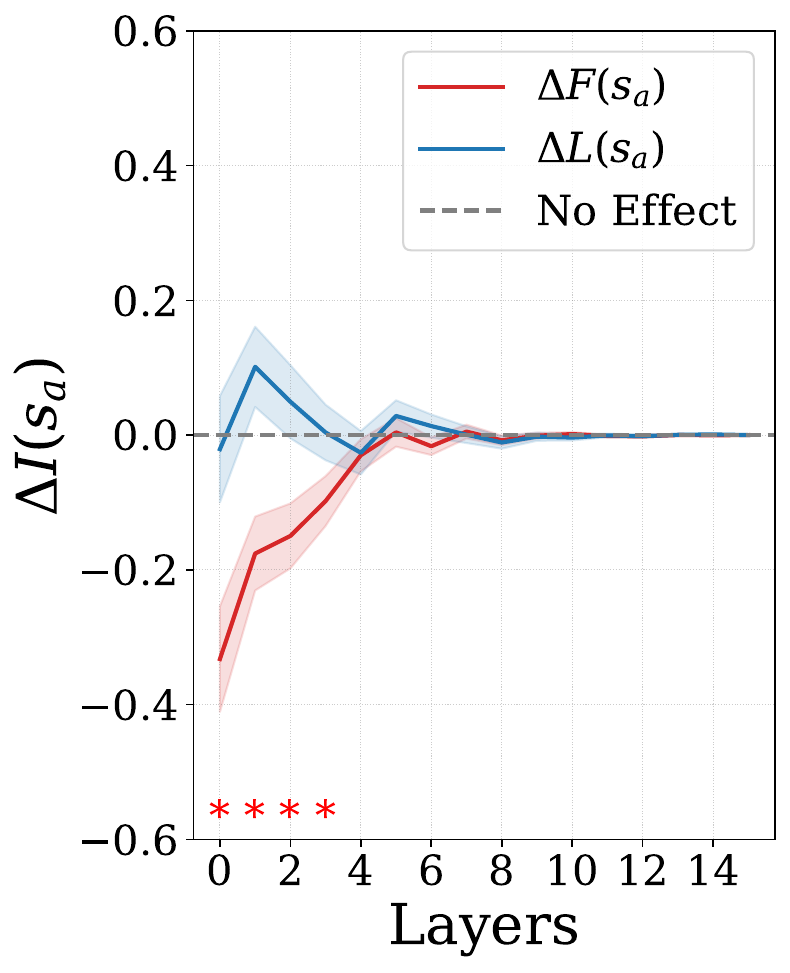}
    \caption{MLP}
    \label{fig:mlp_knockout}
  \end{subfigure}
  \begin{subfigure}[b]{0.23\textwidth}
    \centering
    \includegraphics[width=\textwidth]{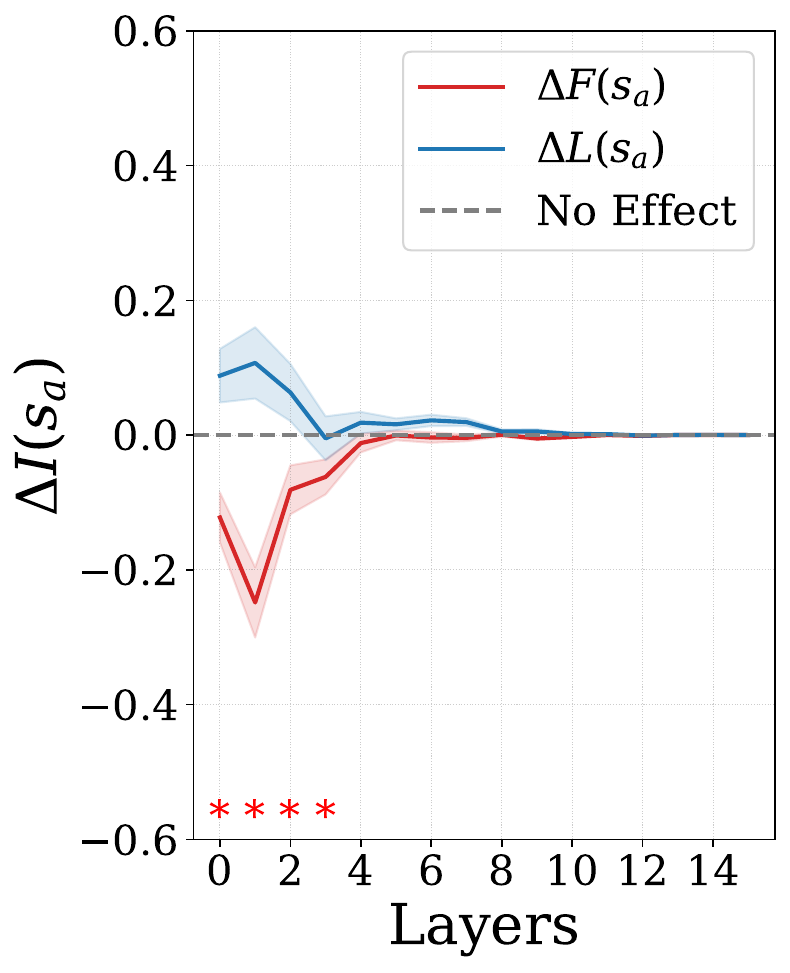}
    \caption{MHSA}
    \label{fig:attn_knockout}
  \end{subfigure} 
      \begin{subfigure}[b]{0.23\textwidth}
    \centering
    \includegraphics[width=\textwidth]{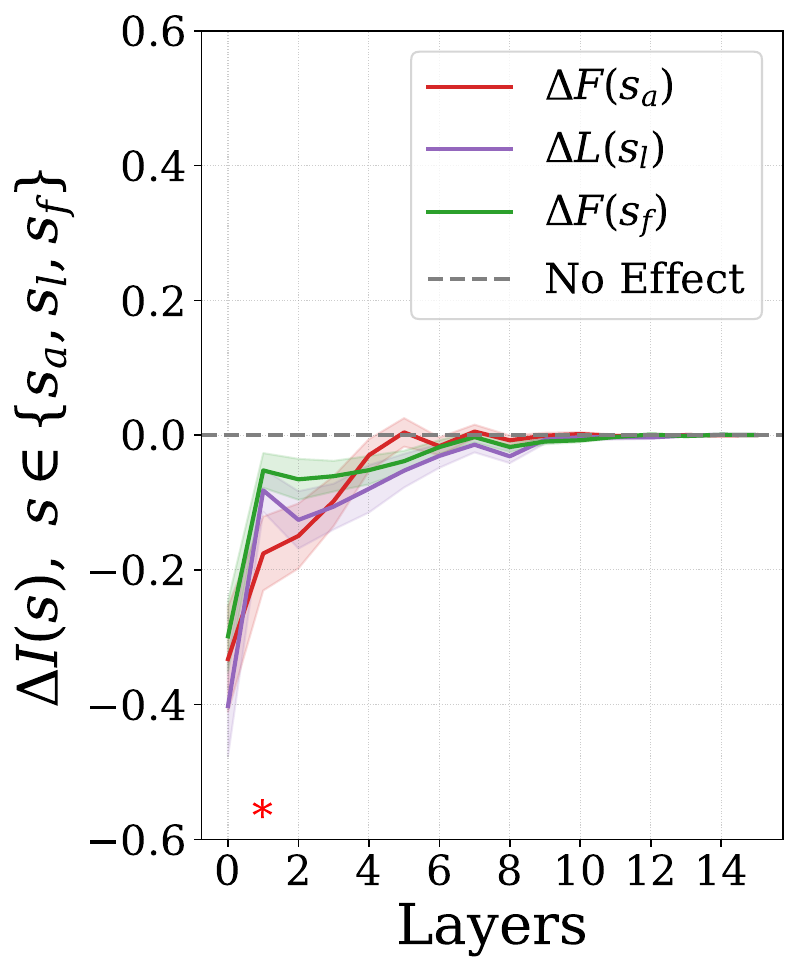}
    \caption{MLP}
    \label{fig:mlp_knockout_all}
  \end{subfigure}
    \begin{subfigure}[b]{0.23\textwidth}
    \centering
    \includegraphics[width=\textwidth]{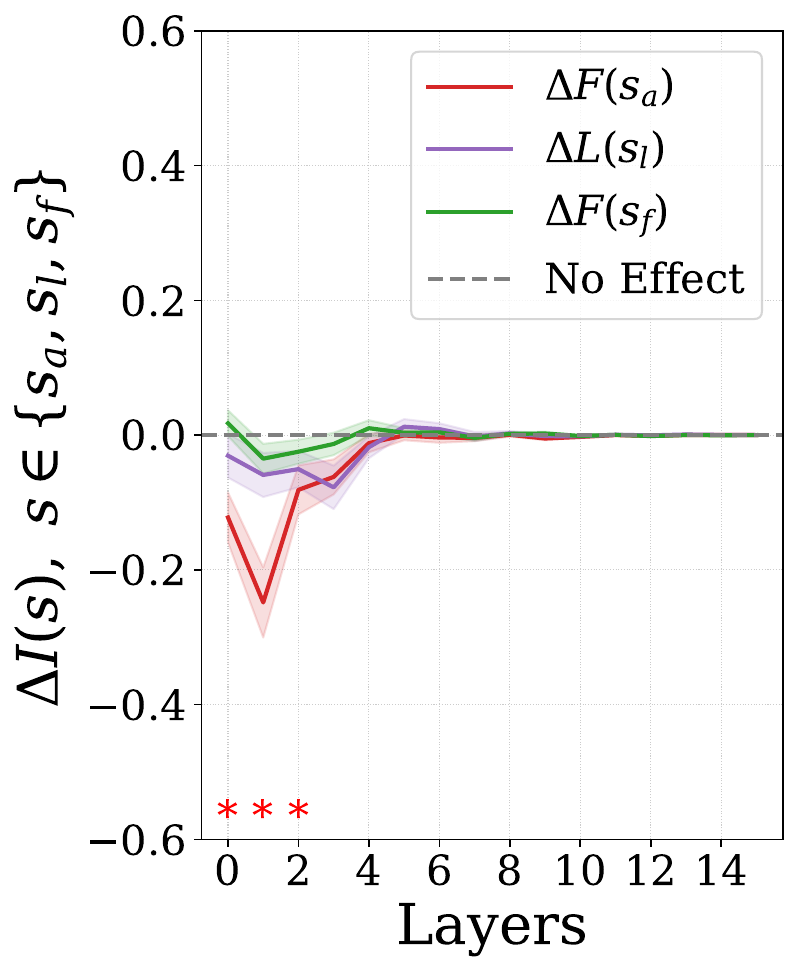}
    \caption{MHSA}
    \label{fig:attn_knockout_all}
  \end{subfigure}

  \caption{Sublayer-wise interpretation shift $\Delta I(s)$ after ablating activations at idiom span, for sentences $s \in \{s_a, s_f, s_l\}$. \textbf{Y-axis:} Mean values of \textcolor[RGB]{31,119,180}{$\Delta L(s_a)$}, \textcolor[RGB]{214, 39, 40}{$\Delta F(s_a)$}, \textcolor[RGB]{148,103,189}{$\Delta L(s_l)$}, \textcolor[RGB]{44,160,44}{$\Delta F(s_f)$} with 95\% confidence intervals across idiom sentences. \textbf{X-axis:} Layers. \textbf{Gray dashed line:} $\Delta I = 0$ (no effect). The red asterisk (\textcolor{red}{*}) marks layers where the difference between \textcolor[RGB]{31,119,180}{$\Delta L$} and \textcolor[RGB]{214,39,40}{$\Delta F$} exceeds the average difference across layers (paired $t$-test, $p<0.05$).}
  \label{fig:knockout_so}
\end{figure}

\noindent\textbf{Results}
\indent Figure~\ref{fig:mlp_knockout} shows by how much the model's interpretation of the idiom changes when the MLP sublayer is knocked out. We observe substantial drops of $\Delta F(s_a)$ when knocking out early MLP sublayers (0-2). At the same time, we see an increase in the probability of the literal completions: $\Delta L(s_a)$ rises above zero. And these shifts are significantly different in 0-3 layers ($p<0.05$). This indicates that early layer MLPs are causally important for reading out the figurative interpretation while potentially suppressing the literal one. For later layers ($\geq$ 4), the knockout does not affect the model's interpretation, where the corresponding patching effect converges to zero. 

Figure~\ref{fig:attn_knockout} shows the effect of MHSA knockout, which is similar to what we observed with MLP sublayer knockout. Specifically, knocking out early layers MHSA (0-3) leads to a drop in $\Delta F(s_a)$ and increase in $\Delta L(s_a)$. The most significant difference between $\Delta F(s_a)$ and $\Delta L(s_a)$ is in layer 1, where $\Delta F(s_a)$ dropped by $-0.30$ and $\Delta L(s_a)$ increased by $0.16$ ($p<0.05$). After layer 4, knocking out MHSA has little to no effect on interpretation, as the values converge toward zero.

Together, the two interventions track the \bluecirc{1} \textbf{idiom's figurative interpretation retrieval}: Early MHSA layers (0-3) gather and bind token-wise interactions, while early MLP layers (0-3) transform those bound representations into a figurative interpretation, while suppressing the literal one. \textcolor{black}{This early layer idiom interpretation retrieval aligns with the findings of~\citet{haviv2022understanding}, who argue that idiomatic information is accessed in the initial layers of LLMs during inference.} After layer 4, $\Delta F(s_a)$ and $\Delta L(s_a)$ stabilize to zero, indicating that later sublayers are not adding new information and that the representations of the interpretations at the idiom span tokens are no longer used by any subsequent tokens.

Moreover, these components are specifically essential for idiom processing, not generally important for semantic interpretation of a sentence. Figure~\ref{fig:mlp_knockout_all} shows that the drop from knocking out early MLP sublayers is significantly smaller in layer 1 for the two paraphrases (\textcolor[RGB]{148,103,189}{$\Delta L(s_l)$}, \textcolor[RGB]{44,160,44}{$\Delta F(s_f)$}) than for the idiomatic expression, (\textcolor[RGB]{214, 39, 40}{$\Delta F(s_a)$}), ($p<0.05$). While, at layer 0, the knockout effect is large for all expressions, as in this case, the MLP block is processing the semantics of the sentence. Figure \ref{fig:attn_knockout_all} shows a similar effect for the early MHSA layers (0-2) where literal paraphrases are not strongly affected by knockout (\textcolor[RGB]{148,103,189}{$\Delta L(s_l)$}, \textcolor[RGB]{44,160,44}{$\Delta F(s_f)$}) compared to idiomatic expression (\textcolor[RGB]{214, 39, 40}{$\Delta F(s_a)$}), ($p<0.05$). The crucial role of figurative interpretation retrieval in early layers is consistent across different models (Appendix~\ref{sec:knockout_others}).

\subsection{Identifying attention heads specific to retrieving figurative interpretations}
\label{sec:attn_head_retrieval}

Next, we investigated which attention heads are specialized for retrieving an idiom's figurative interpretation.
We knock out attention heads iteratively at the idiom span tokens and measure \(\Delta F(s_a)\), \(\Delta L(s_a)\). To operationalize this, we define \emph{idiomatic heads} (\(\mathcal{H}_{\mathrm{idiom}}\)) as heads with a large negative \(\Delta F(s_a)\) and positive \(\Delta L(s_a)\). 

\begin{figure}[h]
\centering
    \begin{subfigure}[b]{0.23\textwidth}
    \centering
    \includegraphics[width=\textwidth]{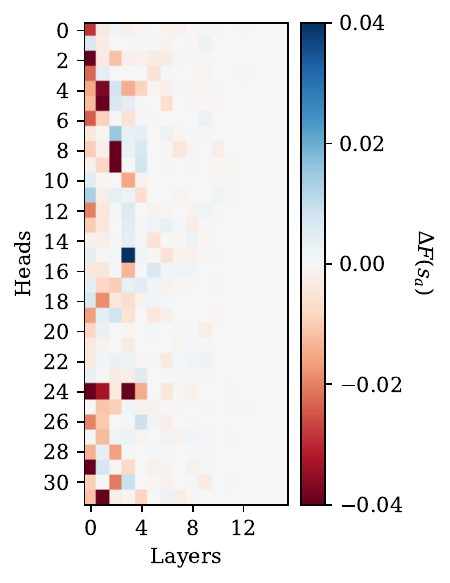}
    \caption{$\Delta F(s_a)$}
    \label{fig:attn_heads_disamb_fl}
  \end{subfigure}
    \begin{subfigure}[b]{0.23\textwidth}
    \centering
    \includegraphics[width=\textwidth]{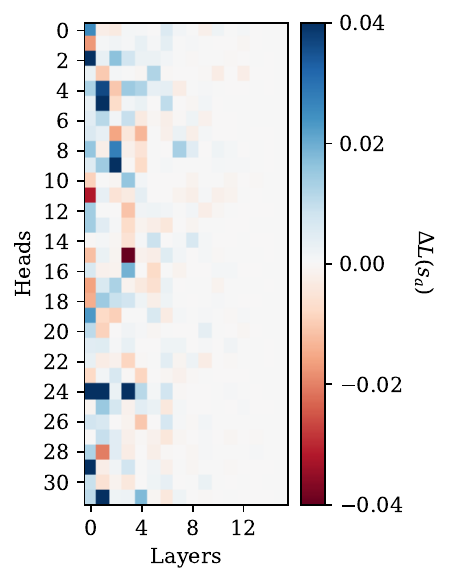}
    \caption{$\Delta L(s_a)$}
    \label{fig:attn_heads_disamb_lf}
  \end{subfigure}

  \caption{Heatmaps of the (a) $\Delta F(s_a)$ (b) $\Delta L(s_a)$ when ablating individual attention heads at the idiom span. \textbf{Idiomatic heads}: Heads those are crucial for retrieving the figurative interpretation of idiom; $-\Delta F(s_a)$ and $+\Delta L(s_a)$, simultaneously.}
  \label{fig:attn_heads_heatmap}
\end{figure}


We identify attention heads whose removal reliably decreases figurative probability ($-\Delta F(s_a)$), while increasing literal probability ($+\Delta L(s_a)$) for the idiom sentences. To identify which of these heads are consistently important across instances of $s_a$, we performed a nonparametric bootstrap with \(B = 1000\) resamples for each head's contribution to the interpretation. We then rank all heads by, on the one hand, \(-\Delta F(s_a)\), and, on the other hand, \(+\Delta L(s_a)\), and select the top 20 heads highly ranked on both orders (selected attention heads listed in Appendix Table~\ref{tab:topk}). Figure~\ref{fig:attn_heads_heatmap} shows that there is a subset of attention heads that simultaneously exhibits large \(-\Delta F(s_a)\) (Figure~\ref{fig:attn_heads_disamb_fl}) and  \(+\Delta L(s_a)\) (Figure~\ref{fig:attn_heads_disamb_lf}).

\subsection{Causal role of MLP sublayers and idiomatic heads in figurative reading}
\label{sec:causal_role}

To show that the components that we found (i.e., early MLP sublayers and $\mathcal{H}_{\mathrm{idiom}}$) are causally sufficient for figurative reading, we patch the activations on $s_a$ at idiom span tokens into the corresponding components when processing $s_l$: patching their activations from the idiom sentence into the literal paraphrase increases $\Delta F(s_l \hookleftarrow s_a)$ and decreases $\Delta L(s_l \hookleftarrow s_a)$, while patching random components has no effect.




\noindent\textbf{Results}\indent 
Patching these components boosts the figurative interpretation, as quantified by \(\Delta F(s_l \hookleftarrow s_a)\) ($M = 0.17, \, SD = 0.49$) while suppressing the literal interpretation, \(\Delta L(s_l \hookleftarrow s_a)\) ($M = -0.29, \, SD = 0.58$).
In contrast, patching random components barely affects the figurative interpretation (\(\Delta F(s_l \hookleftarrow s_a)\): $M = -0.008, \, SD = 0.09$) as well as literal one (\(\Delta L(s_l \hookleftarrow s_a)\): $M = -0.005, \, SD = 0.11$). 
Effects on the experiment and the control are significantly different  (\(p<0.05\)). These contrasting effects suggest that idiomatic components are specialized to enhancing the figurative interpretation while suppressing literal interpretations, so that replacing their activations injects a figurative interpretation and amplifies $\Delta F(s_l)$, while dropping $\Delta L(s_l)$.


\section{Tracing the interpretation flow}
\label{sec:information_flow}

\subsection{Locating idiom interpretation across token positions}
\label{sec:locate-token-pos}
After the retrieval process, the model encodes both literal and figurative interpretations in its residual stream. We next investigate how these two representations are weighted or integrated in the final prediction. To locate the representation of the interpretations, we calculate the mutual nearest-neighbor kernel alignment~\cite{pmlr-v235-huh24a, cho2024revisiting} between sentence embeddings of a corresponding idiom token span from $s_f$ and $s_l$ encoded by BGE M3~\cite{chen-etal-2024-m3} and hidden states from various token positions of idiom sentence $s_a$. This metric quantifies the similarity between representations, with higher values indicating greater semantic similarity.
\begin{figure}[ht]
\centering
    \begin{subfigure}[b]{0.23\textwidth}
    \centering
    \includegraphics[width=\textwidth]{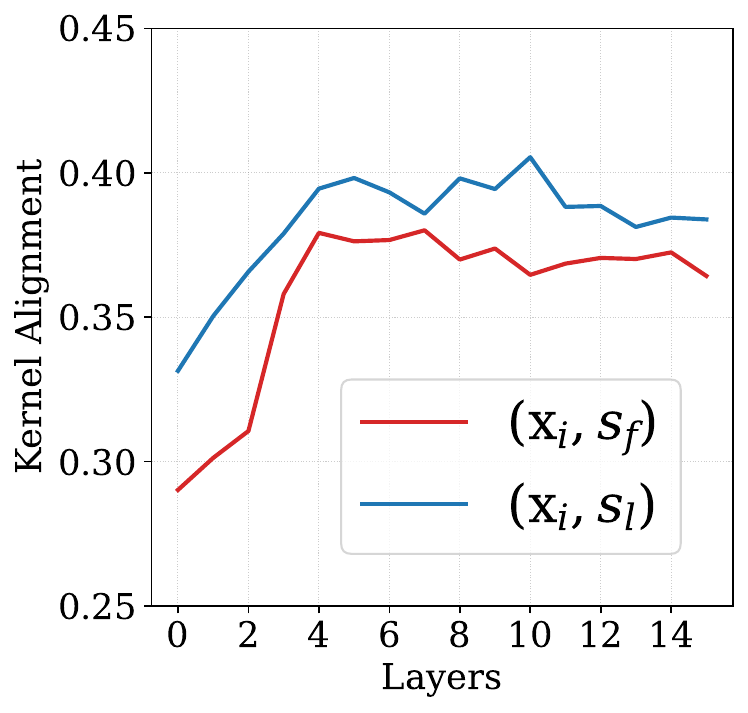}
    \caption{Last idiom token}
    \label{fig:ka_last_idiom}
  \end{subfigure}
    \begin{subfigure}[b]{0.23\textwidth}
    \centering
    \includegraphics[width=\textwidth]{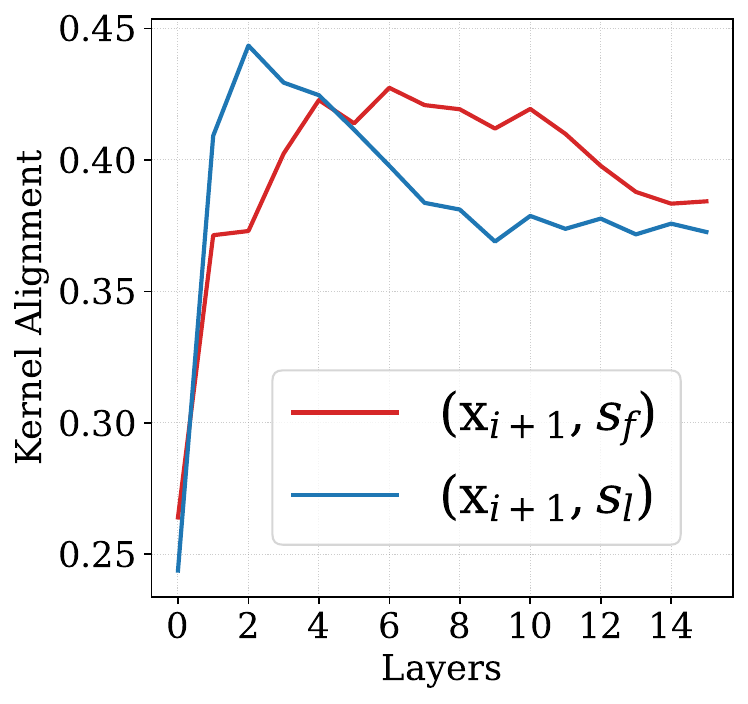}
    \caption{`because' token}
    \label{fig:ka_bc}
  \end{subfigure}
    \begin{subfigure}[b]{0.23\textwidth}
    \centering
    \includegraphics[width=\textwidth]{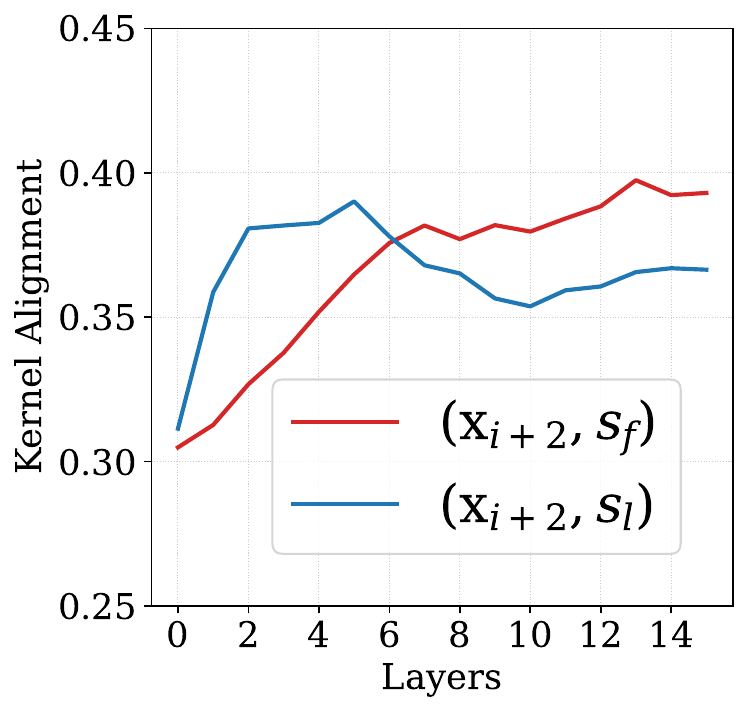}
    \caption{After `because' token}
    \label{fig:ka_after_bc}
  \end{subfigure}
      \begin{subfigure}[b]{0.23\textwidth}
    \centering
    \includegraphics[width=\textwidth]{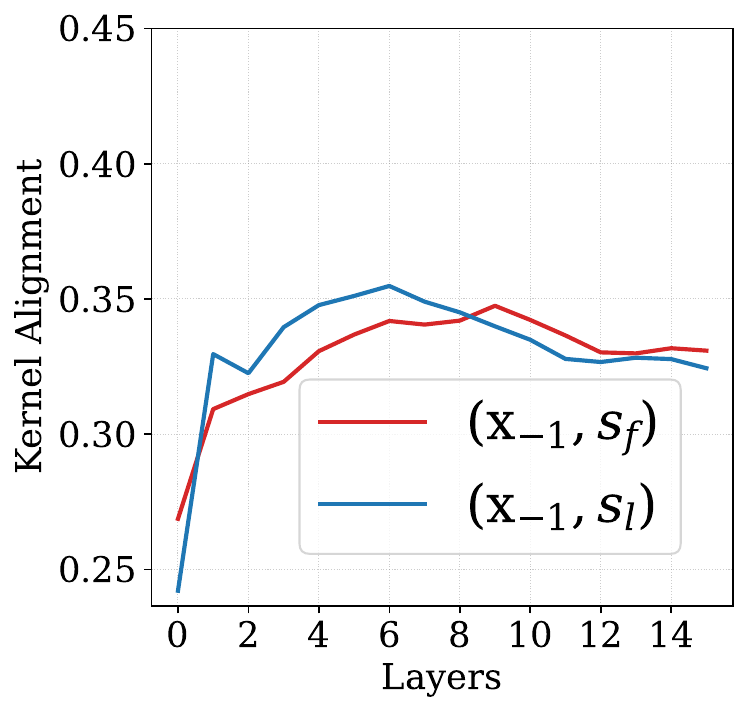}
    \caption{Last token of sentence}
    \label{fig:ka_last}
  \end{subfigure}
  \caption{Kernel alignment between hidden states ($\mathrm{x}$) extracted from four different token positions of $s_a$ and semantic embeddings of paraphrases ($s_f, \, s_l$).} 
  \label{fig:ka}
\end{figure}

\noindent\textbf{Results}\indent We observe that at the `because' token (Figure~\ref{fig:ka_bc}), the kernel alignment between the idiom's hidden states and its figurative interpretation surpasses that of the literal interpretation after layer 4. This suggests that by this point, the model has begun to favor the figurative interpretation (\bluecirc{2} \textbf{selective interpretation step}). Based on the results, we hypothesize an intermediate pathway where token position `because' plays a pivotal role in steering the idiom toward a figurative interpretation. In parallel, to keep the alternative interpretation available at the prediction (as in Figure~\ref{fig:ka_last}), we posit a direct route that favors the literal interpretation by more strongly preserving compositional semantic information.


\subsection{Analyzing layer-wise competing interpretation flow through activation patching}
\label{sec:competition_flow}
\begin{figure}[!h]
\centering
    \begin{subfigure}[b]{0.2\textwidth} 
    \centering
    \includegraphics[width=\textwidth]{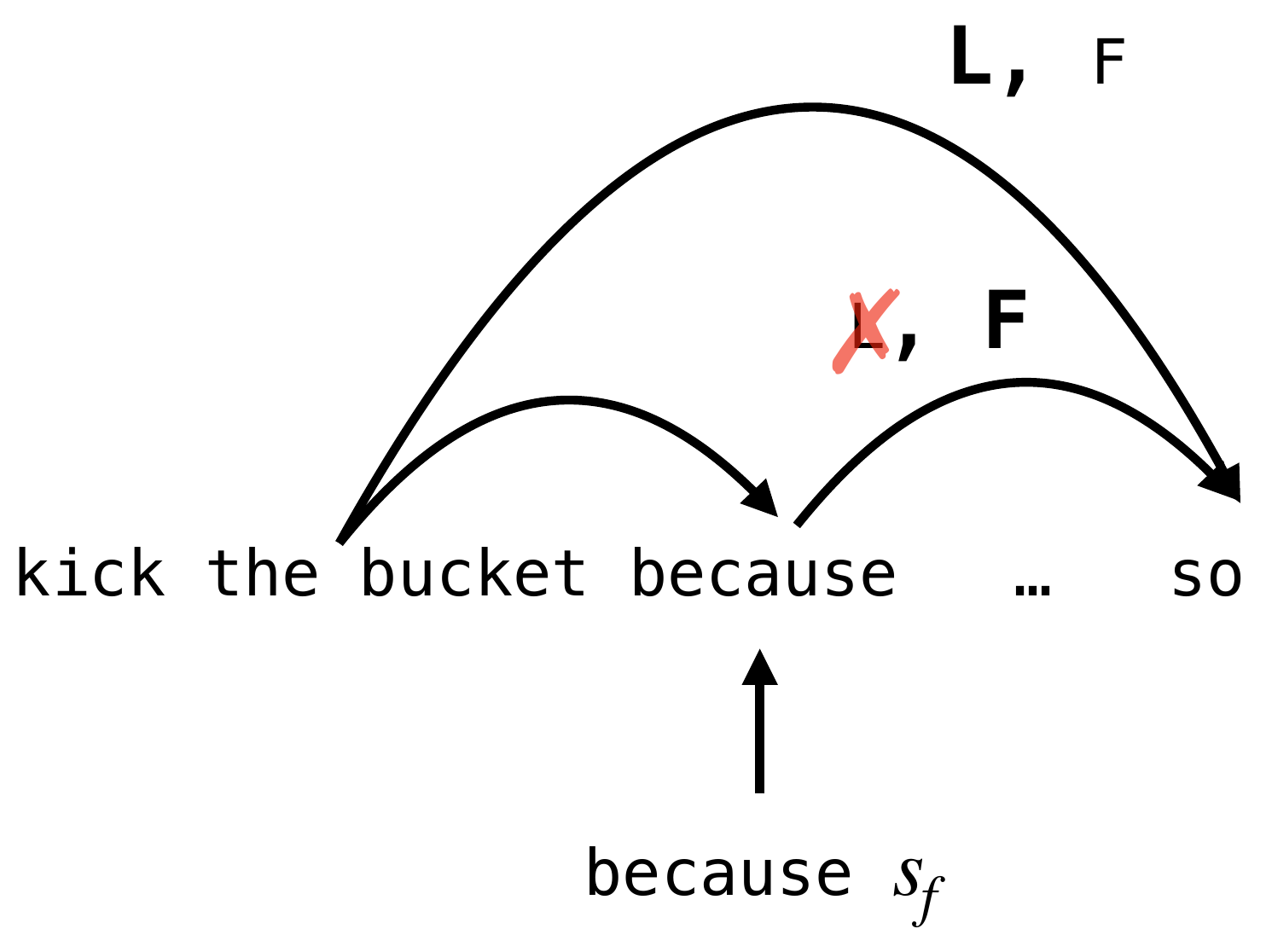}
    \caption{$s_a$ $\hookleftarrow$ $s_f$ at `because'}
    \label{fig:replace_desc1}
  \end{subfigure}
  \hspace{0.01\textwidth}  
    \begin{subfigure}[b]{0.2\textwidth}
    \centering
    \includegraphics[width=\textwidth]{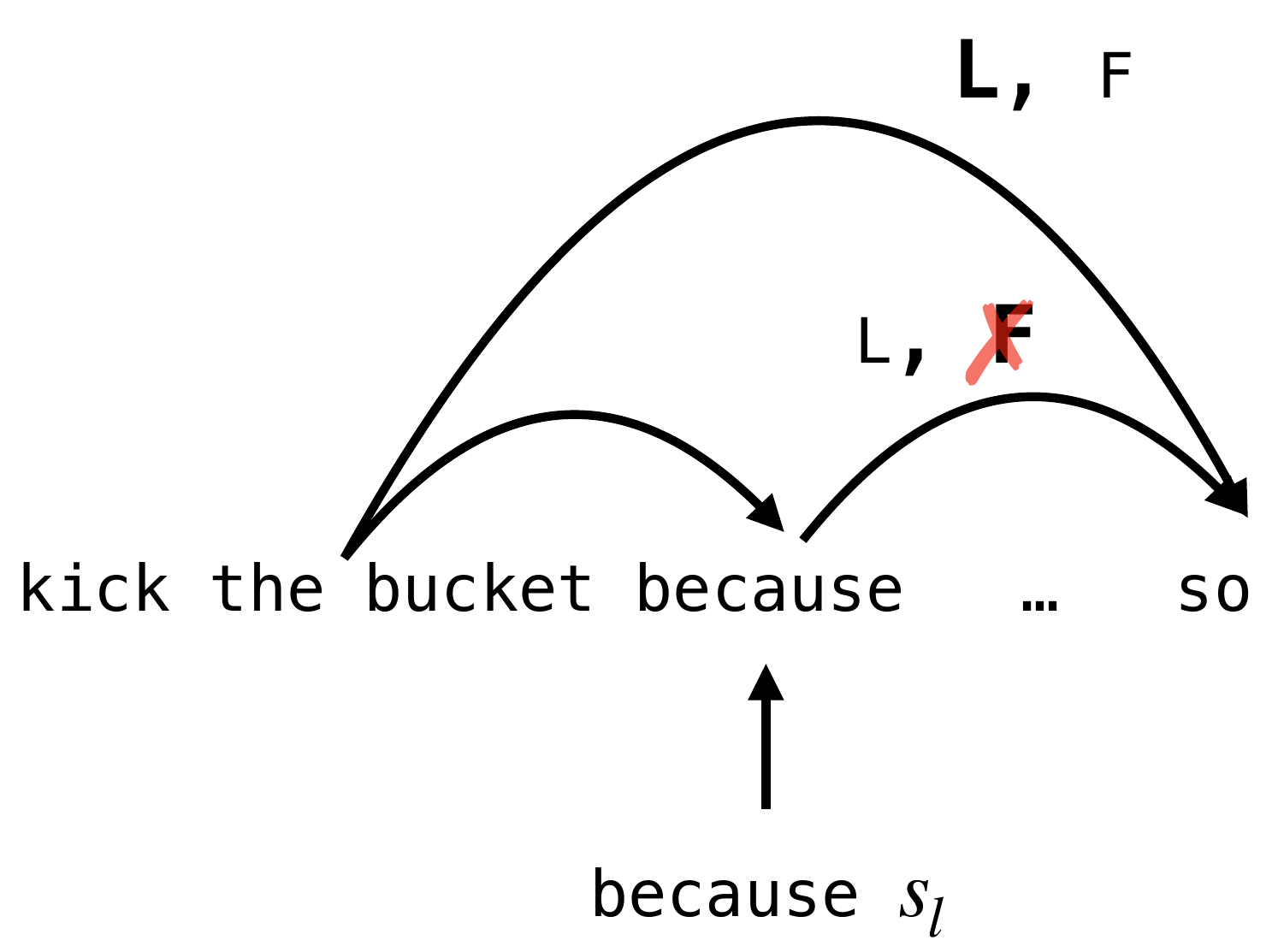}
    \caption{$s_a$ $\hookleftarrow$ $s_l$ at `because'}
    \label{fig:replace_desc2}
  \end{subfigure}
\begin{subfigure}[b]{0.21\textwidth}
    \centering
    \includegraphics[width=\textwidth]{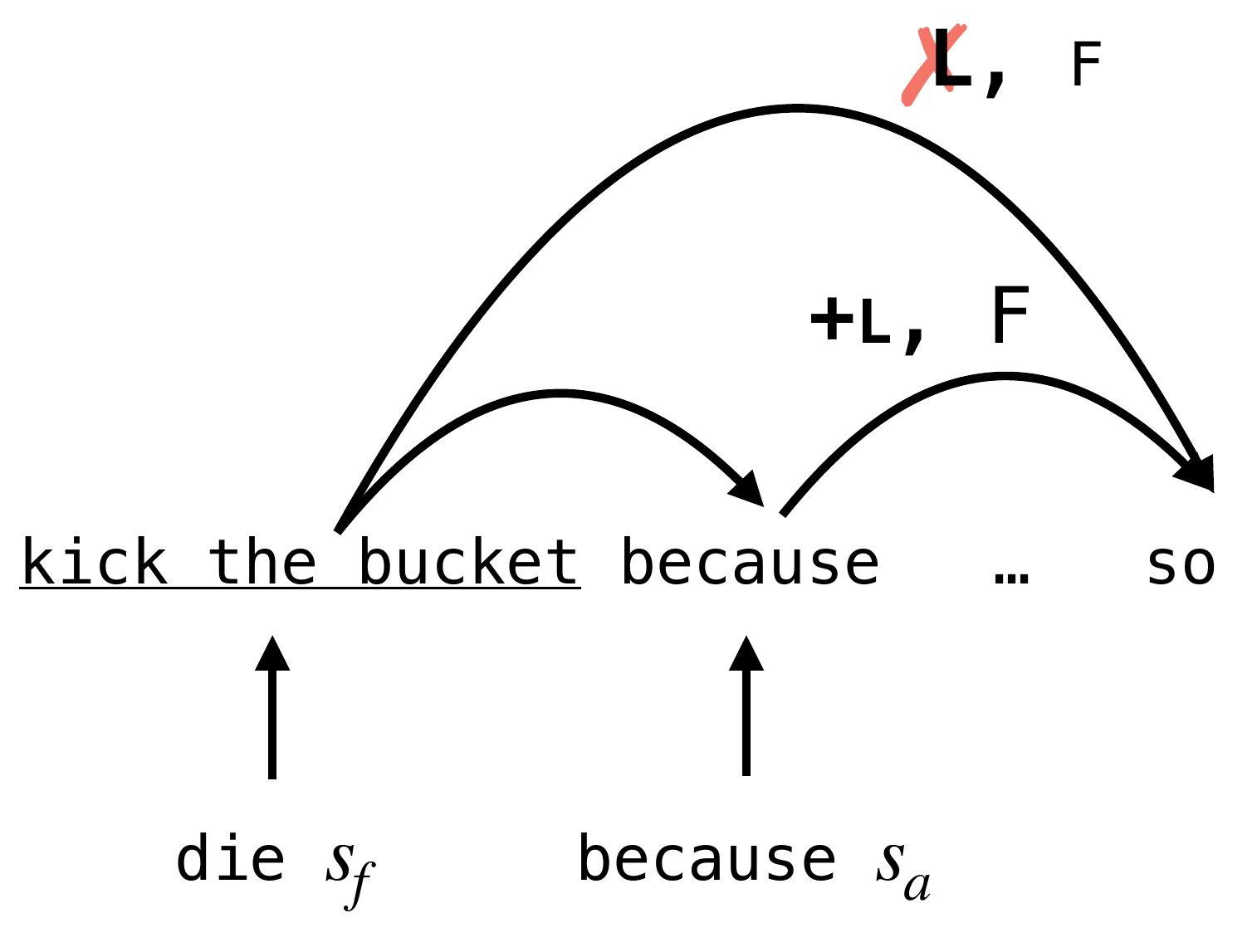}
    \caption{$s_a$ $\hookleftarrow$ $s_f$ at idiom span, \\ $s_a$  $\hookleftarrow$ $s_a$ at `because'}
    \label{fig:replace_desc3}
  \end{subfigure}
\begin{subfigure}[b]{0.26\textwidth}
    \centering
    \includegraphics[width=\textwidth]{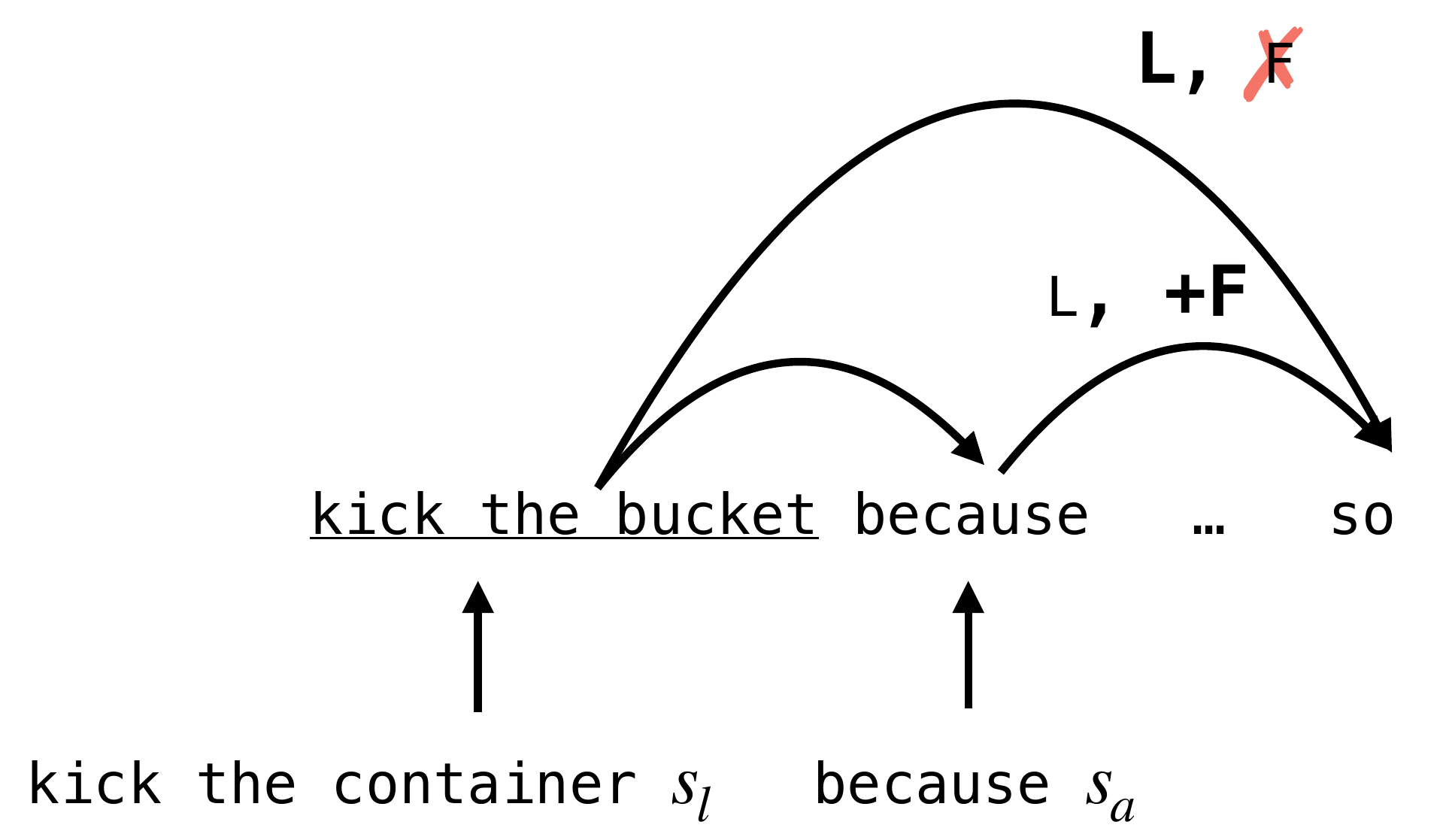}
    \caption{$s_a$ $\hookleftarrow$ $s_l$ at idiom span, \\ $s_a$  $\hookleftarrow$ $s_a$ at `because'}
    \label{fig:replace_desc4}
  \end{subfigure}
  \caption{Conceptual description of the activation patching experiments for tracing information flow (\textbf{L} = literal interpretation; \textbf{F} = figurative interpretation).}
  \label{fig:replace_desc}
\end{figure}

To measure the information flow along the intermediate pathway (through the subsequent token position \textit{`because'}), we patch the activation $s_a$  at the `because' token with $s_{f}$ or $s_{l}$ (Figures~\ref{fig:replace_desc1}, \ref{fig:replace_desc2}). On the other hand, patching the activation  $s_a$ at the idiom token span  with an alternative activation ($s_{f}$ or $s_{l}$) and then re-patching attention at the `because' token with $s_a$ allows us to isolate the information that flows via the direct route (Figures~\ref{fig:replace_desc3}, \ref{fig:replace_desc4}).


\begin{figure}[!h]
\centering
\begin{subfigure}[b]{0.48\linewidth}
  \centering
  \includegraphics[width=\linewidth]{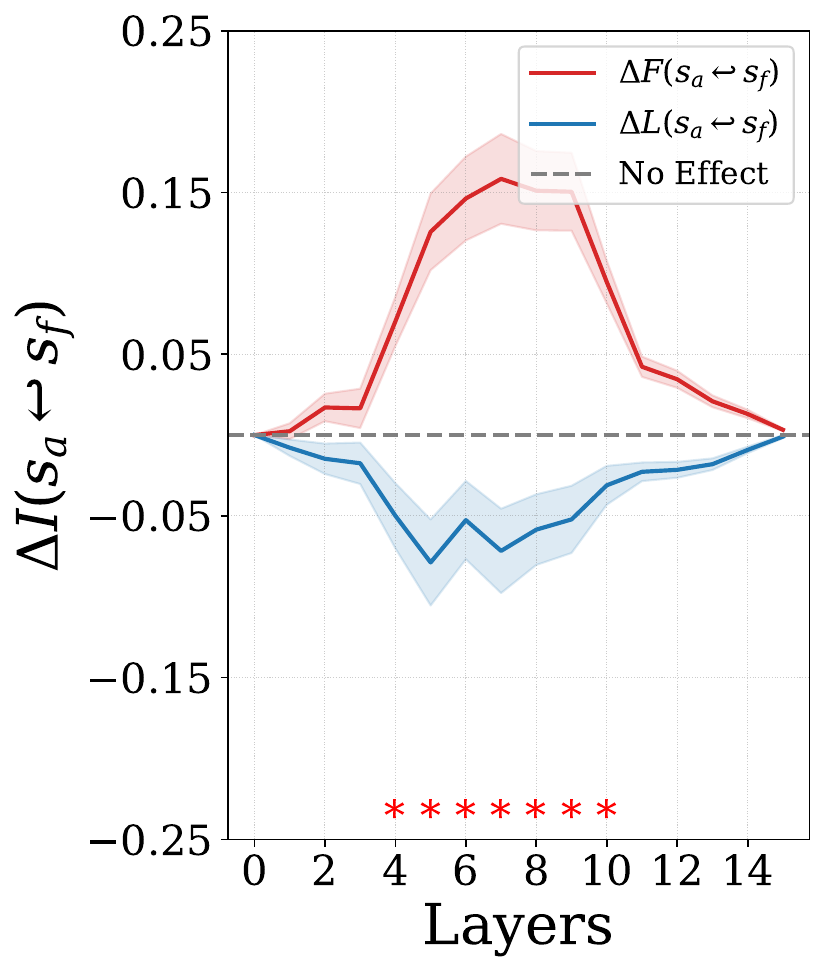}
  \caption{$s_a$ $\hookleftarrow$ $s_f$ at `because'}
  \label{fig:replace_idiom_fig}
\end{subfigure}\hfill
\begin{subfigure}[b]{0.48\linewidth}
  \centering
  \includegraphics[width=\linewidth]{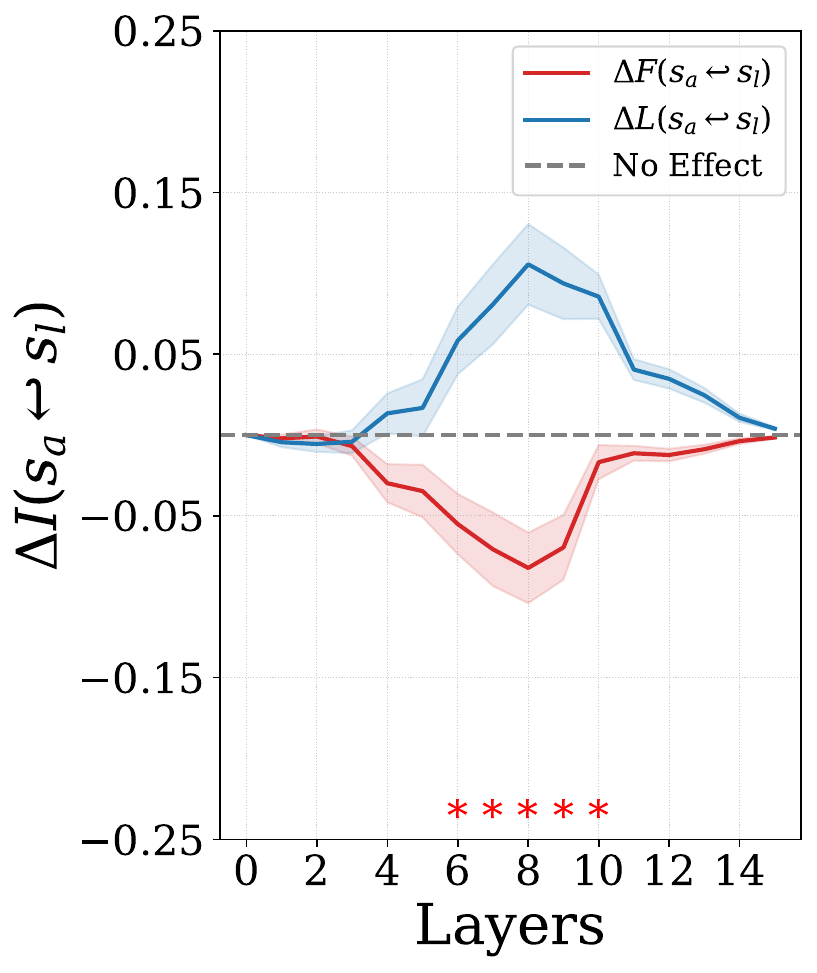}
  \caption{$s_a$ $\hookleftarrow$ $s_l$ at `because'}
  \label{fig:replace_idiom_lit}
\end{subfigure}
\begin{subfigure}[b]{0.48\linewidth}
  \centering
  \includegraphics[width=\linewidth]{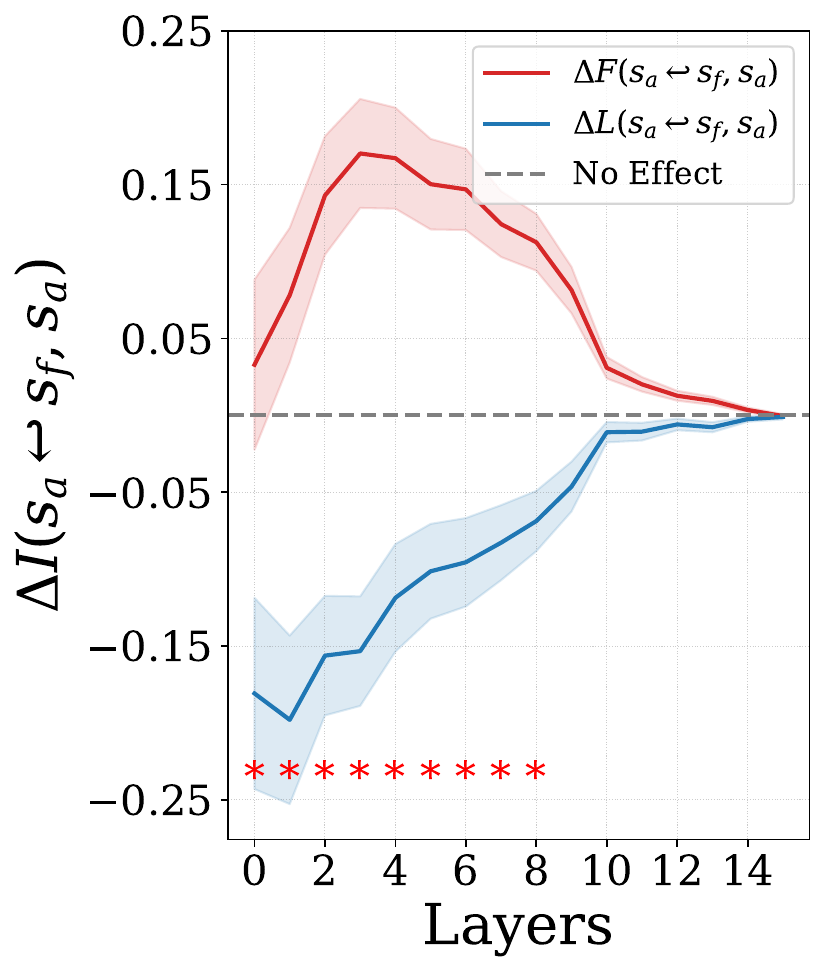}
  \caption{$s_a$ $\hookleftarrow$ $s_f$ at idiom span, \\ $s_a$  $\hookleftarrow$ $s_a$ at `because'}
  \label{fig:replace_fig_idiom_idiom_span}
\end{subfigure}\hfill
\begin{subfigure}[b]{0.48\linewidth}
  \centering
  \includegraphics[width=\linewidth]{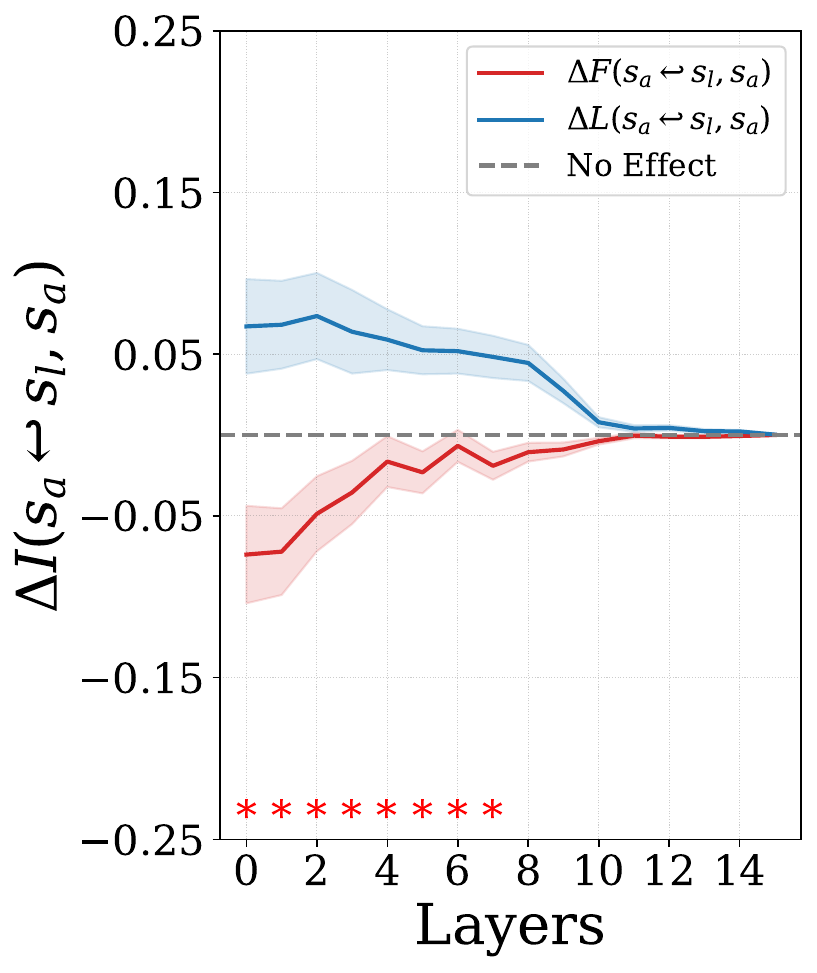}
  \caption{$s_a$ $\hookleftarrow$ $s_l$ at idiom span, \\ $s_a$  $\hookleftarrow$ $s_a$ at `because'}
  \label{fig:replace_lit_idiom_idiom_span}
\end{subfigure}

\caption{Layer-wise interpretation shift after patching in activations. The red asterisk (\textcolor{red}{*}) marks layers where the difference between \textcolor[RGB]{31,119,180}{$\Delta L$} and \textcolor[RGB]{214,39,40}{$\Delta F$} exceeds the average difference across layers (paired $t$-test, $p<0.05$).}
\label{fig:replace}
\end{figure}

\noindent\textbf{Results}\indent Figure~\ref{fig:replace_idiom_fig} shows that the patching as illustrated in Figure~\ref{fig:replace_desc1}, increases the $\Delta F(s_a \hookleftarrow s_f)$ in mid-layers 4-10 (max: $0.17$), but has a relatively small effect on $\Delta L(s_a \hookleftarrow s_f)$, which shows some decrease (min: $-0.09$). Patching in $s_l$ (as in Fig.~\ref{fig:replace_desc2}) removes the intermediate path figurative flow, which leads to an increase in the $\Delta L(s_a \hookleftarrow s_l)$ in mid-layers (6-10) as shown in Figure~\ref{fig:replace_idiom_lit} (max: $0.13$). At the same time, $\Delta F(s_a \hookleftarrow s_l)$ drops symmetrically in the same layers (min: $-0.10$). This indicates that the figurative interpretation is suppressed in favor of the literal interpretation. 

The asymmetry between increases and decreases show that a figurative injection doesn't lead to suppression of the literal interpretation. By contrast, a literal injection must first suppress the figurative interpretation in that subspace to have an effect, producing a more symmetric push–pull tradeoff. This indicates that for the mid-layers at the `because' token position, the model's residual stream is the predominant route for the figurative interpretation (\bluecircc{3-1} \textbf{figurative path}). 

Meanwhile, Figure~\ref{fig:ka_last} indicates that literal interpretation remains competitive with the figurative one. We assume that there's a different path that favors the semantically literal interpretation (\bluecircc{3-2} \textbf{compositional semantic direct path}) to convey its information to the final prediction, directly bypassing the intermediate path. 

To probe this, we replace the idiom sentence's idiom token span activation with the activation from a paraphrased sentence, and re-patch the `because' token with the idiom sentence's original activation (as shown in Figures~\ref{fig:replace_desc3}, \ref{fig:replace_desc4}). This preserves the intermediate pathway and isolates the contribution of the direct route. Figure~\ref{fig:replace_fig_idiom_idiom_span} shows that removing direct path's literal interpretation produces a large drop in $\Delta L$ (min: $-0.25$) and rise in $\Delta F$ (max: $0.20$). Conversely, removing the direct path's figurative interpretation yields only a modest drop in $\Delta F$ (min: $-0.10$) and a small increase in $\Delta L$ (max: $0.10$). These results indicate that the direct path predominantly conveys literal information. This pattern is consistent across different models (Appendix~\ref{sec:information_flow_others}).

\section{Localizing idiom disambiguation in context}
\label{sec:context}
To assess the role of context in idiom disambiguation, we're now looking at the setup with context. We retain an instance only when the model assigns higher cumulative probability to the context-consistent candidate than to the alternative (i.e., $F(s_a|FC) > F(s_a|LC)$, $L(s_a|LC) > L(s_a|FC)$).
\subsection{Probing sublayers through knockout with figurative vs. literal context}
We conduct a layer-wise knockout experiment to assess how MLP and MHSA components contribute to idiom processing when the same idiomatic expressions are presented in different contexts.

\begin{figure}[h]
\centering
    \begin{subfigure}[b]{0.23\textwidth}
    \centering
    \includegraphics[width=\textwidth]{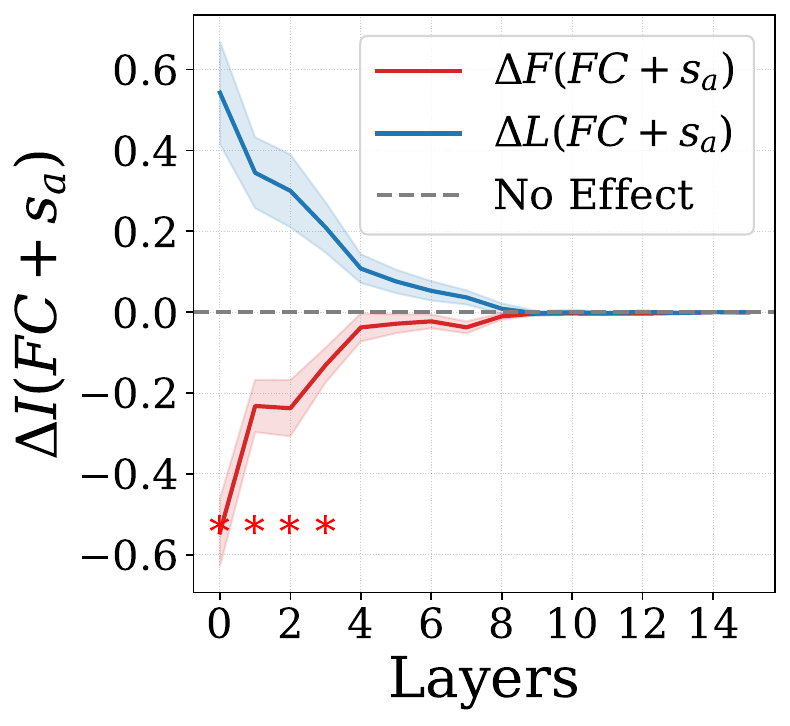}
    \caption{MLP FC}
    \label{fig:knockout_context_mlp_fig_llama1b}
  \end{subfigure}
    \begin{subfigure}[b]{0.23\textwidth}
    \centering
    \includegraphics[width=\textwidth]{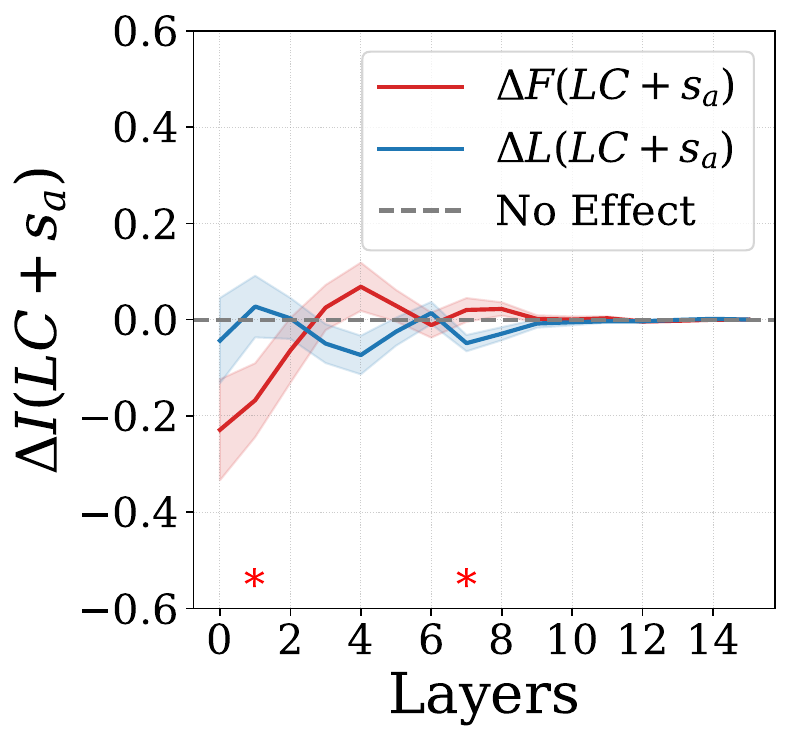}
    \caption{MLP LC}
    \label{fig:knockout_context_mlp_lit_llama1b}
  \end{subfigure}
    \begin{subfigure}[b]{0.23\textwidth}
    \centering
    \includegraphics[width=\textwidth]{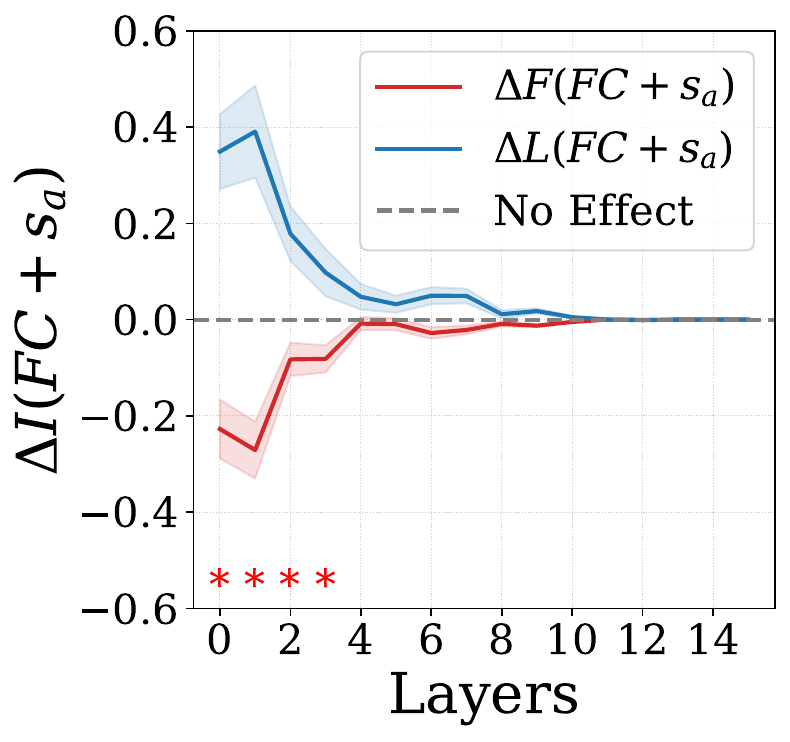}
    \caption{MHSA FC}
    \label{fig:knockout_context_attn_fig_llama1b}
  \end{subfigure}
      \begin{subfigure}[b]{0.23\textwidth}
    \centering
    \includegraphics[width=\textwidth]{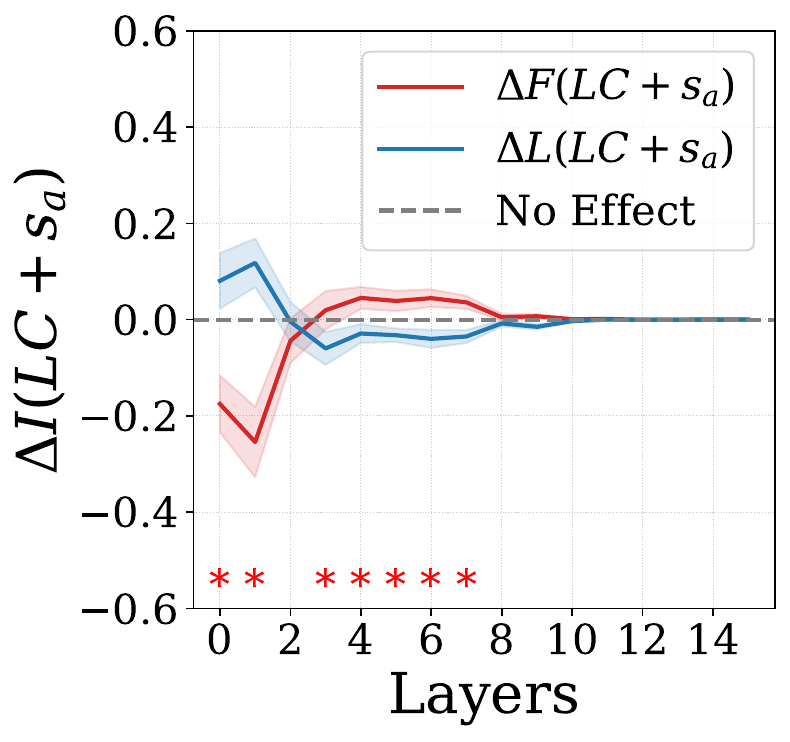}
    \caption{MHSA LC}
    \label{fig:knockout_context_attn_lit_llama1b}
  \end{subfigure}
  \caption{Sublayer-wise interpretation shift $\Delta I$ after knockout activations at idiom span with contexts. The red asterisk (\textcolor{red}{*}) marks layers where the difference between \textcolor[RGB]{31,119,180}{$\Delta L$} and \textcolor[RGB]{214,39,40}{$\Delta F$} exceeds the average difference across layers (paired $t$-test, $p<0.05$).}
  \label{fig:knockout_context}
\end{figure}

\noindent\textbf{Results}\indent Figure~\ref{fig:knockout_context_mlp_fig_llama1b} shows that for idioms in a figurative context, knocking out the MLP in the early layers (0–3) results in a significant drop in $\Delta F$ (min: $-0.62$) and increase in $\Delta L$ (max: $0.66$), consistent with the pattern observed for idioms without context (Figure~\ref{fig:mlp_knockout}). This indicates that these layers are critical for retrieving the figurative interpretation independently of context. This result is consistent across different models (Appendix~\ref{sec:knockout_context_others}). For idioms in a literal context (\ref{fig:knockout_context_mlp_lit_llama1b}), ablating the MLP layer 1 leads to a significant drop in $\Delta F$ (min: $-0.33$) and an increase in $\Delta L$ (max: $0.09$), indicating that the figurative interpretation is still being retrieved. The effect persists but is small compared with the figurative context condition, where early layers at the idiom token appear to encode context that suppresses figurative retrieval while biasing toward the literal interpretation.



A similar pattern is observed in the MHSA layers. As shown in Figure~\ref{fig:knockout_context_attn_fig_llama1b}, under figurative context, ablating early layers (0–3) produces a pronounced decrease in $\Delta F$ (min: $-0.33$) and an increase in $\Delta L$ (max: $0.48$). Under literal context, ablating the earliest layers (0–1) yields a drop in $\Delta F$ (min: $-0.32$), whereas only modest changes for $\Delta L$ (max: $0.16$) relative to the figurative context condition. That is, the literal context suppresses retrieval of the figurative interpretation from the very earliest layers. Moreover, ablating the layers (3–7) under literal context flips the pattern, where $\Delta F$ is increasing while $\Delta L$ is decreasing with a significant difference (Figure~\ref{fig:knockout_context_attn_lit_llama1b}). This indicates that, after the initial retrieval of figurative interpretation, the literal interpretation, which is disambiguated from context, is refined through mainly MHSA mediated integration with surrounding tokens.  

In sum, the results point to a two stage context disambiguation: (1) early layers retrieve figurative evidence while beginning to encode context, followed by (2) mid-layers, MHSA-driven contextual disambiguation that resolves conflicts in favor of the context-aligned interpretation.

\subsection{Probing MHSA contextual disambiguation with Query-preserved KV patching}\label{sec:kvq}
To examine in detail how MHSA mediates contextual disambiguation in early-mid layers, we perform Query-preserved Key-Value patching, in which MHSA's Key–Value (KV) activations are swapped while selectively preserving idiom tokens' query (Q) activations across layers, thereby redirecting the model's attention toward alternative contexts. That is, we switch idiom token span's attention from a figurative-context to the literal context, and vice versa. If the model's final prediction changes from figurative to literal (or vice versa) after patching, it demonstrates that the MHSA mechanism in those specific layers is responsible for using the context in the disambiguation process. 

\begin{figure}[h]
\centering
    \begin{subfigure}[b]{0.23\textwidth}
    \centering
    \includegraphics[width=\textwidth]{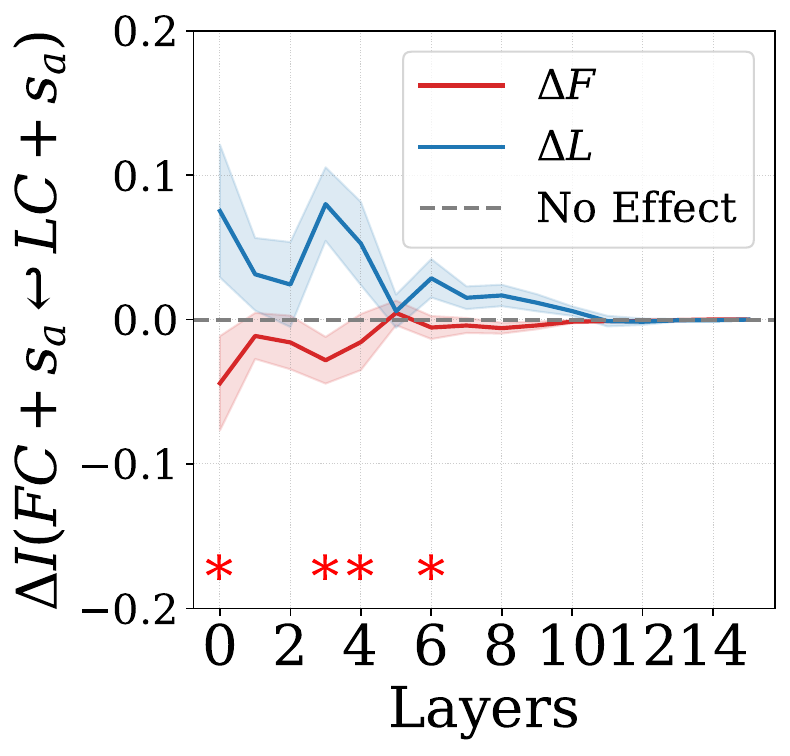}
    \caption{FC $+s_a$ $\hookleftarrow$ LC $+s_a$ \\ at idiom span}
    \label{fig:kvq_patching_FC_llama1b}
  \end{subfigure}
    \begin{subfigure}[b]{0.23\textwidth}
    \centering
    \includegraphics[width=\textwidth]{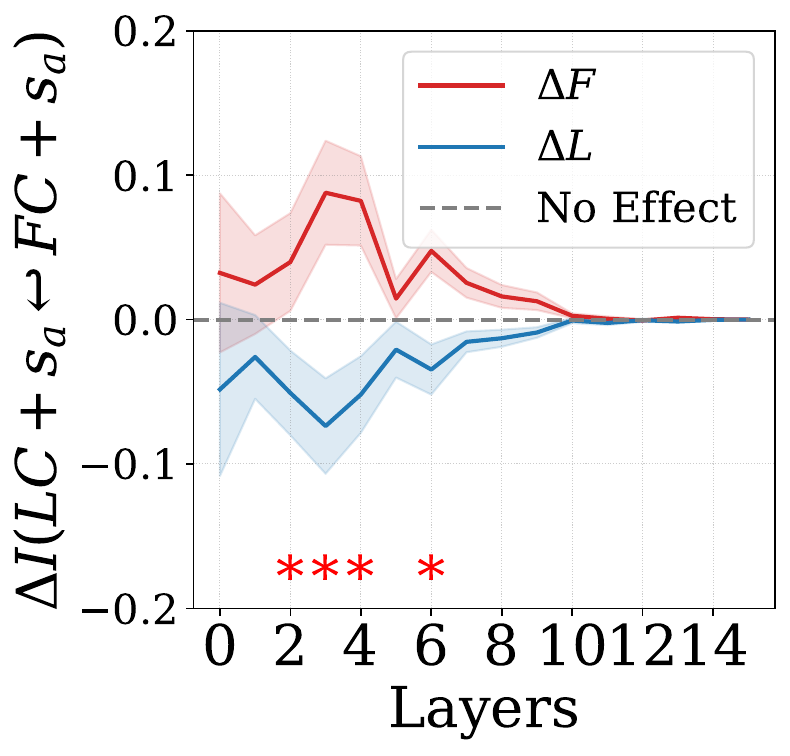}
    \caption{LC $+s_a$ $\hookleftarrow$ FC $+s_a$ \\ at idiom span}
    \label{fig:kvq_patching_LC_llama1b}
  \end{subfigure}
 
  \caption{Layer-wise interpretation shifts $\Delta I$ under Q-preserved KV patching. \textbf{Y-axis:} Mean \textcolor[RGB]{31,119,180}{$\Delta L(s_a)$} and \textcolor[RGB]{214,39,40}{$\Delta F(s_a)$} across idiom sentences. \textbf{X-axis:} Layers. \textbf{Gray dashed line:} $\Delta I=0$ (no effect). The red asterisk (\textcolor{red}{*}) marks layers where the difference between \textcolor[RGB]{31,119,180}{$\Delta L$} and \textcolor[RGB]{214,39,40}{$\Delta F$} exceeds the average difference across layers (paired $t$-test, $p<0.05$).}
  \label{fig:kvq_patching}
\end{figure}

\noindent\textbf{Results}\indent As shown in Figure~\ref{fig:kvq_patching}, patching in KVs with alternative context yields a shift in interpretation of idiom. Under \textit{FC}, effects appears already at layer 0, indicating immediate context sensitivity, with additional peaks at layers 2-3 and 6. Under \textit{LC}, the probability of the context matched interpretation drops across layers 2–6, while the alternative interpretation rises. Taken together, these results suggest that the idiom tokens exploit contextual evidence primarily in layers 2-6, after the figurative interpretation is first retrieved in layers 0-1, with layer 0 playing a crucial early role under \textit{FC}. Later layers tend to consolidate rather than redirect the interpretation.

\section{Conclusion}
We present a multi-stage mechanism of idiom comprehension in causal transformers, by identifying distinct components and flows for figurative and literal interpretations. Causal interventions reveal (i) early retrieval of the idiomatic interpretation, (ii) immediate use of preceding context, with later refinement when it conflicts with the retrieved interpretation, (iii) a selective pathway that carries the figurative interpretation through intermediate layers, and a bypass route that favorably delivers the literal interpretation directly to the output. 

Considering idioms as multiword expressions, our results suggest that early layers (especially MLPs) perform a ``detokenization'' step~\cite{gurneefinding, elhage2021mathematical} that compresses the idiom into a unified figurative representation, while later pathways can ``retokenize'' it into a competing literal interpretation. To sum up, these findings link representational dynamics to contextual disambiguation, yielding a concise mechanistic interpretation of idiom processing.

\section*{Limitations}

While causal tracing methods have been widely used in recent work~\cite{geva2023dissecting, dar2022analyzing, meng2022locating, heimersheim2404use, nanda2023progress}, they only approximate the actual information stored in activations. Moreover, knocking out or replacing the activations can lead the model to out-of-distribution behaviors and cast doubt on the robustness of any interpretability claim. 

Moreover, we binarize idiom interpretation (figurative vs. literal), ignoring polysemy among figurative senses (e.g., `break the ice' can be interpreted into initiate talk, ease tension, etc.), which can flatten nuance and bias evaluation. As a future direction, we can use soft labels for multiple figurative sense and distributional metrics.  


\section*{Ethical Considerations}
One of the intended uses of the Llama-3.3-70B-instruct model is content generation, which aligns with our use of the model for generating the candidates for next token prediction task~\cite{grattafiori2024llama}. The candidates generated by the language model may include biased or sensitive attributes (e.g., race, minority status), which reflects stereotypes that the language model already has (See the Appendix Table~\ref{tab:biased_data_ex} for examples).

\section*{Acknowledgements}
This project has received funding from the European Research Council (ERC) under the European Union's
Horizon 2020 research and innovation programme (ERC Starting Grant “Individualized Interaction in Discourse”,
grant agreement No. 948878). Funded in part by the Deutsche Forschungsgemeinschaft (DFG, German Research Foundation) – Project-ID 232722074 – SFB 1102. We gratefully acknowledge the stimulating research environment of the GRK 2853/1 ``Neuroexplicit Models of Language, Vision, and Action'', funded by the Deutsche Forschungsgemeinschaft (DFG, German Research Foundation) under project number 471607914.

\begin{figure}[h] 
\centering
\includegraphics[width=0.75\columnwidth]{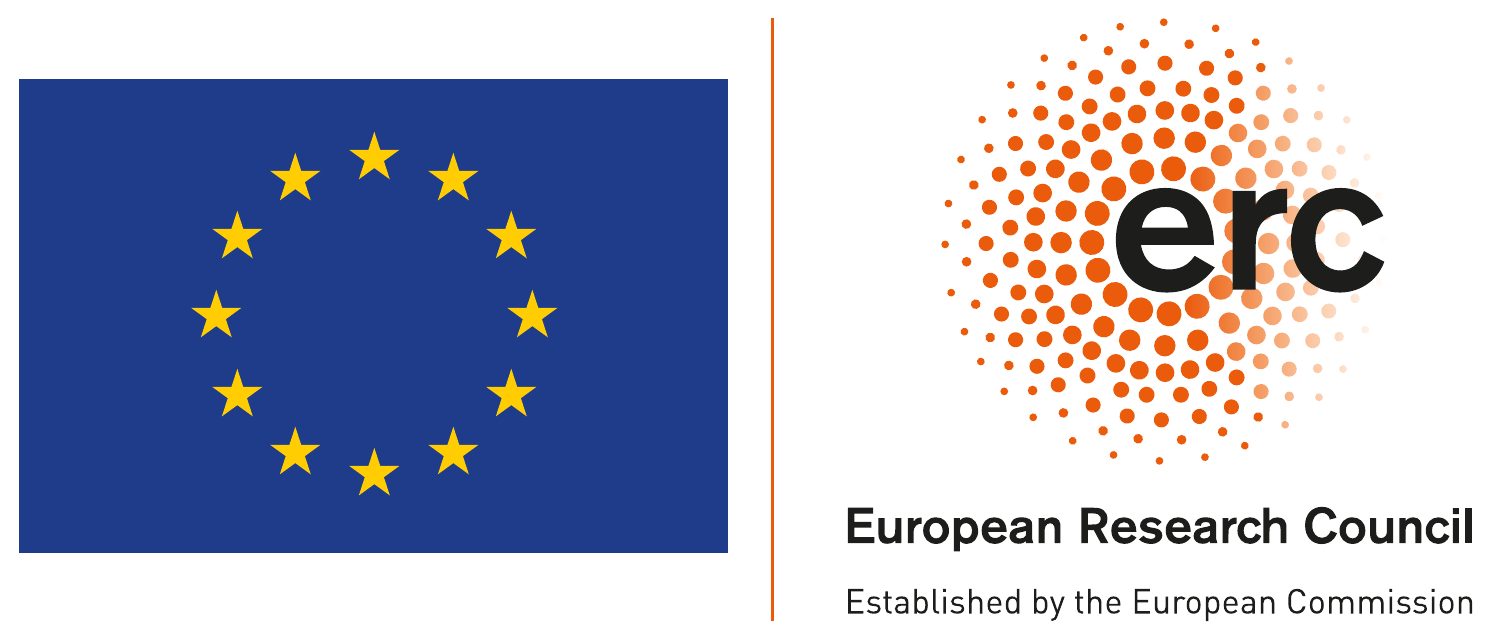}
\end{figure}

\bibliography{custom}
\bibliographystyle{acl_natbib}

\appendix

\appendix

\section{Identified top-20 idiomatic heads}
Table~\ref{tab:topk} lists the attention heads most influential in retrieving figurative meanings while suppressing literal interpretations and randomly selected heads.



\begin{table}[h]
\small
\centering
\begin{tabularx}{\columnwidth}{@{}l|X@{}}
\toprule
\textbf{Type} & \textbf{top-$20$ attention heads (layer, head)} \\
\midrule
Idiomatic & (0, 4), (1, 5), (0, 30), (0, 19), (0, 8), (9, 30), (2, 2), (1, 18), (0, 21), (2, 8), (1, 9), (4, 2), (1, 24), (1, 13), (3, 18), (0, 0), (3, 24), (0, 26), (1, 27), (0, 28)
\\ \hline
Random & (9, 20), (10, 30), (2, 11), (2, 23), (1, 1), (9, 23), (13, 24), (8, 31), (3, 8), (13, 9), (10, 1), (5, 15), (7, 14), (7, 24), (4, 5), (14, 2), (9, 17), (4, 0), (14, 19), (6, 9) \\
\bottomrule
\end{tabularx}
\caption{Top-$20$ attention head sets.}
\label{tab:topk}
\end{table}

\section{Data generation pipeline}

\begin{figure}[!h]
\centering
\includegraphics[width=0.47\textwidth]{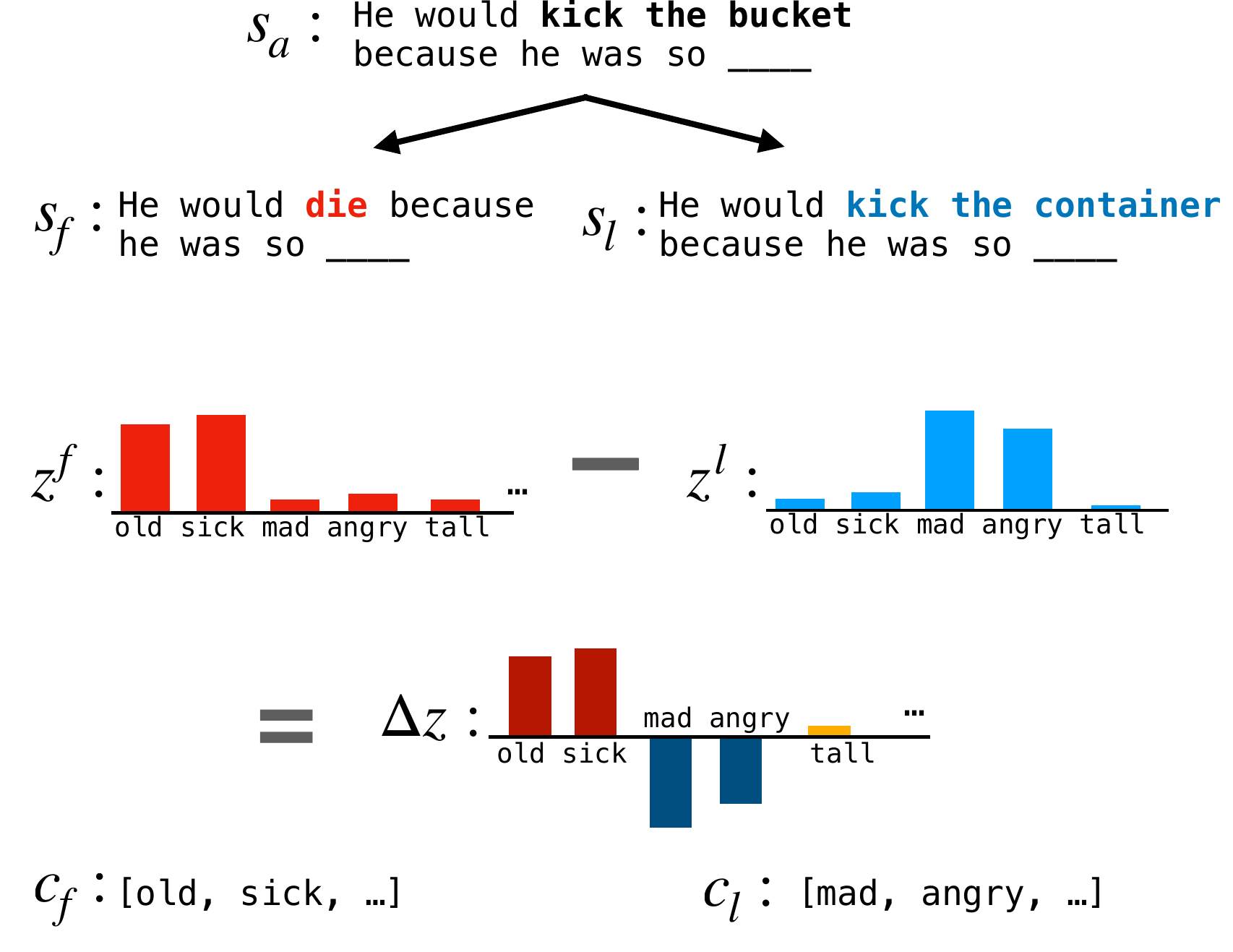}
  \caption{The pipeline of data generation.}
  \label{fig:datagen}
\end{figure}

We illustrate the data generation pipeline in Figure~\ref{fig:datagen}.

\section{Layer-wise context disambiguation via logit lens}
\label{sec:logit_lens_context}
If the MHSA layers 3-7 play a causal role in context disambiguation, then we would expect that the representation at the final idiom token is fully disambiguated towards either the literal or the figurative meaning, depending on the earlier context. 
We employ the logit lens method~\cite{nostalgebraist2020logitlens} to measure the log-probability assigned to figurative and literal interpretations in the figurative ($FC$) vs.~literal context ($LC$) conditions. At each layer, the hidden representations are projected into the output vocabulary space, yielding candidate probabilities. We then measure the difference in log-probability mass between figurative and literal candidates across layers, separately isolating contributions from the MHSA and MLP components.

\begin{figure}[!h]
\centering
    \begin{subfigure}[b]{0.2\textwidth}
    \centering
    \includegraphics[width=\textwidth]{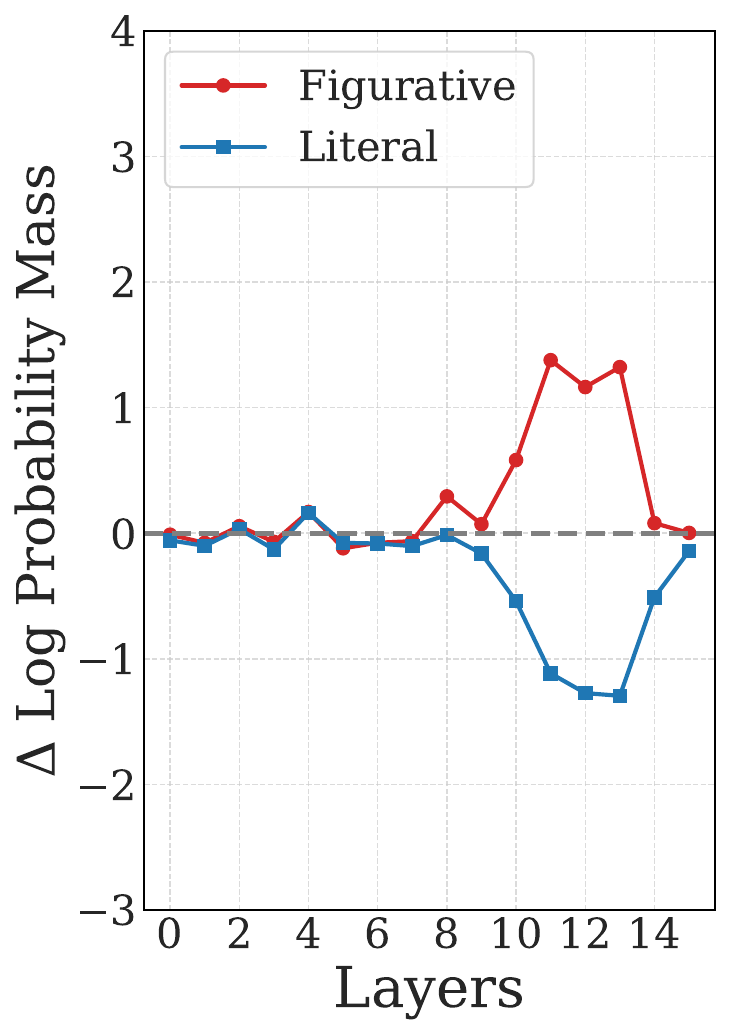}
    \caption{MLP FC$-$LC}
    \label{fig:logitlens_mlp}
  \end{subfigure}
    \begin{subfigure}[b]{0.2\textwidth}
    \centering
    \includegraphics[width=\textwidth]{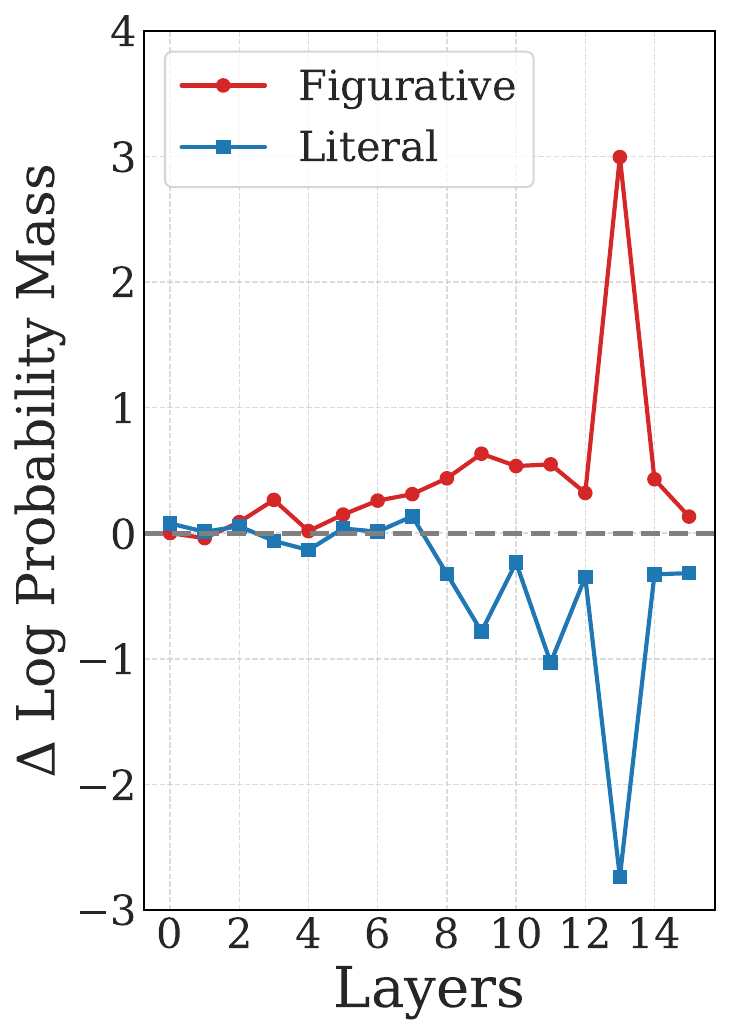}
    \caption{MHSA FC$-$LC}
    \label{fig:logitlens_attn}
  \end{subfigure}
  \caption{Layerwise differences in log-probability between figurative and literal candidates predictions under figurative (FC) vs.~literal (LC) contexts, measured with the logit lens.}
  \label{fig:logitlens_difference_FC-LC}
\end{figure}


\noindent\textbf{Results}\indent 
Figure~\ref{fig:logitlens_difference_FC-LC} shows a clear disambiguation effect for MHSA starting around layer 8. Once MHSA starts modulating activations in a context-dependent manner, it writes this information into the residual stream, which is subsequently processed by the MLP layers. In the early to middle layers (0–9) of the MLP, the difference in log probabilities between figurative and literal candidate predictions under figurative vs.~literal contexts remains minimal. However, beginning at layer 10, MLP activations begin to diverge: idioms presented in figurative contexts are increasingly encoded with figurative interpretations, whereas idioms in literal contexts are increasingly encoded with literal interpretations.


\section{Knockout for crucial role of early layers with other models}
\label{sec:knockout_others}
\begin{figure*}[t]
\centering
{\captionsetup{type=figure}
 \caption*{\textbf{Qwen2.5-0.5B}}}
\begin{subfigure}[b]{0.23\textwidth}
  \centering
  \includegraphics[width=\textwidth]{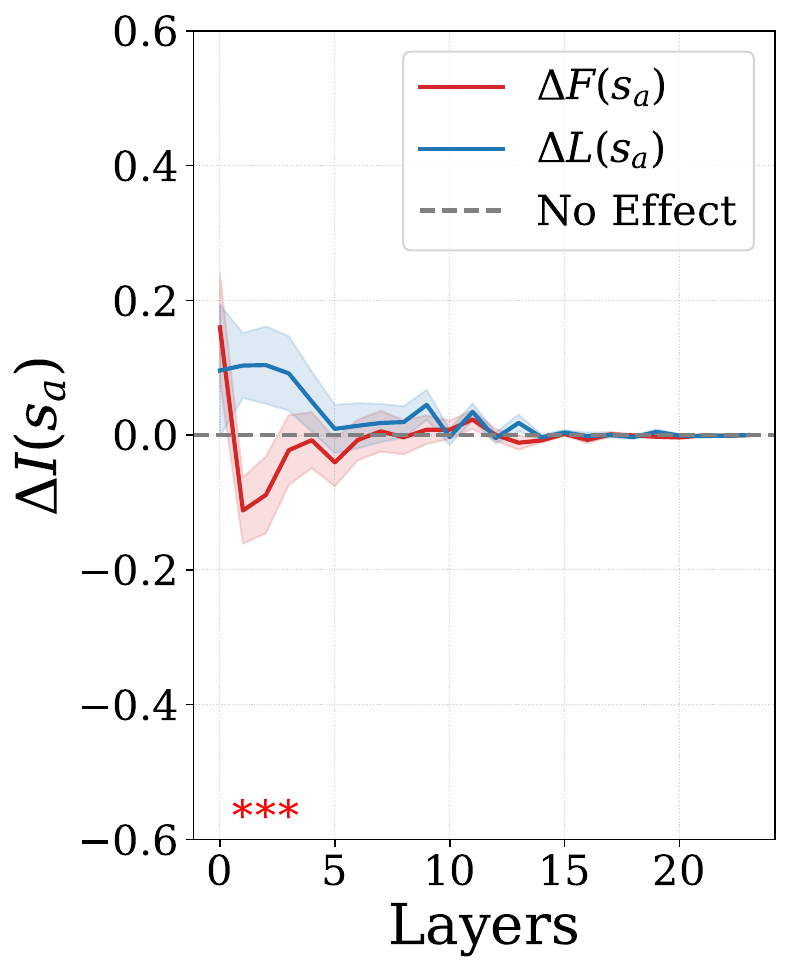}
  \caption{MLP}\label{fig:qwen_mlp_knocout}
\end{subfigure}\hfill
\begin{subfigure}[b]{0.23\textwidth}
  \centering
  \includegraphics[width=\textwidth]{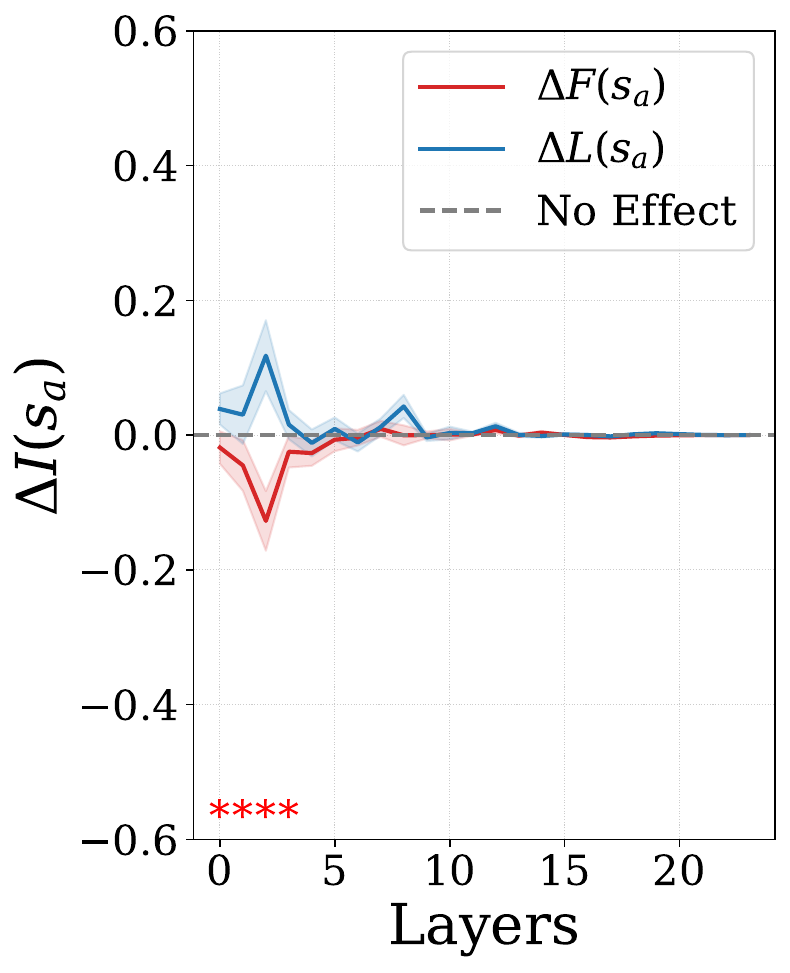}
  \caption{MHSA}\label{fig:qwen_attn_knockout}
\end{subfigure}\hfill
\begin{subfigure}[b]{0.23\textwidth}
  \centering
  \includegraphics[width=\textwidth]{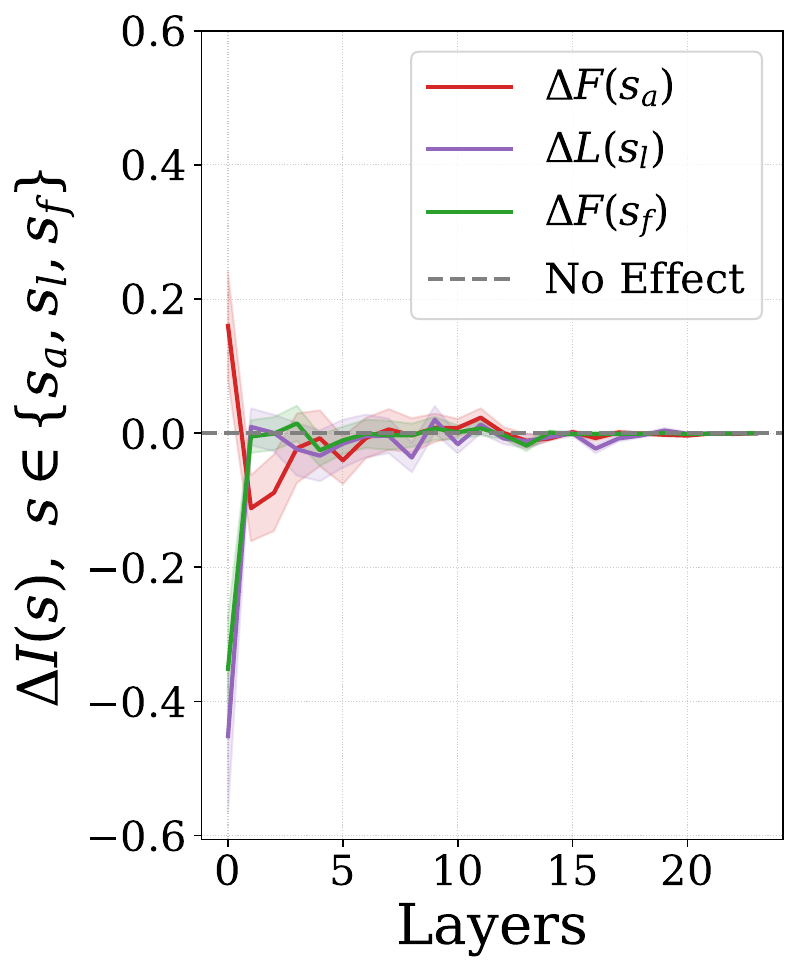}
  \caption{MLP}\label{fig:qwen_mlp_knockout_all}
\end{subfigure}\hfill
\begin{subfigure}[b]{0.23\textwidth}
  \centering
  \includegraphics[width=\textwidth]{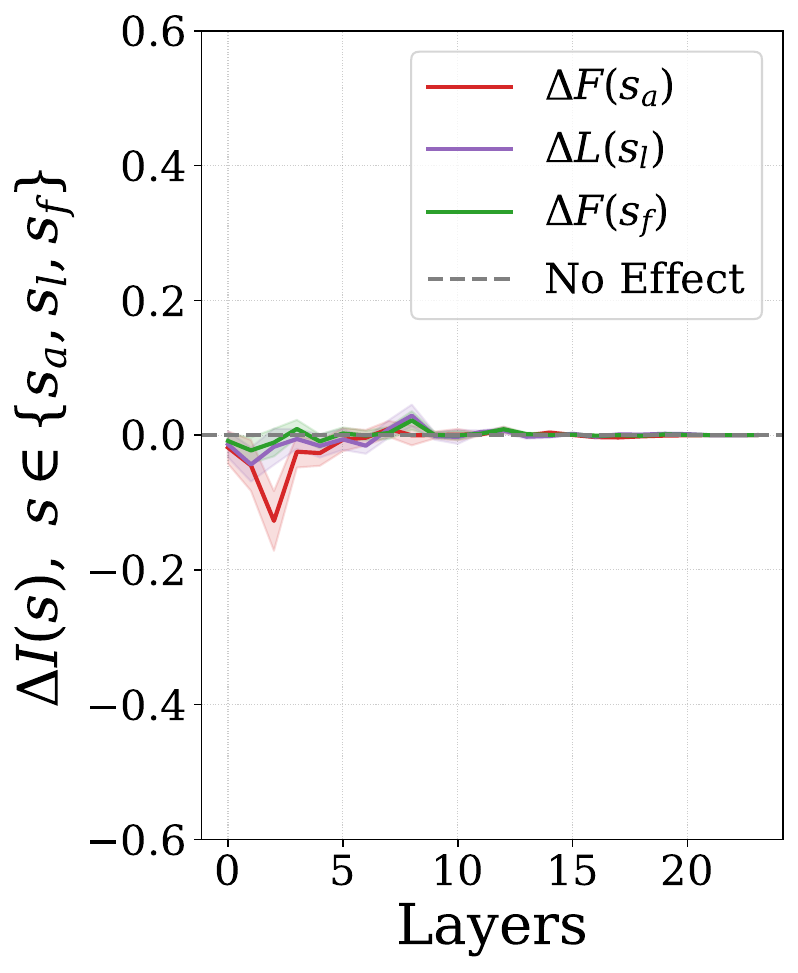}
  \caption{MHSA}\label{fig:qwen_attn_knockout_all}
\end{subfigure}
{\captionsetup{type=figure}
 \caption*{\textbf{Llama3.1-8B}}}
\begin{subfigure}[b]{0.23\textwidth}
  \centering
  \includegraphics[width=\textwidth]{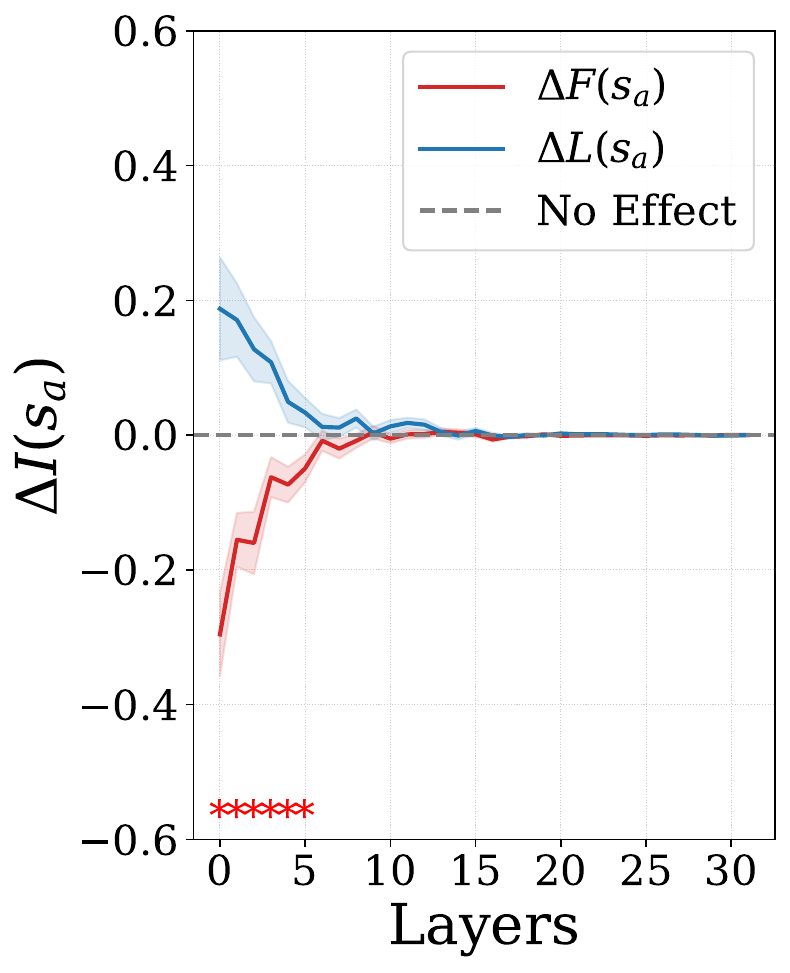}
  \caption{MLP}\label{fig:llama8b_mlp_knocout}
\end{subfigure}\hfill
\begin{subfigure}[b]{0.23\textwidth}
  \centering
  \includegraphics[width=\textwidth]{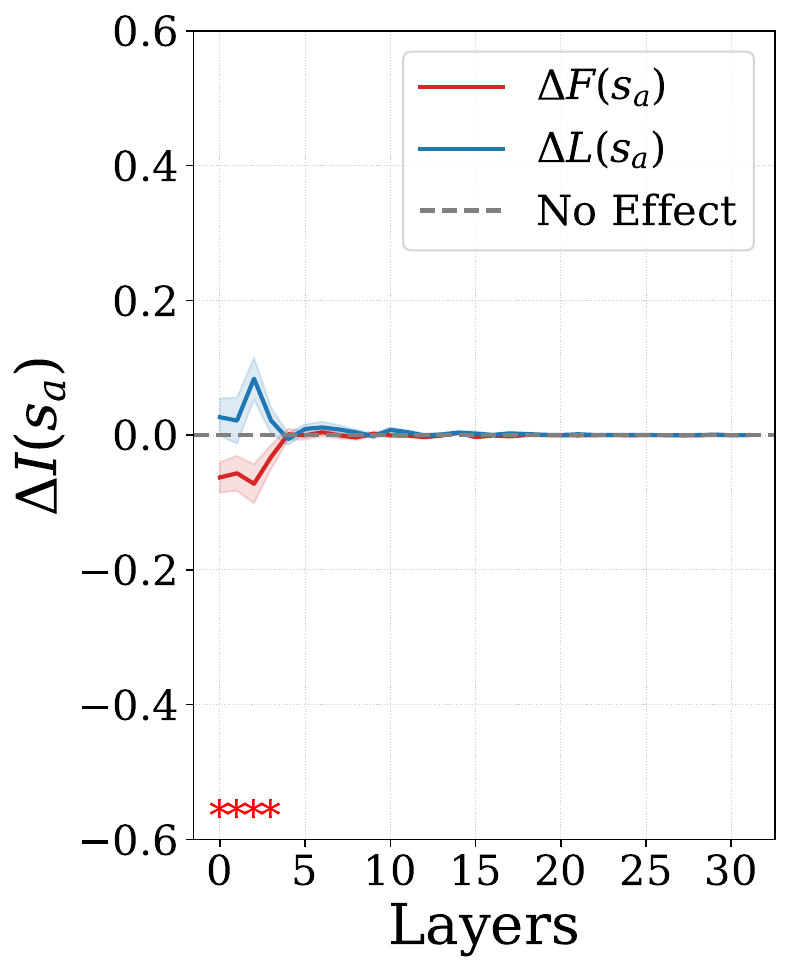}
  \caption{MHSA}\label{fig:llama8b_attn_knockout}
\end{subfigure}\hfill
\begin{subfigure}[b]{0.23\textwidth}
  \centering
  \includegraphics[width=\textwidth]{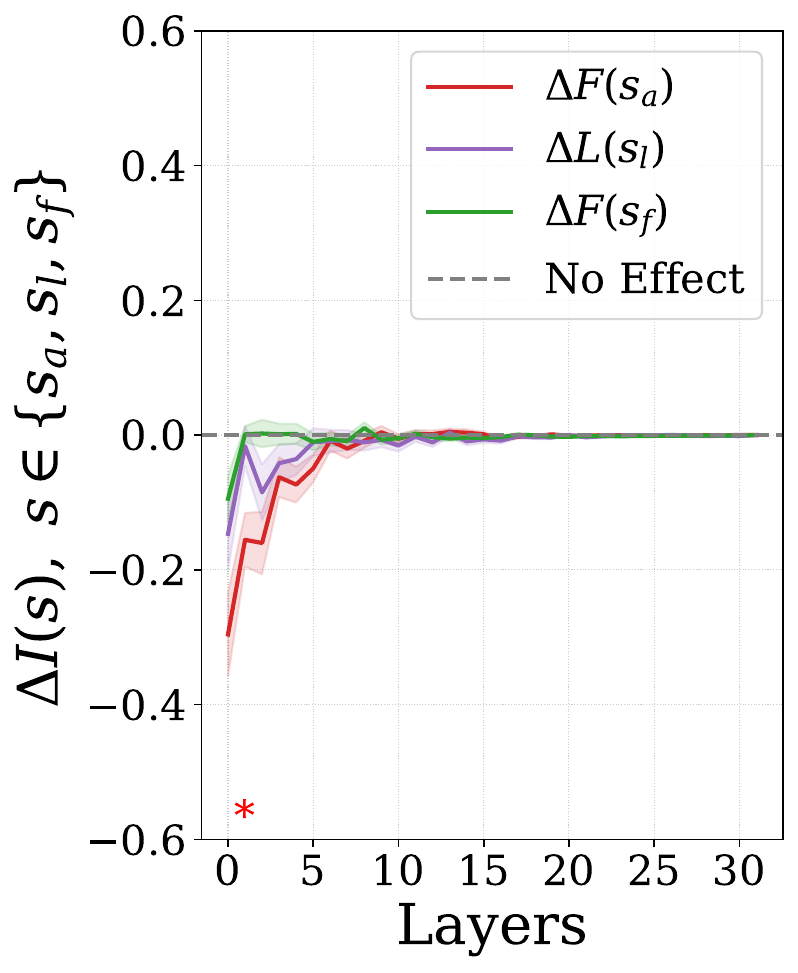}
  \caption{MLP}\label{fig:llama8b_mlp_knockout_all}
\end{subfigure}\hfill
\begin{subfigure}[b]{0.23\textwidth}
  \centering
  \includegraphics[width=\textwidth]{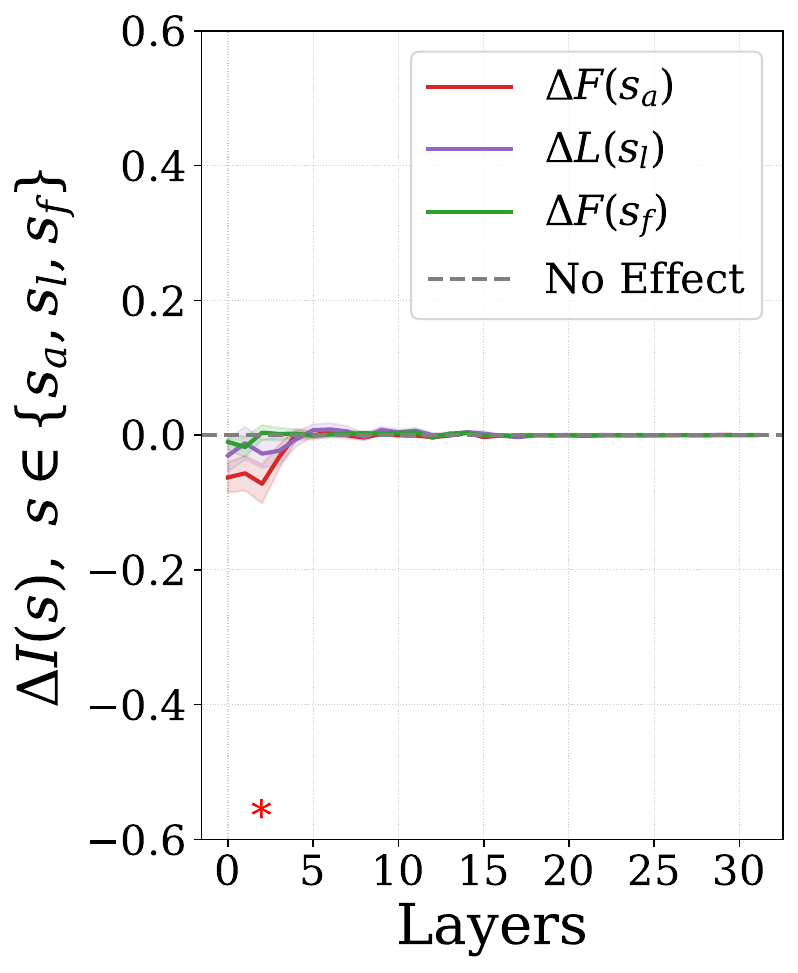}
  \caption{MHSA}\label{fig:llama8b_attn_knockout_all}
\end{subfigure}
{\captionsetup{type=figure}
 \caption*{\textbf{Qwen2.5-7B}}}
\begin{subfigure}[b]{0.23\textwidth}
  \centering
  \includegraphics[width=\textwidth]{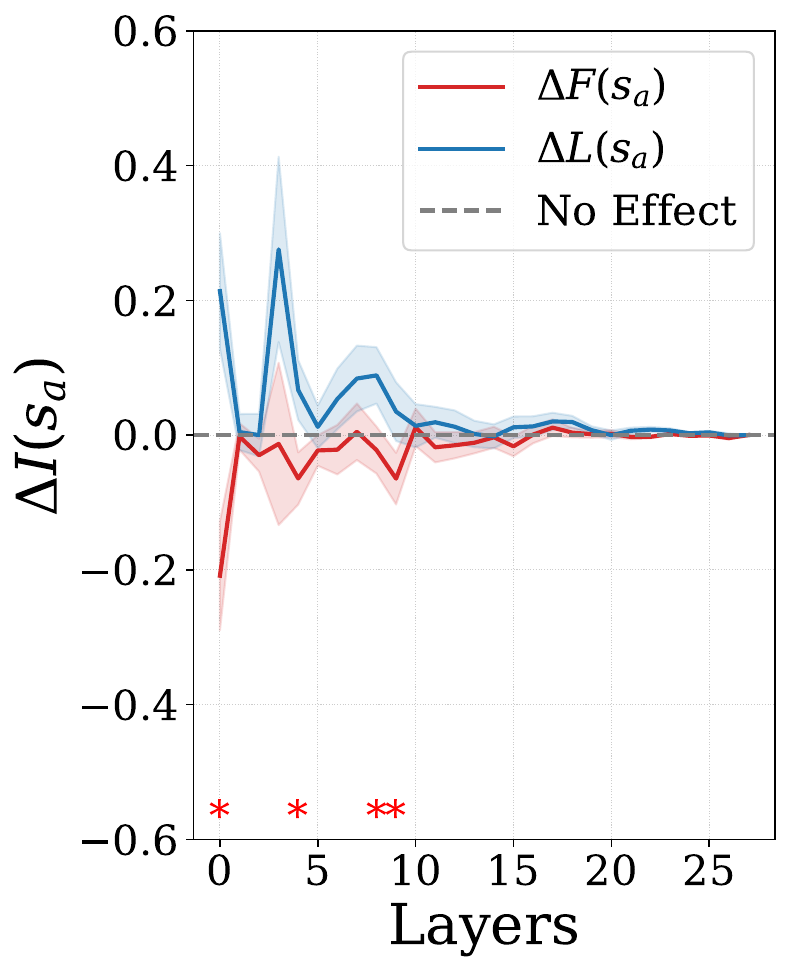}
  \caption{MLP}\label{fig:qwen7b_mlp_knocout}
\end{subfigure}\hfill
\begin{subfigure}[b]{0.23\textwidth}
  \centering
  \includegraphics[width=\textwidth]{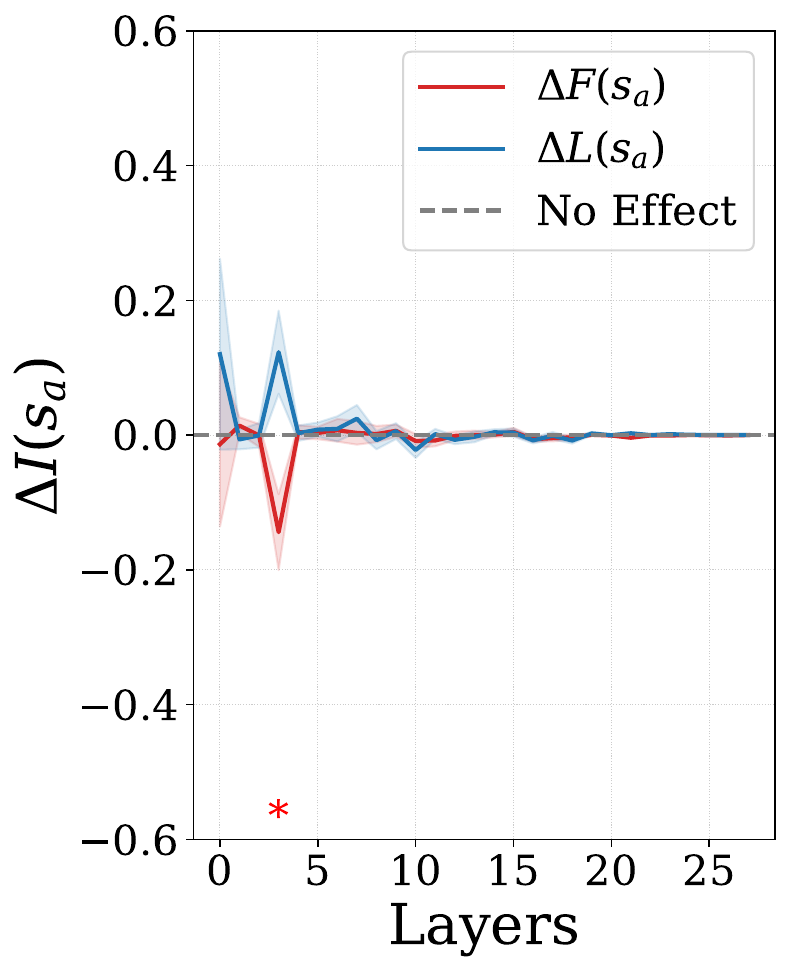}
  \caption{MHSA}\label{fig:qwen7b_attn_knockout}
\end{subfigure}\hfill
\begin{subfigure}[b]{0.23\textwidth}
  \centering
  \includegraphics[width=\textwidth]{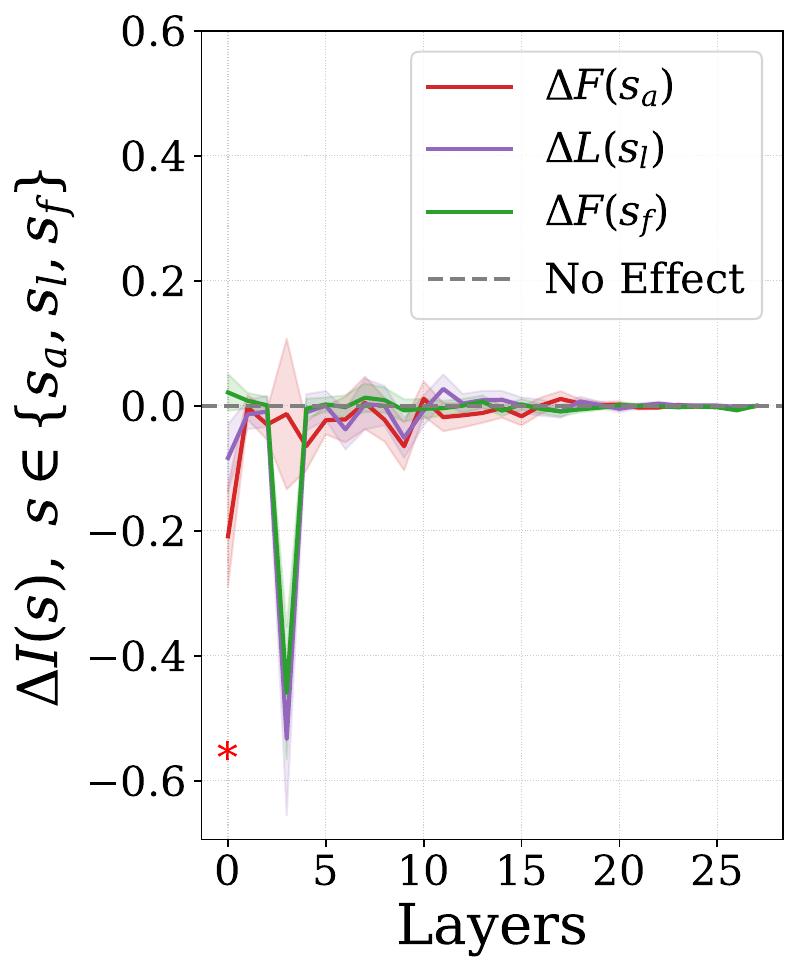}
  \caption{MLP}\label{fig:qwen7b_mlp_knockout_all}
\end{subfigure}\hfill
\begin{subfigure}[b]{0.23\textwidth}
  \centering
  \includegraphics[width=\textwidth]{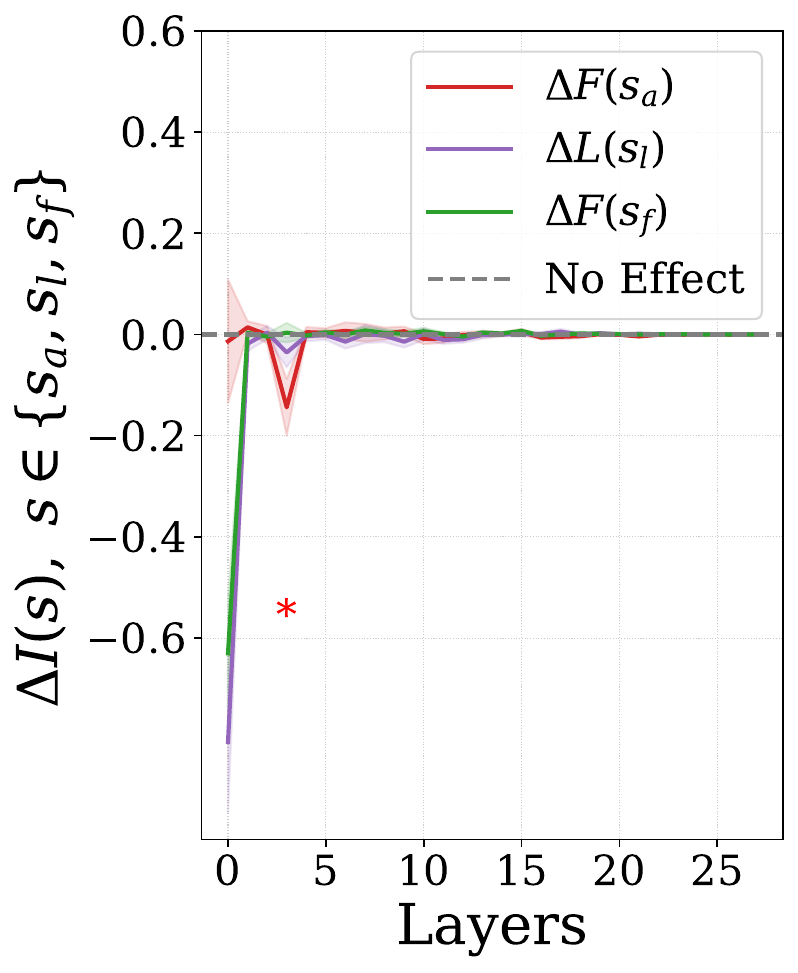}
  \caption{MHSA}\label{fig:qwen7b_attn_knockout_all}
\end{subfigure}
\caption{Sublayer-wise interpretation shift $\Delta I(s)$ after ablating activations at idiom span, for sentences $s \in \{s_a, s_f, s_l\}$. \textbf{Y-axis:} Mean values of \textcolor[RGB]{31,119,180}{$\Delta L(s_a)$}, \textcolor[RGB]{214, 39, 40}{$\Delta F(s_a)$}, \textcolor[RGB]{148,103,189}{$\Delta L(s_l)$}, \textcolor[RGB]{44,160,44}{$\Delta F(s_f)$} with 95\% confidence intervals. \textbf{X-axis:} Layers. \textbf{Gray dashed line:} $\Delta I = 0$ (no effect). \textbf{Red asterisk (\textcolor{red}{*}):} Significant difference between \textcolor[RGB]{214, 39, 40}{$\Delta F(s_a)$} and the others (paired $t$-test, $p<0.05$). The difference at \textcolor{red}{*} marked layer is larger than the average difference across all layers.}
\label{fig:knockout_all_models}
\end{figure*}

Results across different models in Figure~\ref{fig:knockout_all_models} consistently show that early layers retrieve idioms' figurative interpretations while suppressing literal counterparts, unlike unambiguous sentences.

\section{Activation patching for information flow with other models}

\label{sec:information_flow_others}
\begin{figure*}[!h]
\centering
{\captionsetup{type=figure}
 \caption*{\textbf{Qwen2.5-0.5B}}}
\begin{subfigure}[b]{0.23\linewidth}
  \centering
  \includegraphics[width=\linewidth]{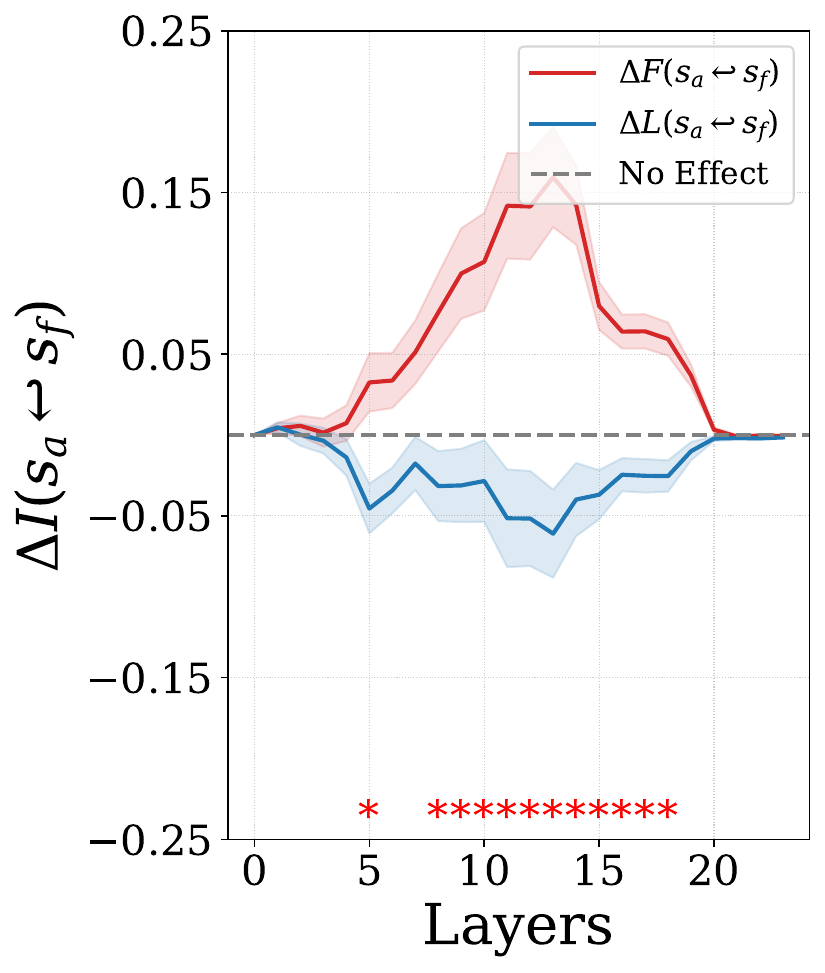}
  \caption{Idiom $\hookleftarrow$ Figurative}
  \label{fig:replace_idiom_fig_qwen0.5b}
\end{subfigure}
\begin{subfigure}[b]{0.23\linewidth}
  \centering
  \includegraphics[width=\linewidth]{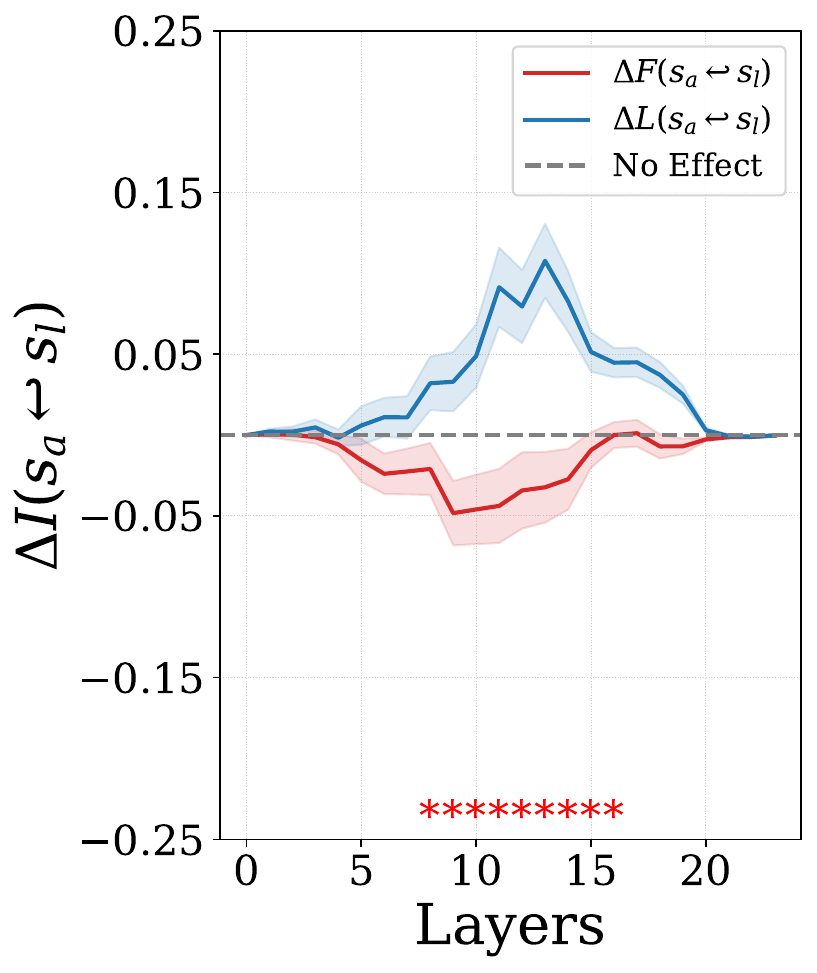}
  \caption{Idiom $\hookleftarrow$ Literal}
  \label{fig:replace_idiom_lit_qwen0.5b}
\end{subfigure}
\begin{subfigure}[b]{0.23\linewidth}
  \centering
  \includegraphics[width=\linewidth]{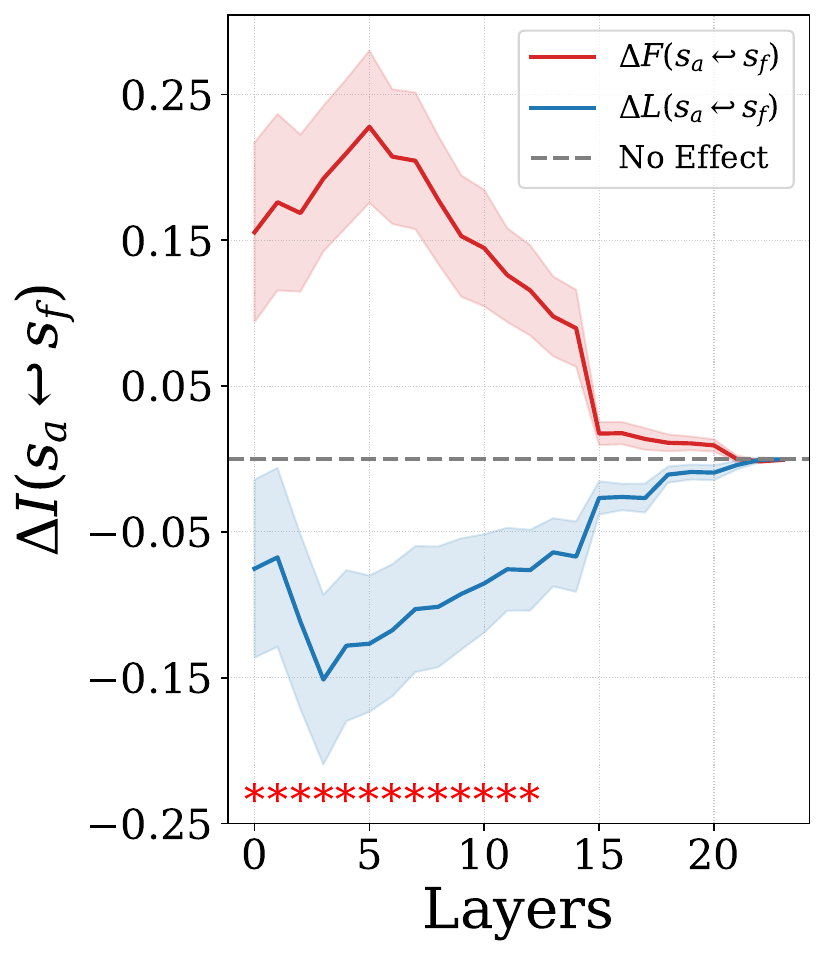}
  \caption{Idiom $\hookleftarrow$ Figurative, Idiom}
  \label{fig:replace_fig_idiom_qwen0.5b}
\end{subfigure}
\begin{subfigure}[b]{0.23\linewidth}
  \centering
  \includegraphics[width=\linewidth]{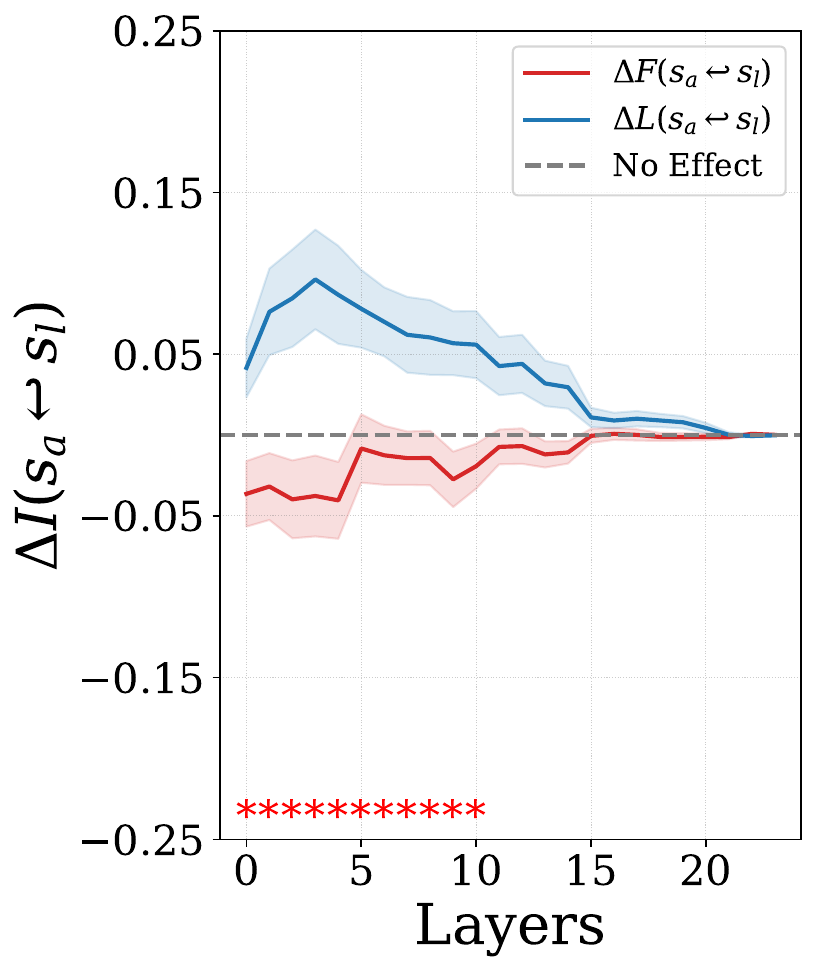}
  \caption{Idiom $\hookleftarrow$ Literal, Idiom}
  \label{fig:replace_lit_idiom_qwen0.5b}
\end{subfigure}
{\captionsetup{type=figure}
 \caption*{\textbf{Llama3.1-8B}}}
\begin{subfigure}[b]{0.23\linewidth}
  \centering
  \includegraphics[width=\linewidth]{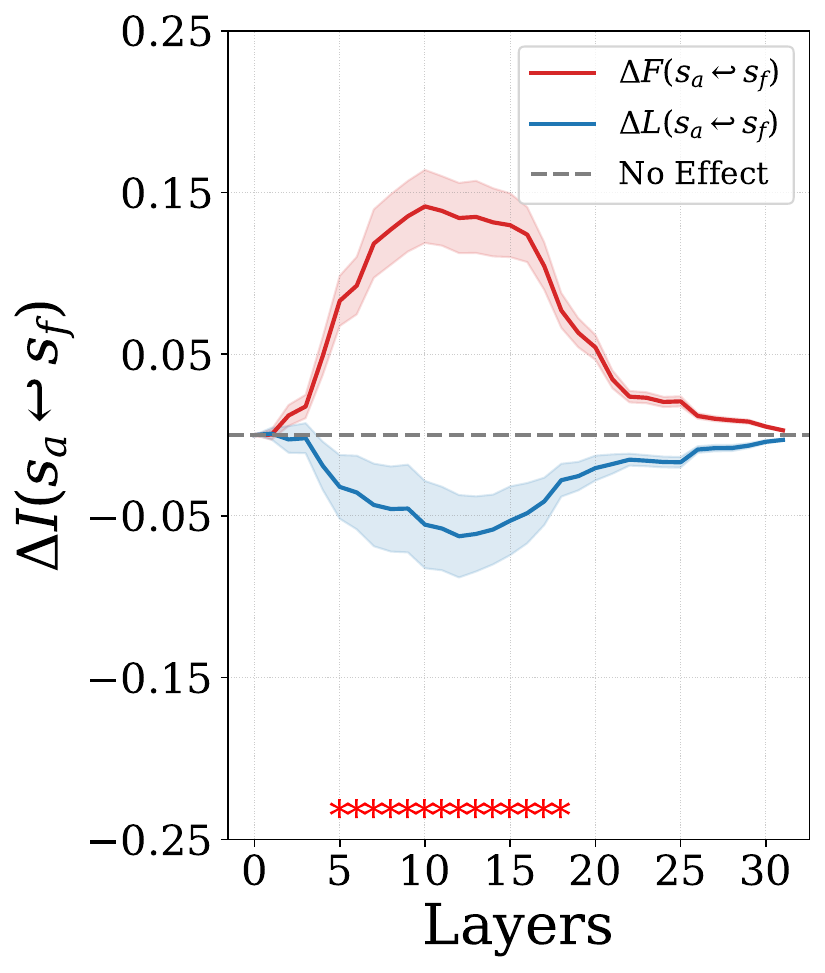}
  \caption{Idiom $\hookleftarrow$ Figurative}
  \label{fig:replace_idiom_fig_llama8b}
\end{subfigure}
\begin{subfigure}[b]{0.23\linewidth}
  \centering
  \includegraphics[width=\linewidth]{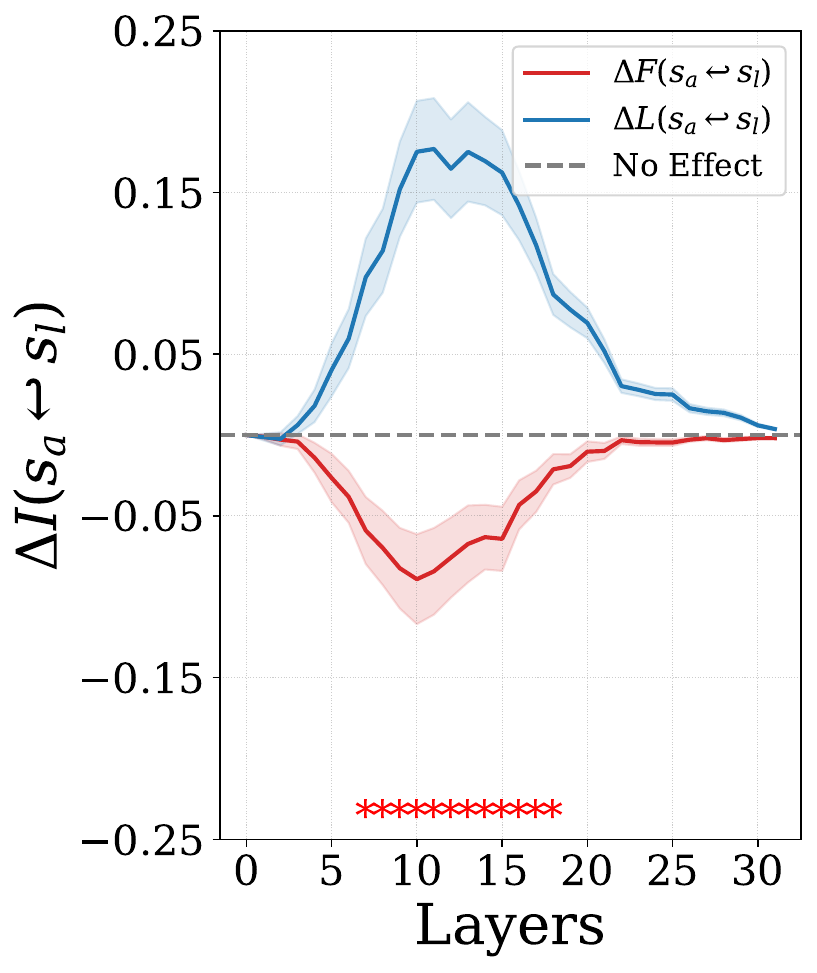}
  \caption{Idiom $\hookleftarrow$ Literal}
  \label{fig:replace_idiom_lit_llama8b}
\end{subfigure}
\begin{subfigure}[b]{0.23\linewidth}
  \centering
  \includegraphics[width=\linewidth]{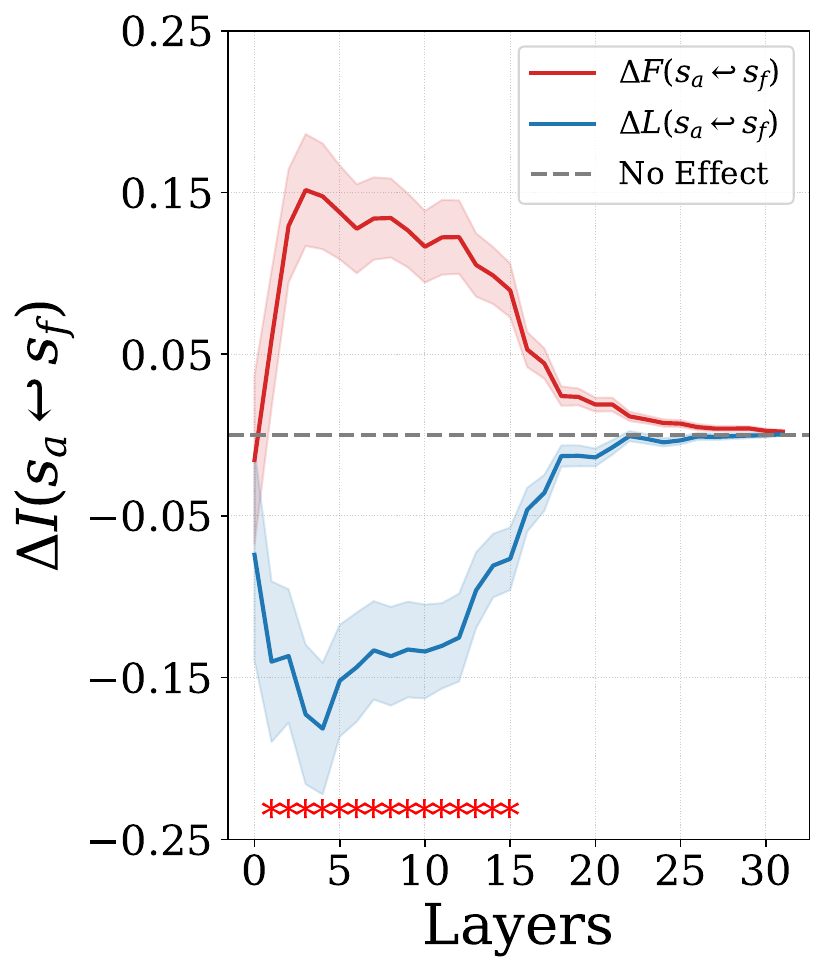}
  \caption{Idiom $\hookleftarrow$ Figurative, Idiom}
  \label{fig:replace_fig_idiom_llama8b}
\end{subfigure}
\begin{subfigure}[b]{0.23\linewidth}
  \centering
  \includegraphics[width=\linewidth]{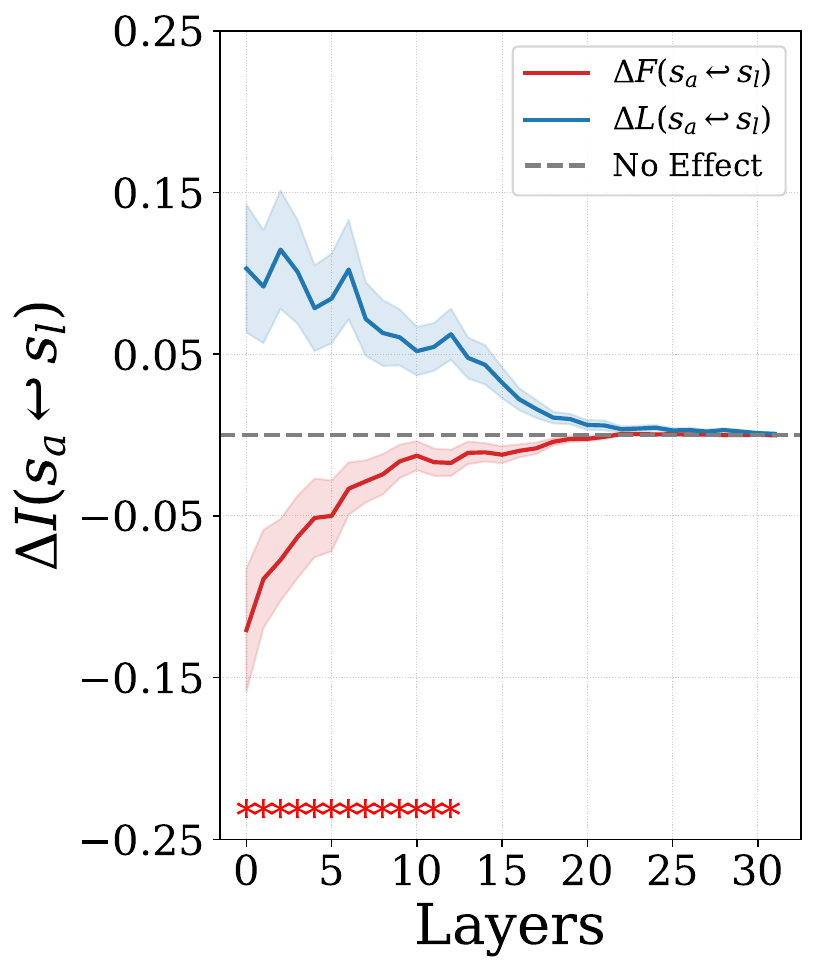}
  \caption{Idiom $\hookleftarrow$ Literal, Idiom}
  \label{fig:replace_lit_idiom_llama8b}
\end{subfigure}
{\captionsetup{type=figure}
 \caption*{\textbf{Qwen2.5-7B}}}
\begin{subfigure}[b]{0.23\linewidth}
  \centering
  \includegraphics[width=\linewidth]{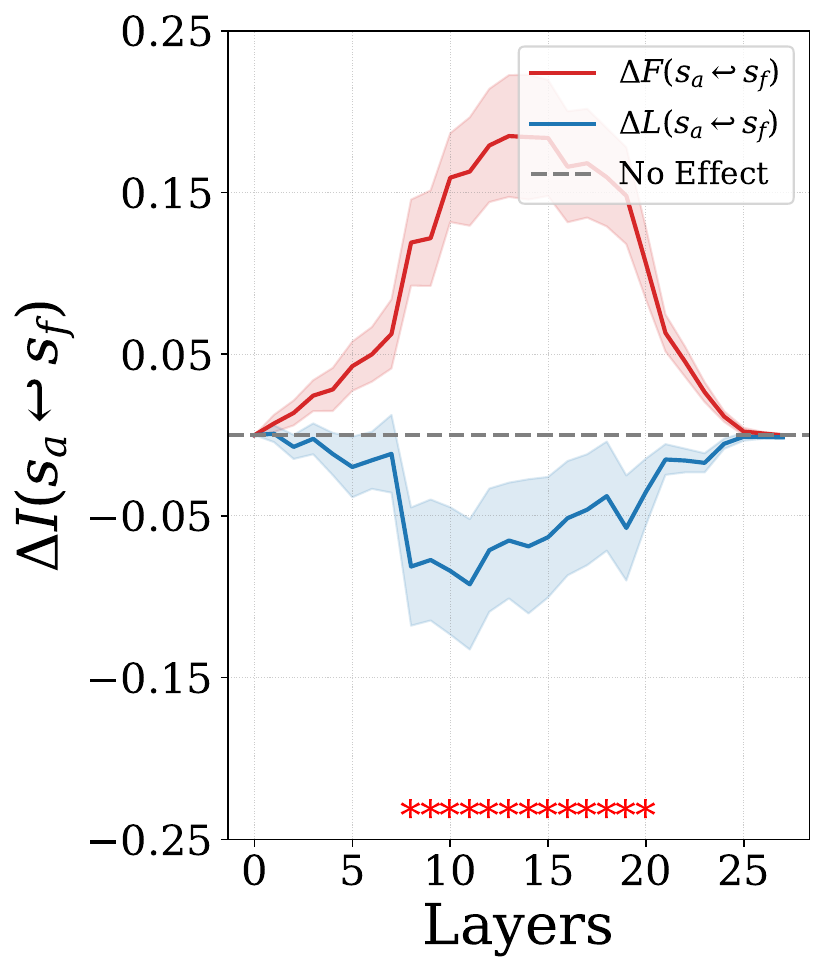}
  \caption{Idiom $\hookleftarrow$ Figurative}
  \label{fig:replace_idiom_fig_qwen7b}
\end{subfigure}
\begin{subfigure}[b]{0.23\linewidth}
  \centering
  \includegraphics[width=\linewidth]{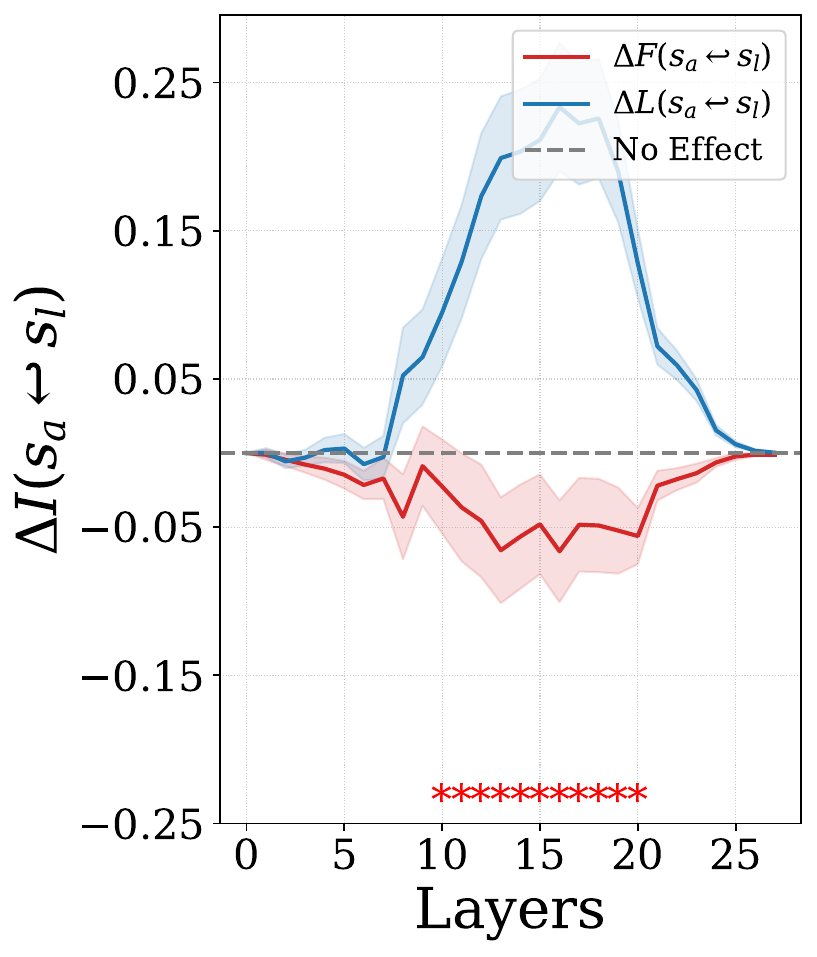}
  \caption{Idiom $\hookleftarrow$ Literal}
  \label{fig:replace_idiom_lit_qwen7b}
\end{subfigure}
\begin{subfigure}[b]{0.23\linewidth}
  \centering
  \includegraphics[width=\linewidth]{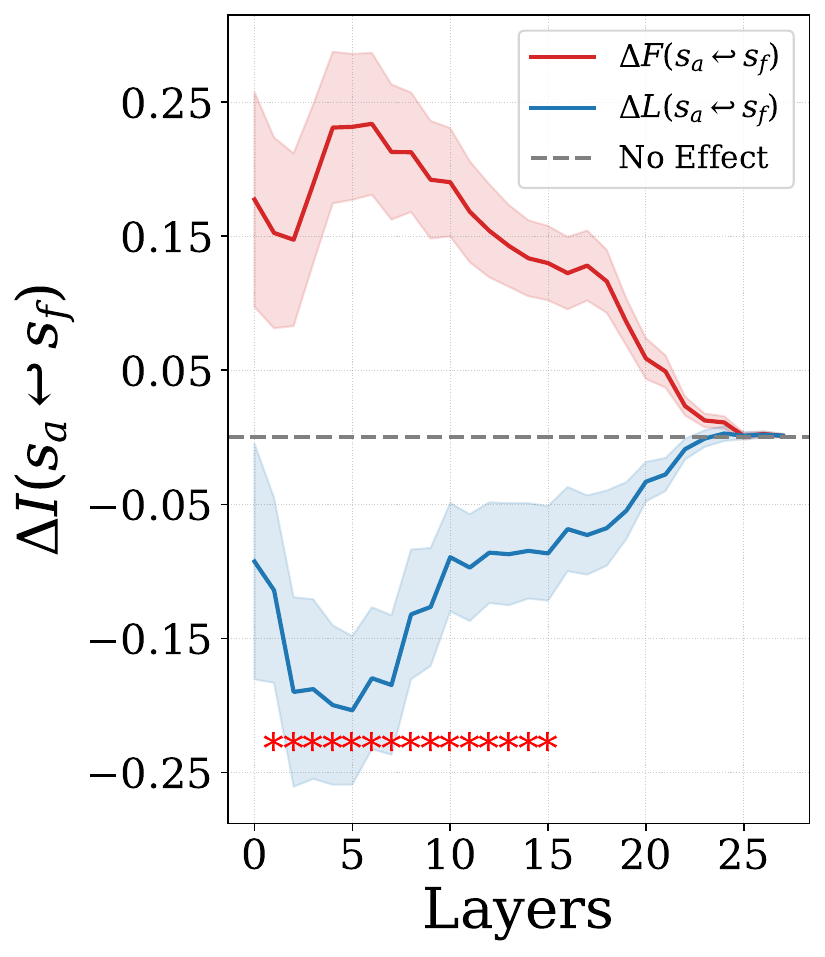}
  \caption{Idiom $\hookleftarrow$ Figurative, Idiom}
  \label{fig:replace_fig_idiom_qwen7b}
\end{subfigure}
\begin{subfigure}[b]{0.23\linewidth}
  \centering
  \includegraphics[width=\linewidth]{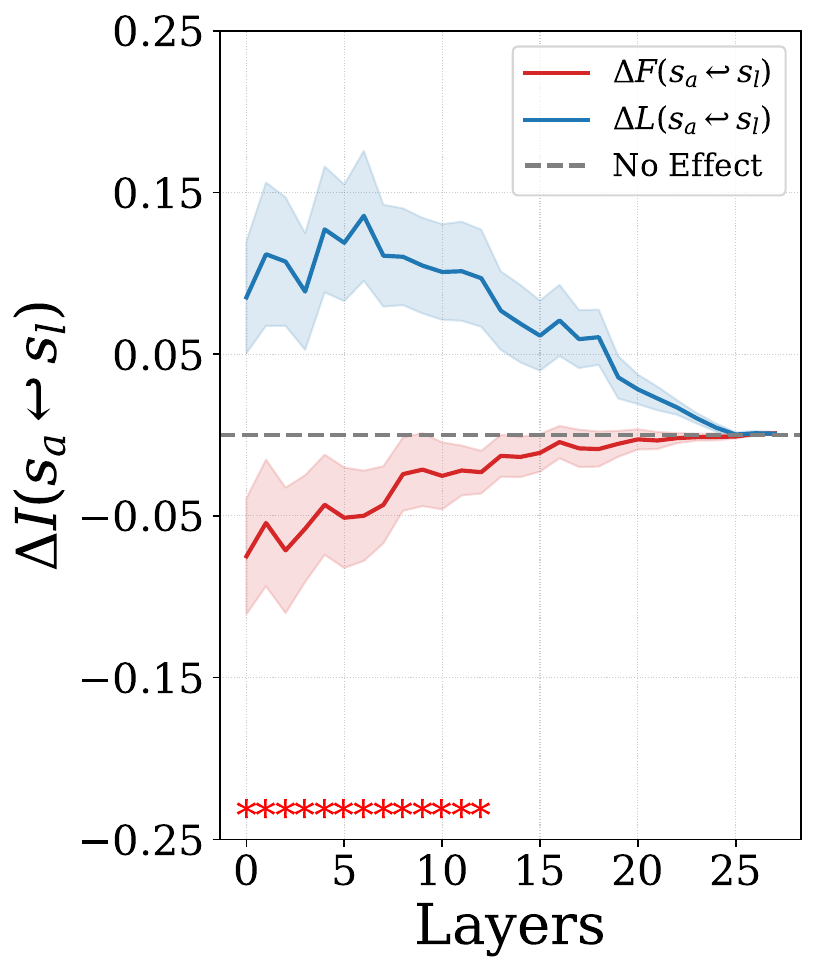}
  \caption{Idiom $\hookleftarrow$ Literal, Idiom}
  \label{fig:replace_lit_idiom_qwen7b}
\end{subfigure}
\caption{Layer-wise interpretation shift after patching in activations from $s_f$ and $s_l$ and vice versa. The red asterisk (\textcolor{red}{*}) marks layers where the difference between \textcolor[RGB]{31,119,180}{$\Delta L$} and \textcolor[RGB]{214,39,40}{$\Delta F$} exceeds the average difference across layers (paired $t$-test, $p<0.05$).}
\label{fig:information_flow_all}
\end{figure*}

Across models, the intermediate and direct pathways consistently prefer different interpretations (see Figure~\ref{fig:information_flow_all}).

\section{Knockout for context disambiguation with other models}
\label{sec:knockout_context_others}
\begin{figure*}[t]
\centering
{\captionsetup{type=figure}
 \caption*{\textbf{Qwen2.5-0.5B}}}
     \begin{subfigure}[b]{0.23\textwidth}
    \centering
    \includegraphics[width=\textwidth]{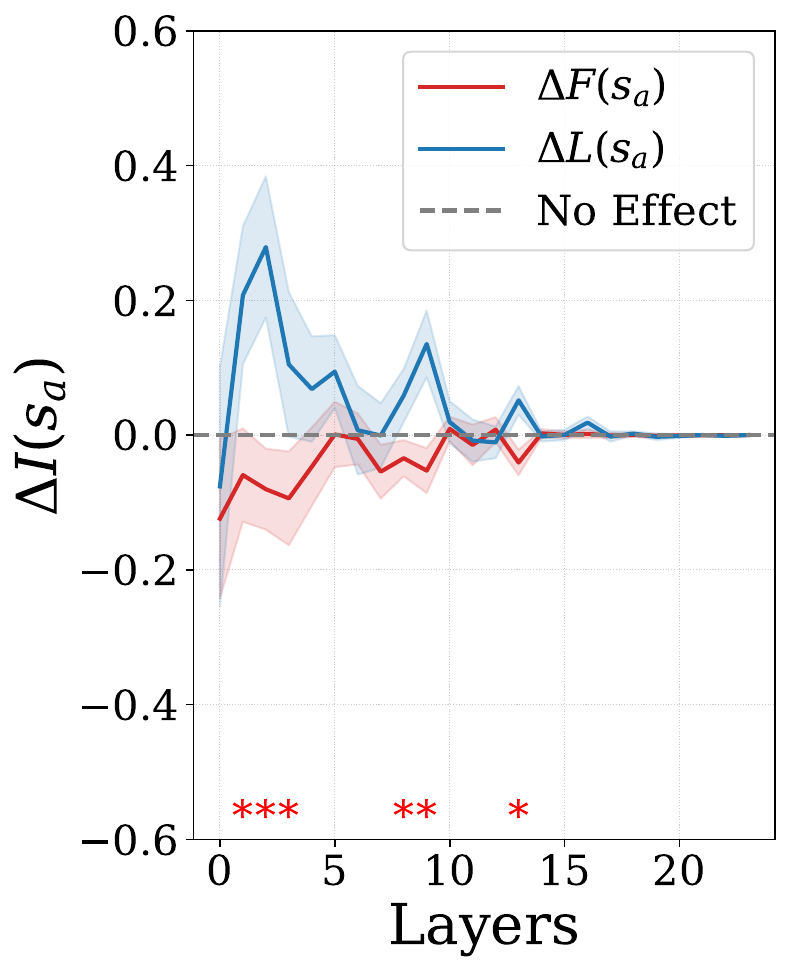}
    \caption{MLP FC}
    \label{fig:qwen0.5b_mlp_knockout_context_fig}
  \end{subfigure}
      \begin{subfigure}[b]{0.23\textwidth}
    \centering
    \includegraphics[width=\textwidth]{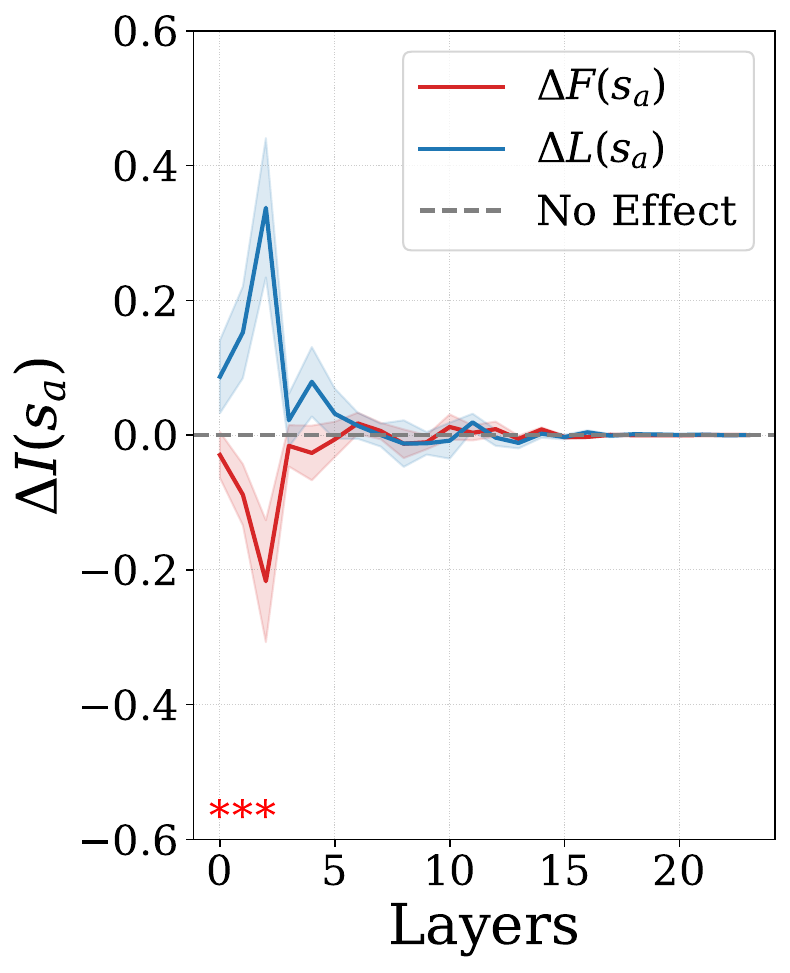}
    \caption{MHSA FC}
    \label{fig:qwen0.5b_attn_knockout_context_fig}
  \end{subfigure}
        \begin{subfigure}[b]{0.23\textwidth}
    \centering
    \includegraphics[width=\textwidth]{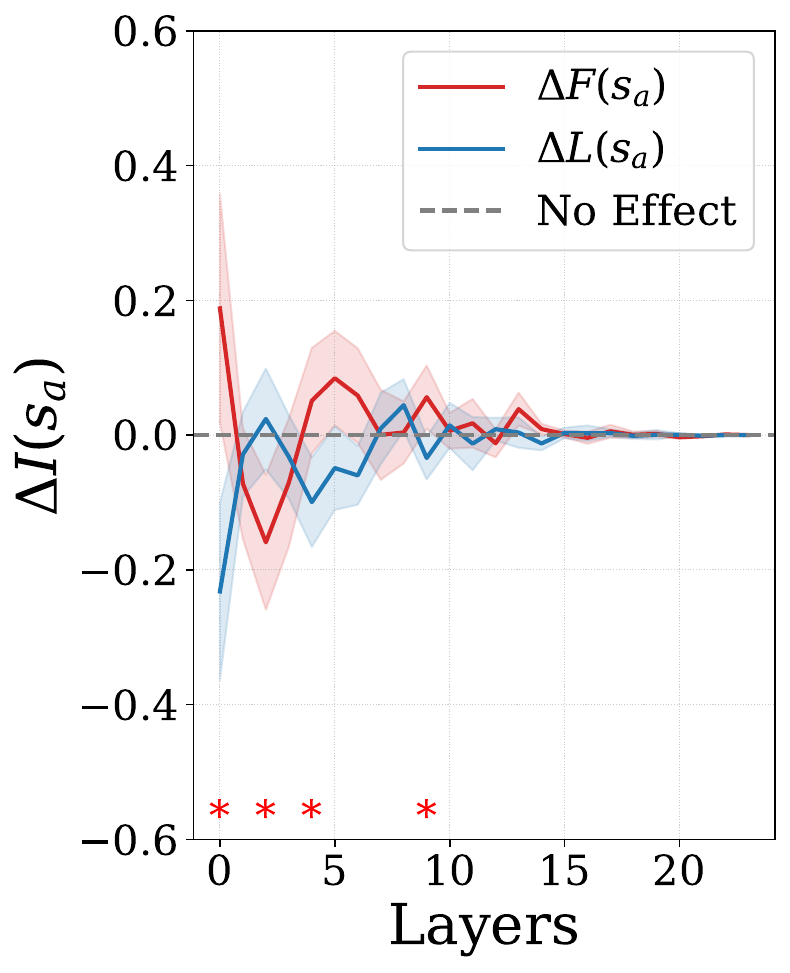}
    \caption{MLP LC}
    \label{fig:qwen0.5b_mlp_knockout_context_lit}
  \end{subfigure}
          \begin{subfigure}[b]{0.23\textwidth}
    \centering
    \includegraphics[width=\textwidth]{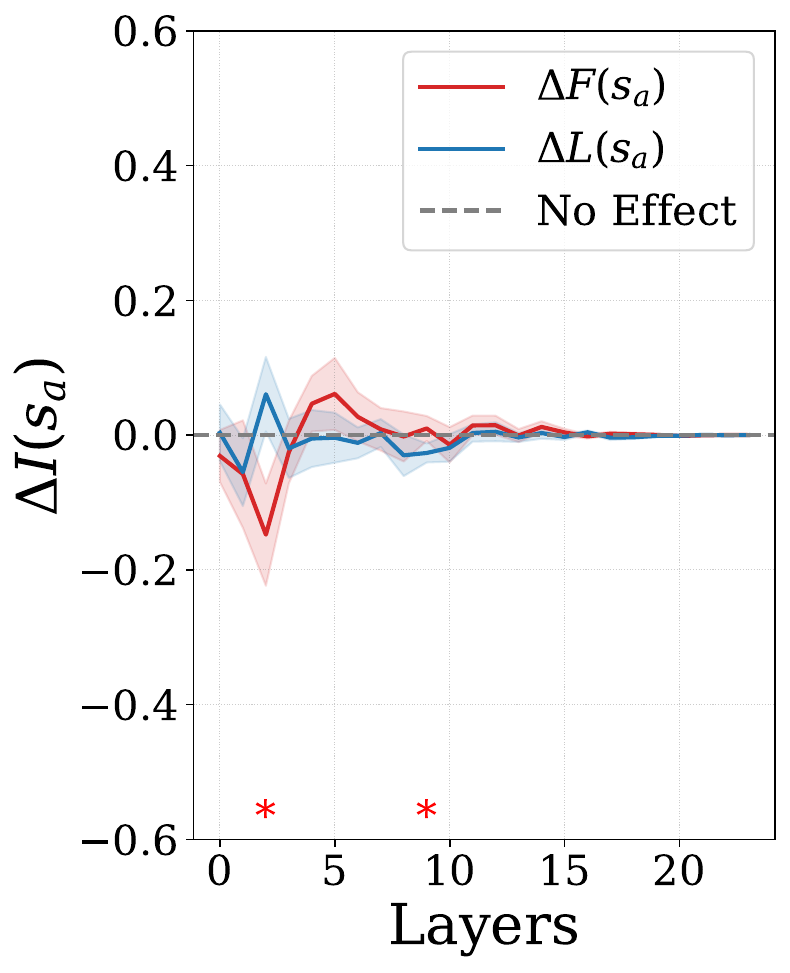}
    \caption{MHSA LC}
    \label{fig:qwen0.5b_attn_knockout_context_lit}
  \end{subfigure}
{\captionsetup{type=figure}
 \caption*{\textbf{Llama3.1-8B}}}
      \begin{subfigure}[b]{0.23\textwidth}
    \centering
    \includegraphics[width=\textwidth]{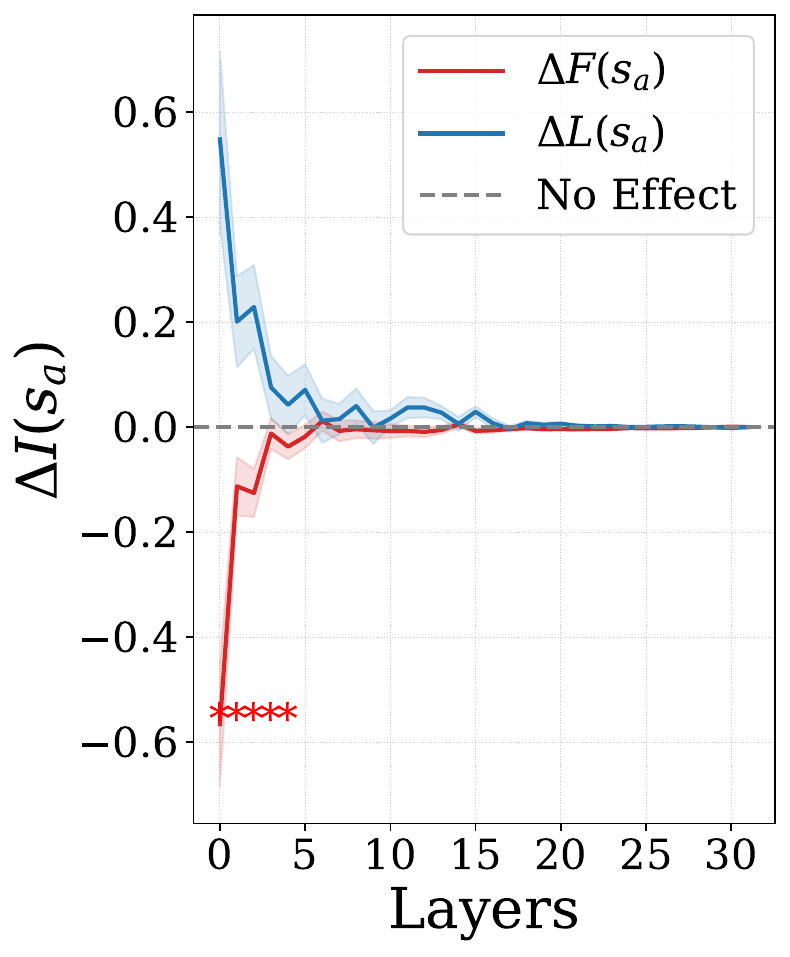}
    \caption{MLP FC}
    \label{fig:llama8b_mlp_knockout_context_fig}
  \end{subfigure}
      \begin{subfigure}[b]{0.23\textwidth}
    \centering
    \includegraphics[width=\textwidth]{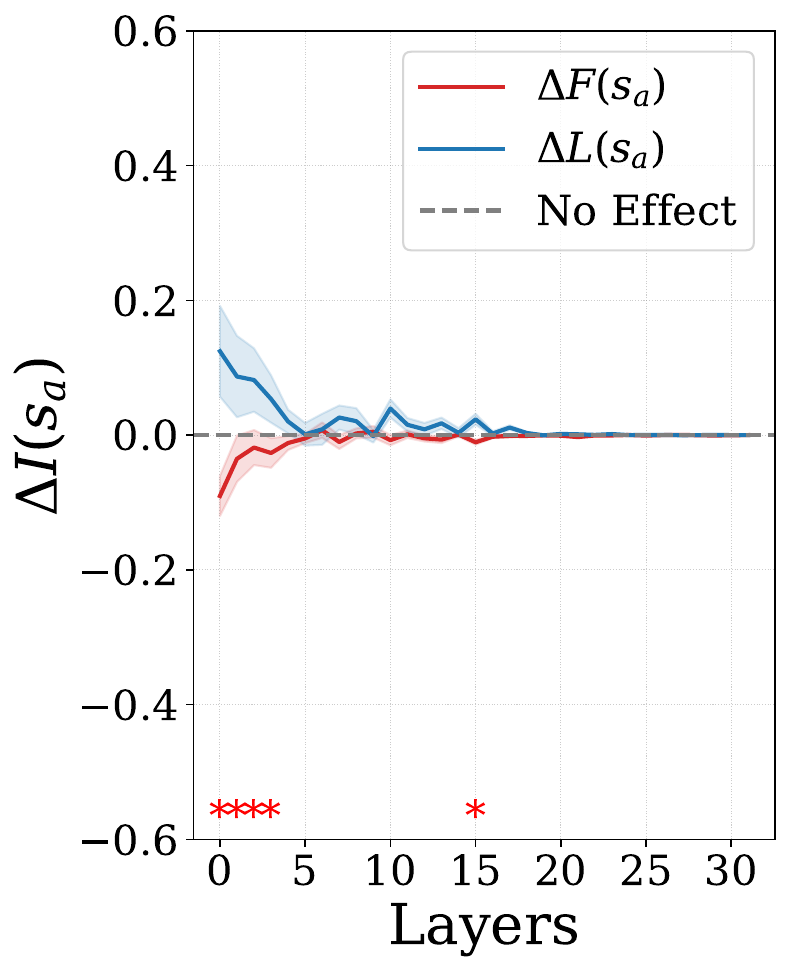}
    \caption{MHSA FC}
    \label{fig:llama8b_attn_knockout_context_fig}
  \end{subfigure}
        \begin{subfigure}[b]{0.23\textwidth}
    \centering
    \includegraphics[width=\textwidth]{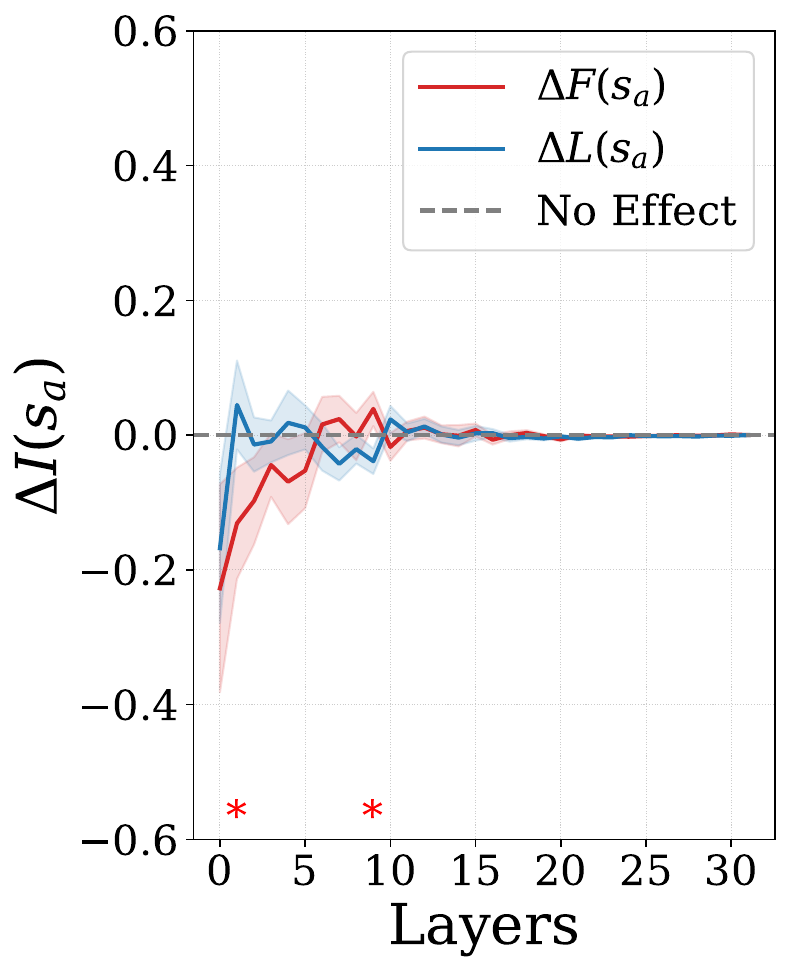}
    \caption{MLP LC}
    \label{fig:llama8b_mlp_knockout_context_lit}
  \end{subfigure}
        \begin{subfigure}[b]{0.23\textwidth}
    \centering
    \includegraphics[width=\textwidth]{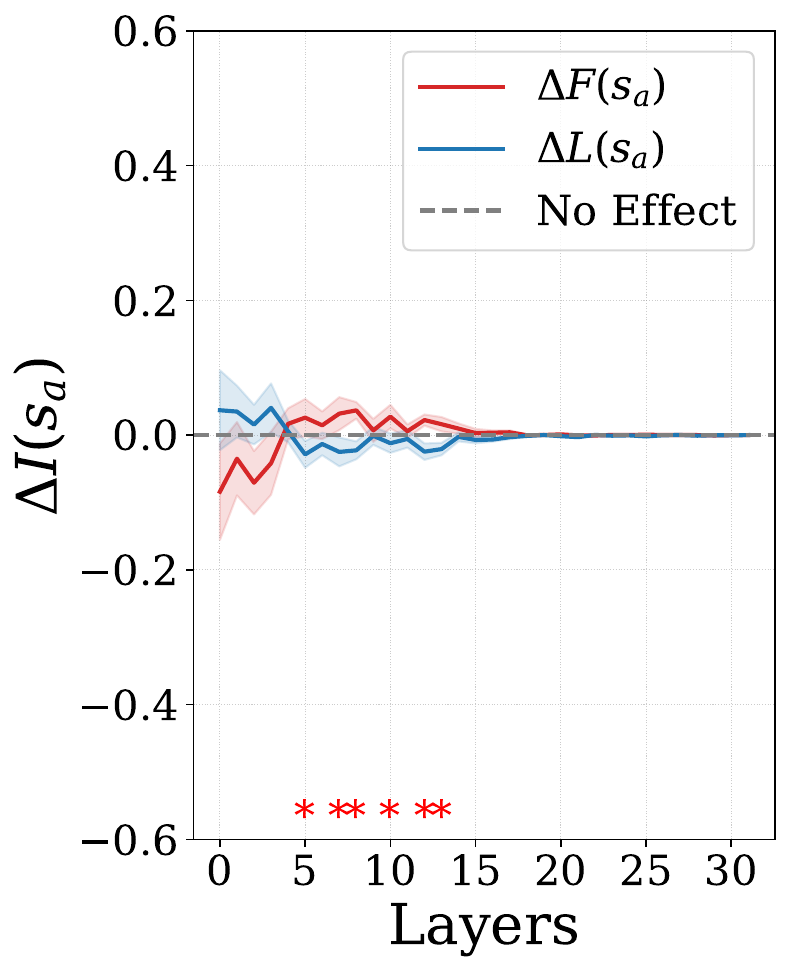}
    \caption{MHSA LC}
\label{fig:llama8b_attn_knockout_context_lit}
  \end{subfigure}

{\captionsetup{type=figure}
 \caption*{\textbf{Qwen2.5-7B}}}
     \begin{subfigure}[b]{0.23\textwidth}
    \centering
    \includegraphics[width=\textwidth]{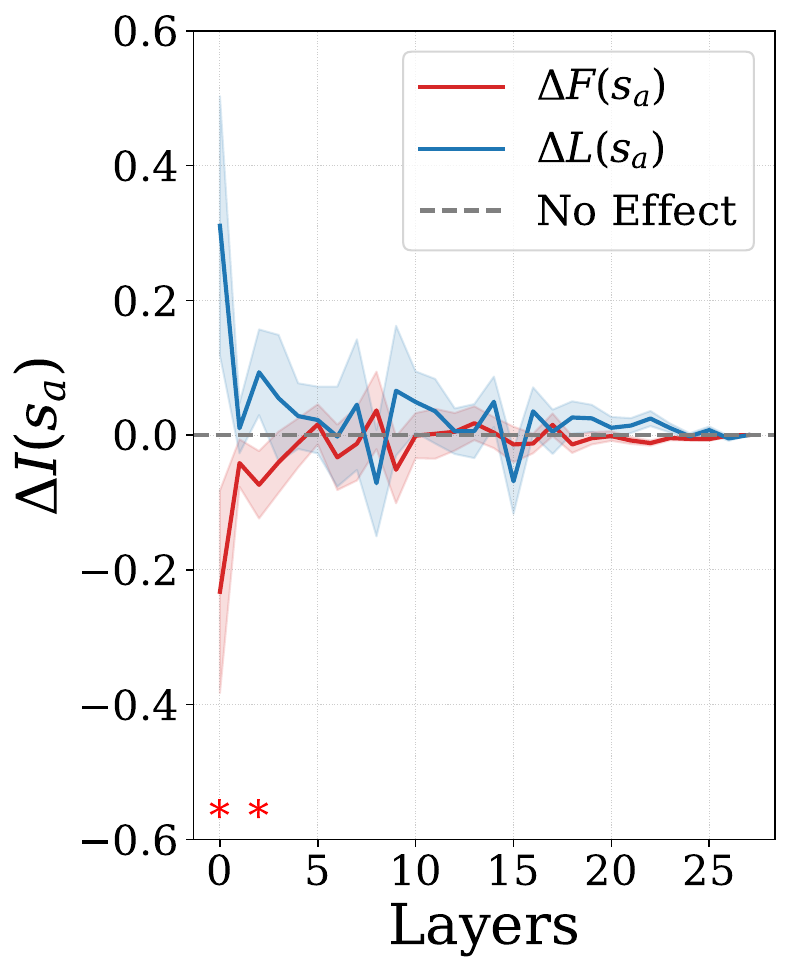}
    \caption{MLP FC}
    \label{fig:qwen7b_mlp_knockout_context_fig}
  \end{subfigure}
      \begin{subfigure}[b]{0.23\textwidth}
    \centering
    \includegraphics[width=\textwidth]{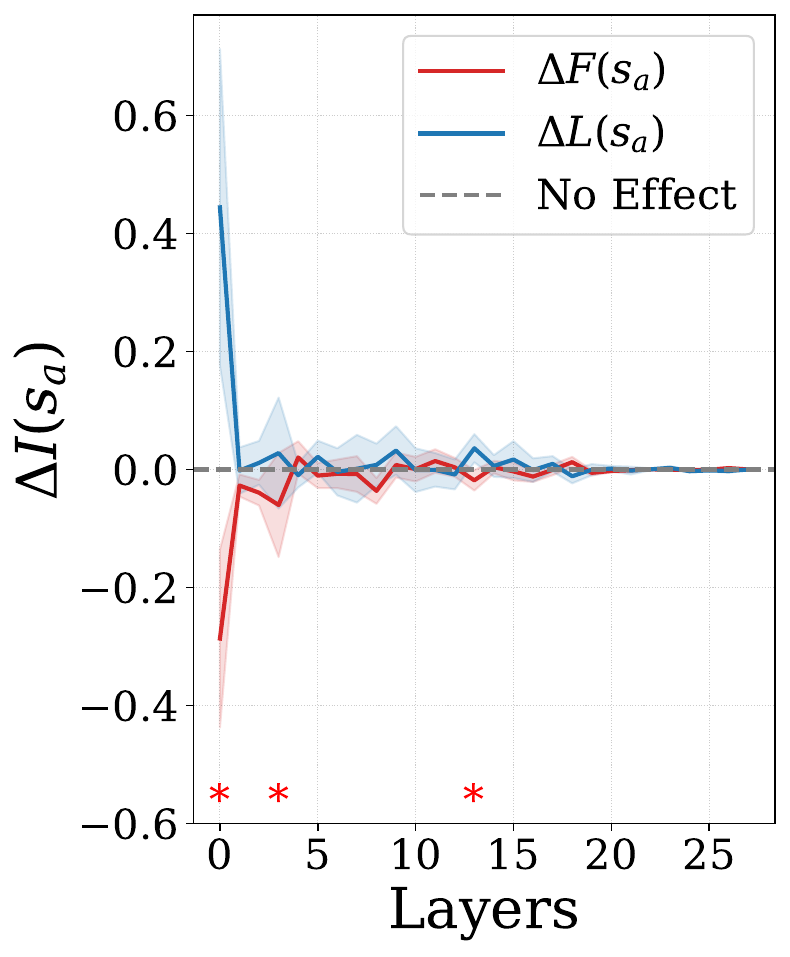}
    \caption{MHSA FC}
    \label{fig:qwen7b_attn_knockout_context_fig}
  \end{subfigure}
        \begin{subfigure}[b]{0.23\textwidth}
    \centering
    \includegraphics[width=\textwidth]{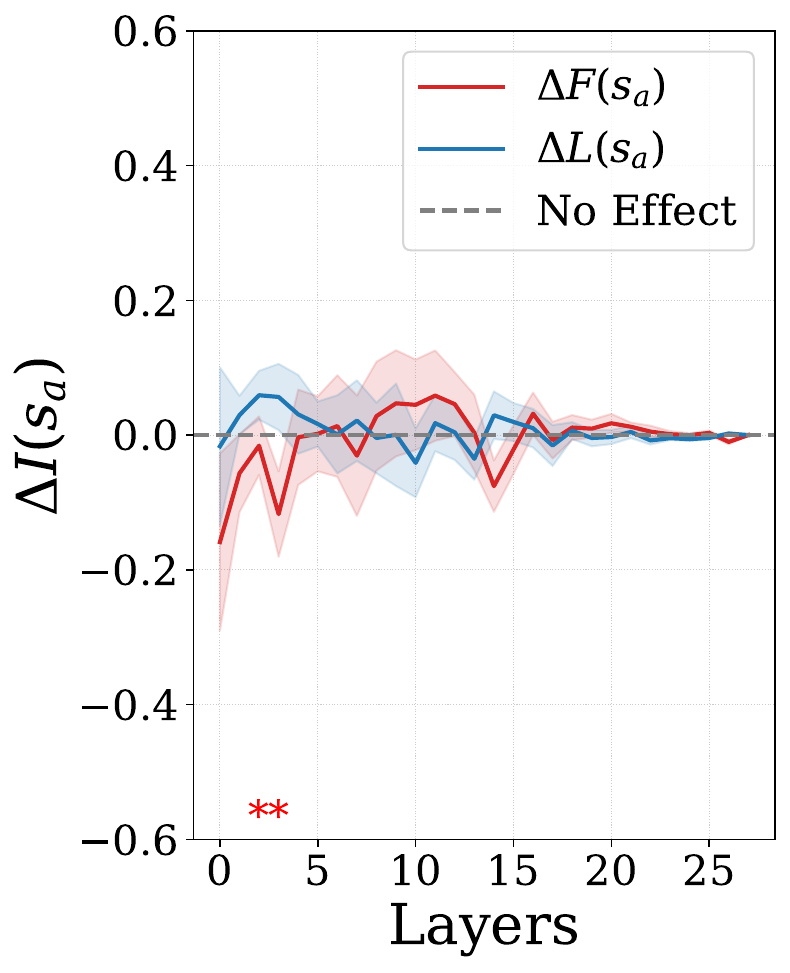}
    \caption{MLP LC}
    \label{fig:qwen7b_mlp_knockout_context_lit}
  \end{subfigure}
        \begin{subfigure}[b]{0.23\textwidth}
    \centering
    \includegraphics[width=\textwidth]{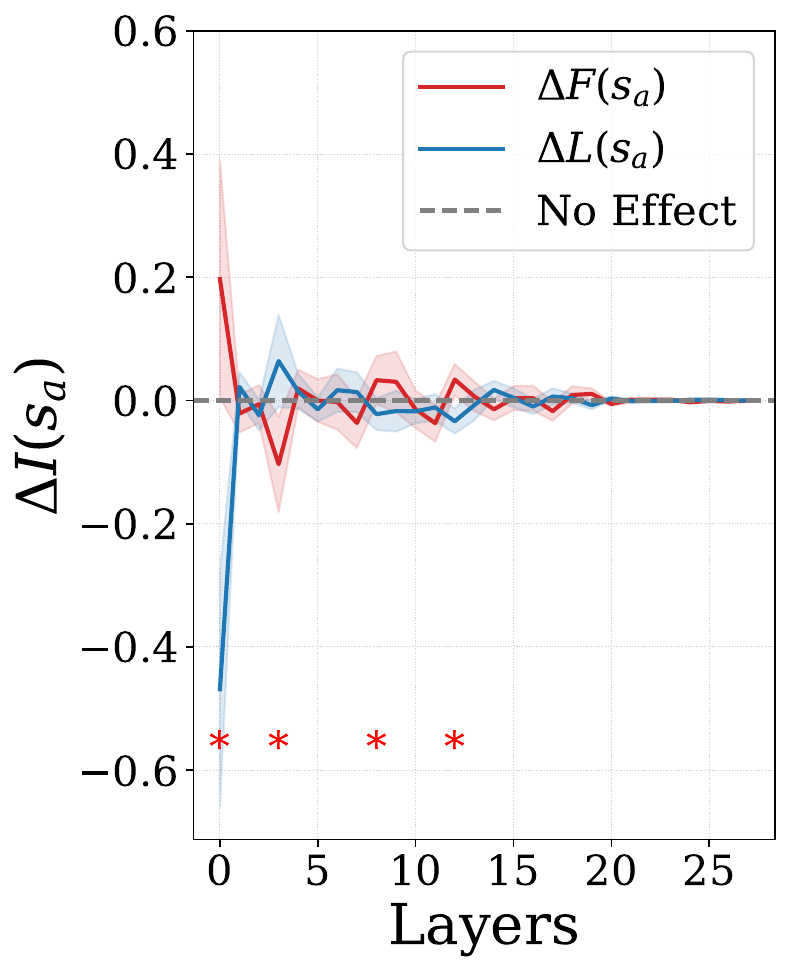}
    \caption{MHSA LC}
\label{fig:qwen7b_attn_knockout_context_lit}
\end{subfigure}

 \caption{Sublayer-wise interpretation shift $\Delta I(C+s_a)$ after ablating activations at idiom span, for contexts $C \in \{FC, LC\}$. \textbf{Y-axis:} Mean values of \textcolor[RGB]{31,119,180}{$\Delta L(C + s_a)$}, \textcolor[RGB]{214, 39, 40}{$\Delta F(C+s_a)$} with 95\% confidence intervals. \textbf{X-axis:} Layers. \textbf{Gray dashed line:} $\Delta I = 0$ (no effect). \textbf{Red asterisk (\textcolor{red}{*}):} Significant difference between \textcolor[RGB]{214, 39, 40}{$\Delta F(C+s_a)$} and the others (paired $t$-test, $p<0.05$). The difference at \textcolor{red}{*} marked layer is larger than the average difference across all layers.}
\label{fig:context_all_models}
\end{figure*}

Across different models, Figure~\ref{fig:context_all_models} consistently indicate that early MLP layers retrieve idioms' figurative meanings, and following layers promotes contextual cues via MHSA when a literal context is present.

\begin{table*}[ht]
  \small            
  \centering
  \begin{tabularx}{\textwidth}{
      >{\raggedright\arraybackslash}p{3.2cm}
      >{\raggedright\arraybackslash}p{3.2cm}
      >{\raggedright\arraybackslash}p{3.2cm}
      >{\raggedright\arraybackslash}X
      >{\raggedright\arraybackslash}X}
    \toprule
    \textbf{Ambiguous sentence} & \textbf{Literal paraphrase} &
    \textbf{Figurative paraphrase} & \textbf{Literal candidates} &
    \textbf{Figurative candidates} \\
    \midrule
    They will bend over backwards because they are so &
    They will arch spine backwards because they are so &
    They will make extra efforts because they are so &
    flexible, used, strong, weak, relaxed, tight, uncomfortable, tall, stiff, short, scared, full, comfortable, tense, small, thin, angry, over, inf, surprised &
    grateful, eager, proud, passionate, close, motivated, committed, keen, desperate, enthusiastic, invested, confident, glad, well, interested, sure, apprec, focused, attracted, dedicated \\ \hline
    He bit off more than he can chew because he was so &
    He took more food than he can swallow because he was so &
    He took on more than he could handle because he was so &
    hungry, greedy, happy, very, nervous, fam, attracted, r, thirsty, poor, star, pleased, starving, sad, delighted, hung, eng, drunk, gl, tempted &
    eager, confident, anxious, desperate, optimistic, enthusiastic, passionate, sure, determined, ambitious, focused, driven, busy, good, full, young, keen, convinced, smart, strong \\ \hline
    He would blow his own horn because he was a &
    He would blow the musical instrument because he was a &
    He would praise himself because he was a &
    professional, musician, fan, p, wind, skilled, shepherd, trumpet, member, trump, pro, fl, virt, jazz, boy, bag, human, brass, flute, musical &
    good, great, man, self, genius, proud, winner, god, true, hero, hard, legend, unique, better, brilliant, successful, clever, narciss, smart, perfection \\ \hline
    It was out in left field because it was a &
    It sat far in baseball's area because it was a &
    It was completely unrealistic because it was a &
    baseball, league, minor, sport, strong, few, popular, difficult, download, home, football, smaller, tough, sports, right, significant, basketball, pitcher, deep, stadium &
    fantasy, dream, very, completely, huge, one, product, total, movie, story, complete, perfect, romantic, fairy, totally, cartoon, fictional, massive, two, film \\ \hline
    They were up the creek because they were a &
    They were near the stream because they were a &
    They were in trouble because they were a &
    group, family, fishing, nom, part, hunting, water, tribe, pair, thirsty, party, people, river, fish, stream, traveling, farming, curious, type, peaceful &
    new, long, threat, small, minority, bunch, few, mixed, very, day, single, large, bad, man, young, mess, tiny, mix, poor, relatively \\
    \bottomrule
  \end{tabularx}
  \caption{Examples of generated data.}
  \label{tab:data_ex}
\end{table*}

\begin{table*}[ht]
  \small           
  \centering
  \begin{tabularx}{\textwidth}{
      >{\raggedright\arraybackslash}p{3.2cm}
      >{\raggedright\arraybackslash}p{3.2cm}
      >{\raggedright\arraybackslash}p{3.2cm}
      >{\raggedright\arraybackslash}X
      >{\raggedright\arraybackslash}X}
    \toprule
    \textbf{Ambiguous sentence} & \textbf{Literal paraphrase} &
    \textbf{Figurative paraphrase} & \textbf{Literal candidates} &
    \textbf{Figurative candidates} \\
    \midrule
    They left him out in the cold because he was a &
    They left him outside in the frost because he was a &
    They abandoned him without support because he was a &
    bad, stranger, little, bit, thief, trouble, poor, drunk, witch, dirty, dog, rebel, beg, danger, trait, s, nuisance, tiny, he, naughty &
    child, foreign, threat, boy, political, baby, \textbf{black}, non, reminder, \textbf{disabled}, male, burden, cripp, product, \textbf{minority}, son, member, difficult, minor, different \\ \hline
    They let him off the hook because he was a &
    They removed him from fishing tackle because he was a &
    They freed him from responsibility because he was a &
    threat, bad, convicted, little, bit, danger, sexual, fish, ped, \textbf{white}, criminal, racist, bully, \textbf{black}, terrorist, thief, political, jerk, rap, big &
    \textbf{minor}, child, foreign, slave, kid, good, mad, juvenile, victim, boy, stranger, young, youth, mere, student, fool, first, teenager, prisoner, friend \\ \hline
    She was looking for a needle in a haystack because she was a &
    She was searching for a needle within a straw because she was a &
    She was facing an impossible search because she was a &
    needle, craft, seam, straw, tiny, hay, detective, master, haystack, witch, farmer, pro, camel, professional, busy, crazy, giant, chicken, cow, neat &
    \textbf{woman}, victim, single, young, \textbf{black}, \textbf{female}, \textbf{girl}, \textbf{slave}, mother, stranger, new, first, non, private, \textbf{white}, novice, \textbf{minority}, prisoner, foreign, mom \\
    \bottomrule
  \end{tabularx}
  \caption{Biased data examples reflecting stereotypes embedded in the language model.}
  \label{tab:biased_data_ex}
\end{table*}

\end{document}